\documentclass[10pt,article,compsoc]{IEEEtran}

\ifCLASSOPTIONcompsoc
  \usepackage[nocompress]{cite}
\else
  \usepackage{cite}
\fi

\usepackage{diagbox}
\usepackage{lipsum}
\usepackage{arydshln}
\usepackage{epsfig}
\usepackage{graphicx}
\usepackage{amsmath}
\usepackage{amssymb}
\usepackage[ruled,vlined]{algorithm2e}
\usepackage{color}
\usepackage{booktabs}
\usepackage{multirow}
\usepackage{threeparttable}
\usepackage[normalem]{ulem}
\usepackage{makecell}
\usepackage{subfigure}
\usepackage{mathrsfs}

\usepackage{amsthm,amsmath,amssymb}

\usepackage{colortbl}
\usepackage{pifont}
\usepackage{xcolor}
\usepackage{array}

\usepackage[pagebackref=false,breaklinks=true,colorlinks,bookmarks=false]{hyperref}

\newcommand{\yxa}{\textcolor[rgb]{0.0,0.0,0.0}}

\newcommand{\md}{\textcolor[rgb]{0.0,0.0,0.0}}

\newcommand{\major}{\textcolor[rgb]{0,0,0}}

\makeatletter
\DeclareRobustCommand\onedot{\futurelet\@let@token\@onedot}
\def\@onedot{\ifx\@let@token.\else.\null\fi\xspace}
\def\eg{\emph{e.g}\onedot} 
\def\ie{\emph{i.e}\onedot} 
 
\def\etc{\emph{etc}\onedot} \def\vs{\emph{vs}\onedot}
 
\def\etal{\emph{et al}\onedot}
\definecolor{yellow}{RGB}{255,255,220}
\definecolor{lightyellow}{RGB}{255,220,121}
\definecolor{lightgray}{RGB}{210,210,210}
\definecolor{gray}{rgb}{0.8, 0.9, 1.0}
\definecolor{blue}{RGB}{157,181,223}

\begin{document}
{\onecolumn

\noindent \vspace{1cm}

\noindent \textbf{\huge{A Versatile Framework for Multi-scene Person\\\\Re-identification}}

\vspace{2cm}

\noindent {\LARGE{Wei-Shi Zheng, Junkai Yan, Yi-Xing Peng}}
\\\\
\Large{Code: \href{https://github.com/iSEE-Laboratory/VersReID}{{https://github.com/iSEE-Laboratory/VersReID}}}

\vspace{1cm}

\noindent\Large{Submission date: 05-Jun-2023 to IEEE Transactions on Pattern Analysis and Machine Intelligence}

\vspace{1cm}

\noindent\Large{For reference of this work, please cite:}

\vspace{1cm}
\noindent\Large{Wei-Shi Zheng, Junkai Yan, Yi-Xing Peng.
``A Versatile Framework for Multi-scene Person Re-identification''. \textit{IEEE Transactions on Pattern Analysis and Machine Intelligence}, 2024.}

\vspace{1cm}
\Large{
\noindent Bib:\\
\noindent @article\{zheng2024versreid,\\
\ \ \  title     = \{A Versatile Framework for Multi-scene Person Re-identification\}, \\
 \ \ \   author    = \{Zheng, Wei-Shi and Yan, Junkai and Peng, Yi-Xing\},\\
\ \ \  journal   = \{IEEE Transactions on Pattern Analysis and Machine Intelligence\},\\
\ \ \  year      = \{2024\}\\
\}
}}
\newpage

{
\twocolumn

\title{A Versatile Framework for\\Multi-scene Person Re-identification}

\author{
	Wei-Shi~Zheng,~Junkai~Yan,~and~Yi-Xing~Peng%
    \IEEEcompsocitemizethanks{
        \IEEEcompsocthanksitem W.-S. Zheng is with the School of Computer Science and Engineering, Sun Yat-sen University, Guangzhou 510275, China, and also with the Key Laboratory of Machine Intelligence and Advanced Computing, Ministry of Education, 510006, China. E-mail: wszheng@ieee.org.
        \IEEEcompsocthanksitem J. Yan and Y.-X Peng are with the School of Computer Science and Engineering, Sun Yat-sen University, Guangzhou 510275, China. E-mail: yanjk3@mail2.sysu.edu.cn, pengyx23@mail2.sysu.edu.cn.
    }
}



\IEEEtitleabstractindextext{%
\begin{abstract} 
Person Re-identification (ReID) has been extensively developed for a decade in order to learn the association of images of the same person across non-overlapping camera views. To overcome significant variations between images across camera views, mountains of variants of ReID models were developed for solving a number of challenges, such as resolution change, clothing change, occlusion, modality change, and so on. Despite the impressive performance of many ReID variants, these variants \md{typically} function distinctly and cannot be applied to other challenges. To our best knowledge, there is no versatile ReID model that can handle various ReID challenges at the same time. This work contributes to the first attempt at learning a versatile ReID model to solve such a problem. Our main idea is to form a \md{two-stage prompt-based twin modeling framework called \textbf{VersReID}. Our VersReID firstly leverages the scene label to train a ReID Bank that contains abundant knowledge for handling various scenes, where several groups of scene-specific prompts are used to encode different scene-specific knowledge. In the second stage, we distill a \textbf{V-Branch} model with versatile prompts from the ReID Bank for adaptively solving the ReID of different scenes, eliminating the demand for scene labels during the inference stage.} To facilitate training VersReID, we further introduce the multi-scene properties into self-supervised learning of ReID via a multi-scene prioris data augmentation (MPDA) strategy. Through extensive experiments, we demonstrate the success of learning an effective and versatile ReID model for handling ReID tasks under multi-scene conditions without manual assignment of scene labels in the inference stage, including general, low-resolution, clothing change, occlusion, and cross-modality scenes.
Codes and models are available at \href{https://github.com/iSEE-Laboratory/VersReID}{{https://github.com/iSEE-Laboratory/VersReID}}.
\end{abstract}

\begin{IEEEkeywords}
person re-identification, visual surveillance
\end{IEEEkeywords}}

\maketitle

\IEEEdisplaynontitleabstractindextext

\IEEEpeerreviewmaketitle

\IEEEraisesectionheading{
\section{Introduction}
\label{s.1}}
\begin{figure}[t]
  \centering
  \includegraphics[width=1\linewidth]{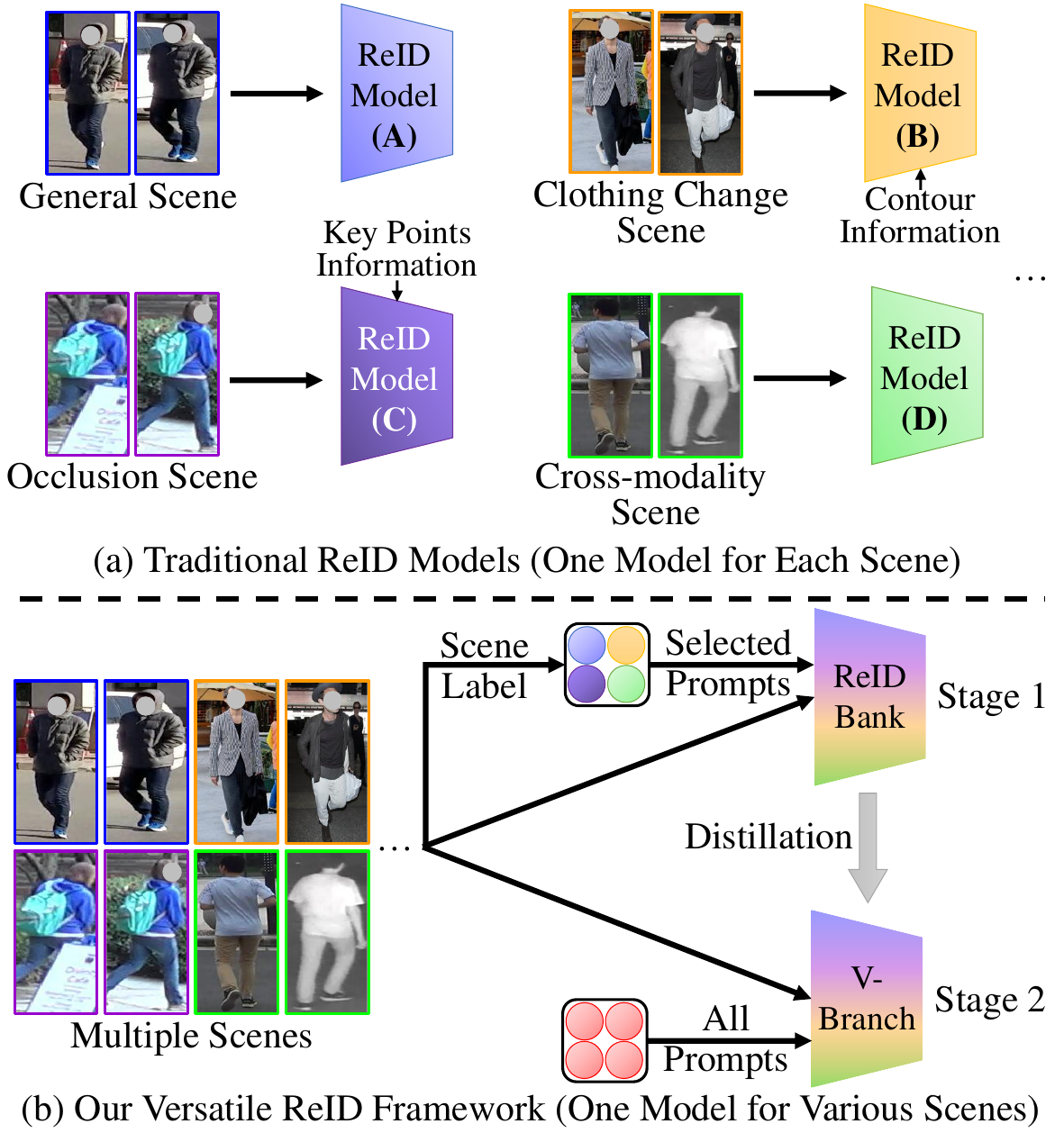}\vspace{-1em}
  \caption{(a) Traditional frameworks train distinctive models for different scenes and usually require auxiliary information such as the contour and key points information. (b) Differently, our versatile ReID framework (VersReID) can solve multiple ReID scenes simultaneously without auxiliary information. The colored circles in the figure indicate prompts.}\vspace{-0.8em}
   \label{f.1}
\end{figure}

Person Re-identification (ReID) aims to retrieve the same identity across various non-overlapping camera views, serving as a fundamental function in visual surveillance systems.
The ReID is challenging due to significant appearance variations of a person across camera views, including pose variation, lighting variation, resolution variation, clothing change, occlusion, modality variation, \etc. 
In the last decade, significant processes have been witnessed in ReID. 
A number of ReID models have been developed to solve the pose variation problem~\cite{pose_li_2022_pami,pvpm,pfd}, the low-resolution problem~\cite{zheng_2022,dslr_market,cheng_2020,li_2015}, the clothing change problem~\cite{iccv-2-cc,iccv-4-cc,prcc,fsam,chen_2021_cvpr_3d,cg_chen_2022_tmm,cg_wu_2017,ccfa}, the occlusion problem~\cite{occ_zheng_2015,occ_yang_2021,pgfa,pat}, and the cross-modality search ReID~\cite{wu2017rgb,zhang2022fmcnet,mm01,zheng_2020_pr_cm,jiawei,partmix,deen,iccv-3-cm,iccv-5-cm,iccv-6-cm,iccv-7-cm}.

Although performing specific ReID modeling for solving each specific challenge seems more effective (as illustrated by Figure~\ref{f.1} (a)), it hampers the deployment of ReID models because the challenges in coming images are always unpredictable, and thus it is impractical to switch different models for processing different images.
Besides, those challenges are \md{usually} existing at the same time, while existing models are supposed to solve only one challenge.
The above concerns hurt the scalability of the ReID system.
Moreover, despite the promising performance, most methods are built with auxiliary models for additional discriminative cues, 
such as key points~\cite{pfd,pgfa} or contour information~\cite{prcc,fsam}, making them inflexible and difficult to deploy in practice.

In this work, for the first time, we aim to learn a versatile person ReID model to handle various ReID scenes\footnote{In this work, we regard each specific ReID problem/challenge as a specific scene, \eg, the low-resolution ReID scene indicates the ReID problem under the low-resolution condition~\cite{dslr_market,zheng_2022}.\label{scene}} simultaneously without additional hand-craft or predicted auxiliary information used in the inference stage.
To achieve this goal, we work on a multi-scene ReID dataset based on existing ReID datasets designed for various challenges to facilitate a versatile ReID model.
Different from an intuitive solution for forming a versatile model by training a model on the multi-scene dataset, we propose a two-stage prompt-based twin modeling framework called \textbf{VersReID}. 
Our VersReID aims to distill a universal model specified by versatile prompts from a ReID Bank that is learned on the multi-scene ReID dataset so that VersReID is able to process different ReID scenes adaptively.

Specifically, for learning the \textbf{ReID Bank} in the first stage of our VersReID, we utilize the learnable prompts to adaptively model scene-specific knowledge with the help of scene labels\footnote{\md{As the multi-scene ReID dataset is based on existing ReID datasets, we assign a scene label to each dataset according to its main property. For example, the PRCC~\cite{prcc} dataset is designed for the clothing change ReID problem. Thus, its scene label is ``clothing change''.}\label{scenelabels}}.
By augmenting a scene-shared backbone containing scene-invariant knowledge via scene-specific prompts, we form a powerful ReID Bank to handle multi-scene ReID under the guidance of scene labels.
Since we expect our ReID model to work in a versatile way without manually assigning scene labels in the inference stage, we learn a sister stream network \textbf{V-Branch} with a group of versatile prompts, which is the second stage of the VersReID. 
The distillation enables simultaneously exploiting different specific knowledge encoded by scene-specific prompts, which is unavailable in the intuitive solution, so that the knowledge is unified into V-Branch. Thus, V-Branch can extract unified feature representation for images across multiple scenes without using scene labels.
Notably, our modeling is different from conventional multi-task learning~\cite{mt1,mt2,mt3,mt4,mt5,hardparashare,maskrcnn}, because they either require explicit task indicators\footnote{In multi-task learning, there is explicit differentiation among tasks, \eg, classification, detection, and image generation, and therefore they have to leverage a task indicator to distinguish different tasks.}\label{tasklabels} to instruct which task to handle~\cite{mt1} or execute all tasks at the same time~\cite{maskrcnn}. 
Conversely, our ReID model can work under various scenes without accessing scene labels in the inference stage.
Besides, beyond the elaborately designed task-specific layers required by traditional multi-task learning, we employ scene-specific prompts for effectively adapting the ReID backbone to multiple scenes. Since the prompts modeling is independent of the backbone, our learning process is more flexible without manually designing which layer to share or not.

In addition, we introduce a Multi-scene Prioris Data Augmentation (MPDA) strategy for the \emph{pre-training stage} of ReID by simulating each ReID scene via generating a specific view for each image, where the view contains similar visual characteristics to the corresponding scene, so as to guide the model to overcome the cross-scene variations in a self-supervised contrastive manner~\cite{dino,mocov3}.
By applying the MPDA to typical self-supervised contrastive learning methods~\cite{dino,mocov3}, we can learn our model on the large-scale unlabeled person dataset LUPerson~\cite{lup} in a self-supervised manner for improving the generalization ability on a wide range of ReID scenes.

In our experiments, we demonstrate learning an effective and versatile ReID model for simultaneously handling ReID tasks under different scenes, including the general~\cite{market,msmt}, low-resolution~\cite{dslr_market,zheng_2022}, clothing change~\cite{prcc,cele}, occlusion~\cite{pgfa,occ_zheng_2015}, and cross-modality scenes~\cite{mm01}. 
We also build a multi-scene joint testing set where different scenes occur and show that our proposed VersReID achieves impressive performance not only in each scene separately but also in multi-scene conditions.
Moreover, we provide extensive ablation studies to analyze how the proposed VersReID works.
In summary, for the first time, we show the probability of constructing a versatile model for handling multiple ReID scenes that frequently appear in realistic scenarios.

 \vspace{-10pt}
\section{Related Works}
\label{s.2}
\subsection{Person ReID under Different Scenes}
\label{s.2.1}
The pioneering works on ReID focus on designing hand-crafted features~\cite{FE_1,FE_2,FE_3,FE_4,LOMO,FE_20,kale2004identification,kale2004fusion} to represent the identity information in the images and learning a distance metric~\cite{metric_1,metric_10,metric_2,metric_7,metric_8,metric_9, metric_20, metric_21,metric_cvpr23} so that the features of the same person are closer than those of different persons.
Due to the dramatic cross-view variations in lighting, background, and so forth, the ReID performance of hand-crafted features is far from satisfactory.

Recently, with the rapid development of deep learning, deep neural networks have been employed to learn discriminative feature representations automatically and have dominated the ReID field~\cite{pcb,transreid,q_loss,yixing,Suh_2018_ECCV,song2018mask, camera_selection}. 
Free from hand-crafted features, researchers focus on improving the network architectures~\cite{pcb,OSNET,Pyramid,ABDNet,DG7,DG9,transreid} and designing objective functions~\cite{q_loss,triplet_loss,MVP,Loss_1,consistency}.
For example, Sun~\etal~\cite{pcb} exploited the local discriminative cues in the images by multi-branch classifiers.
He~\etal~\cite{transreid} adopted the powerful transformer backbone for ReID.

Despite the impressive performance of the advanced ReID models, they are developed for general ReID. 
Unfortunately, specific challenges in practical application, including occlusion~\cite{occ_reid,pgfa,occ_zheng_2015,zhao2021incremental, pami_partial_1}, low-resolution~\cite{dslr_market,zheng_2022,zhang_2021_TIP,han2020prediction,han2021adaptive}, lighting change~\cite{mm01,wu2017rgb}, clothing change~\cite{prcc,ltcc,ccfa}, \etc, will seriously degrade the ReID performance~\cite{pami_noise_1}.

\vspace{0.6\baselineskip}\noindent\textbf{- Person ReID under low-resolution.}
Due to the different qualities of cameras and the complicated environmental factors, the resolution of the images varies from different cameras. 
Consequently, matching the low-resolution (LR) images with the high-resolution (HR) images has attracted increasing attention.
To bridge the gap between different resolutions, Jing~\etal~\cite{jing_2015} proposed learning two dictionaries for the high-resolution and low-resolution images, respectively, and developed patch-level mapping functions to overcome the resolution gap. 
Li~\etal~\cite{li_2015} proposed a joint multi-scale discriminant component analysis framework for aligning the feature spaces of images in different resolutions while keeping the discriminative information for distinguishing persons.
Wang~\etal~\cite{wang_2016} explored the characteristics of the scale-distance functions for scale-adaptive ReID.
Although they made considerable improvement in low-resolution ReID, they ignored effectively exploiting the abundant informative cues in high-resolution images and relied on hand-crafted features.

Recently, the super-resolution (SR) technique has been introduced for low-resolution ReID~\cite{dslr_market,wang_2018,zhang_2021_TIP,zhang_ICIP}.
Jiao~\etal~\cite{dslr_market} first proposed to utilize the super-resolution network to assist low-resolution ReID.
Since the conventional SR network focuses on image visual fidelity, learning a ReID-oriented SR network is crucial for SR-based low-resolution ReID models~\cite{dslr_market}.
Wang~\etal~\cite{wang_2018} developed a cascaded super-resolution network to enhance the LR images, making them as informative as the HR images.  
Chen~\etal~\cite{chen_2019} employed a decoder for super-resolution and introduced a discriminator to encourage the features to become resolution-invariant.
Mao~\etal~\cite{mao_2019} enforced the SR network to concentrate on the foreground of the images and proposed a two-steam network to handle LR and HR images for resolution invariant features. 
Li~\etal~\cite{li_2019} further enforced the super-resolved LR images to contain identity discriminative information. 
Cheng~\etal~\cite{cheng_2020} modeled the association between the super-resolution task and the person ReID task for regularizing the SR network so that the SR network can better bridge the gap between the LR images and the HR images for ReID. 
Huang~\etal~\cite{huang_2020} exploited the Generative Adversarial Network (GAN) to decomposite the identity features and other features.
More recently, Zheng~\etal~\cite{zheng_2022} jointly learned an SR network and two resolution-specific ReID extractors to leverage the specific information from both HR and LR images.

\vspace{0.6\baselineskip}\noindent\textbf{- Person ReID under clothing change.}
Since the clothes contain abundant appearance information, ReID models tend to focus on the cloth for discriminative cues.
As shown in existing works, these ReID models easily get trapped by the clothing change, and hard to associate the images under the clothing change~\cite{prcc,fsam,cg_chen_2022_tmm,cg_wu_2017,cg_zhang_2023,cg_gao_2022,cg_wang_2022_ACCV,iccv-2-cc,iccv-4-cc,ccfa, pami_cc_1}. 

Because the clothing and illumination change will inevitably affect the RGB images, depth-based ReID models~\cite{cg_li_2016,cg_wu_2017,PALA201969} are also proposed to utilize the shape, skeleton, and motion information for ReID.
Wu~\etal~\cite{cg_wu_2017} proposed to mine robust shape and skeleton information from depth information and bridged the gap between the RGB images and depth images. 
Fan~\etal~\cite{rfreid} exploited the information from radio frequency signals related to body size and shape for robust ReID. 
Han~\etal~\cite{cg_han_2022_ACCV} proposed utilizing the temporal and 3D shape information to resolve video-based clothing-change ReID.
Despite the progress, depth statistics and radio frequency signals are not widely available in real-world applications.

Instead of using 3D information, Zheng~\etal~\cite{dgnet} disentangled the appearance information (\eg, clothing color) and the structure information (\eg, body size) from images for ReID only using the person's identity. 
Li~\etal~\cite{casenet} enforced the network to focus on shape information against clothing change. 
Subsequently, several methods exploiting extra information such as body shape have been explored for the clothing change ReID.
Yu~\etal~\cite{cocas,cocas_plus} developed a two-stream network for biometrics and clothes features demanding the predefined clothes template for assisting image matching. 
Wan~\etal~\cite{cg_wan_2020} detected the face and extracted face features to overcome clothing change. 
Besides, abundant clothing variations such as synthetic data and augmented real data are introduced to improve models' robustness to clothing change~\cite{cele_light,cocas,casenet,cg_wan_2020,ltcc, cg_jia_2022}.
Moreover, Yang~\etal~\cite{prcc} proposed a learnable spatial polar transformation schema and a multi-stream model for extracting discriminative information from contour sketch images. 
To attenuate the information loss in sketch images, Chen~\etal~\cite{cg_chen_2022_tmm} exploited RGB and sketch images to learn shape-aware features. 
Xue~\etal~\cite{ltcc} extracted identity-discriminative body shape information via key points. 
Hong~\etal~\cite{fsam} learned fine-grained body shape knowledge via human masks.
Jin~\etal~\cite{cg_jin_2022} utilized the gait information to assist feature learning. 
Gu~\etal~\cite{cg_gu_2022} adversarially learned the cloth-agnostic features based on an auxiliary clothes classifier. 
Huang~\etal~\cite{rcsanet} improved the robustness of the model by learning the awareness of the clothing status. 

\vspace{0.6\baselineskip}\noindent\textbf{- Person ReID under occlusion.}
Occlusion ubiquitously exists in person images, resulting in only part of the person being visible~\cite{occ_zheng_2015}.
To overcome the occlusion, existing works mainly extracted the discriminative information from the visible part.
Zheng~\etal~\cite{occ_zheng_2015} designed an Ambiguity-sensitive Matching Classifier for patch-level matching and a Sliding Window Matching model to overcome the occlusion. Luo~\etal~\cite{occ_luo_2020} introduced the spatial transformer networks to bridge the gap between the occluded images and holistic images. 
He~\etal~\cite{occ_he_2018} proposed feature reconstruction against occlusion. 
Afterward, He~\etal~\cite{occ_he_2019} further developed a foreground probability generator to overcome the background interference. 
Sun~\etal~\cite{occ_sun_2019} proposed a Visibility-aware Part Model to focus the shared regions between images.
Based on the availability of large-scale data, some works~\cite{occ_jia_2022_tmm,occ_chen_2021_iccv,occ_yan_2021_iccv,occ_tan_2022_tnnls, occ_tan_2022} simulated the occlusion by data augmentation and enhanced the feature extractor by attention mechanism to learn discriminative features against occlusion.
Yan~\etal~\cite{occ_yan_2021_iccv} designed Compound Batch Erasing to create synthetic occlusions and enhanced the network with a Disentangled Non-Local module as well as the reconstructive pooling. 
Li~\etal~\cite{pat} employed the transformer for discriminative part-level features.

Moreover, some works~\cite{pvpm,pgfa,occ_zhang_2019,occ_yang_2021,iccv-8-part} introduced external cues for discovering visible parts and feature alignment between images.
Miao~\etal~\cite{pgfa} exploited poses to discover visible parts of the person in images and proposed matching visible parts of different images. 
Based on the information of key points, Wang~\etal~\cite{horeid} modeled a graph between different regions of the person for feature alignment.
Zheng~\etal~\cite{occ_zheng_kd} guided the network to focus on the human body by pose-guided knowledge distillation.
Wang~\etal~\cite{pfd} employed a transformer to extract discriminative part-level features and utilized the pose information to overcome occlusion interference.
Besides, researchers also investigated recovering the discriminative information from the occlusion part based on generative models~\cite{occ_recover_1,occ_recover_2} or spatial-temporal relations in videos~\cite{occ_recover_3,occ_recover_4}.

\vspace{0.6\baselineskip}\noindent\textbf{- Person ReID under lighting change and modality change.}
Learning to match images under various lighting conditions is crucial for ReID since there naturally exists lighting change in surveillance systems~\cite{wang2013camera,javed2005appearance,porikli2003inter,xiang2022rethinking,huang2019illumination}.
Kviatkovsky~\etal~\cite{ci_reid} mined the invariant and discriminative information in the color distribution for ReID.
Bhuiyan~\etal~\cite{btf_reid} learned the brightness transfer functions to resolve the illumination change.
Zeng~\etal~\cite{zeng2020illumination} disentangled the identity information to alleviate the interference from the illumination.

Since the images captured at day and night have dramatic illumination changes, Wu~\etal~\cite{wu2017rgb} proposed capturing the infrared (IR) images when the illumination is poor while capturing the RGB images in other cases, where IR images are always regarded as another \emph{modality} different from RGB images. 
Consequently, visible-infrared ReID (VI-ReID), which aims to match the RGB and IR images, becomes the core challenge for overcoming the lighting change~\cite{dai2018cross,ye2018visible,mm01,zhang2022modality,lin2022learning}. 

A stream of works aim at extracting modality-shared features~\cite{wei2020co,zhang2021learning,ye2020dynamic,ye2021dynamic,ye2018hierarchical,iccv-7-cm,ye2020bi,chen2022structure,huang2023exploring,mm01,zhao2022spatial,zheng2022visible,zhang2022visible_2,hu2022adversarial,iccv-3-cm}. 
Dai~\etal~\cite{dai2018cross} mitigated the modality discrepancy by adversarial learning.
Ye~\etal~\cite{ye2018visible} developed a two-stream network to extract features from different modalities.
For better feature learning, a bi-directional dual-constrained top-ranking loss is proposed~\cite{ye2018visible}.
Zhang~\etal~\cite{zhang2021learning} introduced an angle metric space and a cyclic projection network.
Liu~\etal~\cite{liu2020parameter} investigated the design of shared parameters in the two-stream network and developed a hetero-center triplet loss. 
Chen~\etal~\cite{chen2022structure} leveraged transformer architecture for robust features against pose variations and so forth.
Wang~\etal~\cite{wang2022optimal} utilized annotated RGB images to alleviate the demand for labeled IR images.
For better network architectures, Neural Architecture Search is explored~\cite{fu2021cm,chen2021neural}.
To better mitigate the modality gap, immediate modality is introduced~\cite{hat,li2020infrared}. 
Li~\etal~\cite{li2020infrared} learned an X modality with a modality gap constraint. 
Wei~\etal~\cite{wei2021syncretic} proposed combining the information of RGB and IR images to learn a syncretic modality to bridge the modality gap. 

Another stream of works exploits the modality-specific discriminative information~\cite{jsia,zhong2021grayscale,liu2022revisiting,li2022counterfactual}. Wang~\etal~\cite{wang2019learning} proposed generating cross-modality counterparts for images by GAN. 
Lu~\etal~\cite{lu2020cross} modeled the intra- and inter-modality affinity to propagate the modality shared and specific information among samples. 
Jiang~\etal~\cite{jiang2022cross} learned modality prototypes and adopted an encoder-decoder for modality compensation.
Zhang~\etal~\cite{zhang2022fmcnet} designed a feature-level modality compensation network to generate cross-modality features and employed a shared-specific feature fusion module to combine shared and specific modal information.
Li~\etal~\cite{li2022visible} proposed a memory network to complete missing modality information.
 
\vspace{0.6\baselineskip}\noindent\textbf{- Discussions.}
Until now, a large number of specific models have been developed for distinctive ReID scenes, but how to handle ReID under different scenes simultaneously is under exploration.
Differently, our work proposes using a single versatile model to overcome this challenge, which first independently learns diverse knowledge for adapting distinctive scenes by leveraging learnable parameters called scene-specific prompts.
The diverse knowledge from the specific prompts is then further adaptively aggregated into versatile prompts to enable the ReID model to overcome different scenes.

\subsection{Multi-task and Multi-scene Learning}
\label{s.2.2}
Recently, multi-task learning~\cite{mt1,mt2,mt3,mt4,mt5,hardparashare} has attracted increasing attention in the computer vision (CV) community, which trains a unified model to handle various CV tasks, \eg, classification, detection, and generation.
A typical technology in multi-task learning is modeling the task-shared and task-specific knowledge separately.
For example, Baxter~\etal~\cite{hardparashare} and Liang~\etal~\cite{mt5} utilized task-shared and task-specific layers to learn the above knowledge.
Lu~\etal~\cite{mt1} introduced a task-shared backbone network and task-specific prompts to capture task-shared and task-specific cues, respectively.

Multi-scene learning, a related field of multi-task learning, mainly focuses on developing a unified model capable of carrying out a specific task in various scenes, such as outdoor, indoor, and so on. 
Yoli~\etal~\cite{ms1} explored a multi-scene pose regressor by applying a shared-weight transformer backbone and scene-specific queries for modeling scene-shared and scene-specific knowledge in the absolute pose regression task. 

In the above modelings, there is an explicit difference among different tasks/scenes, and the previous modeling is necessary to assign a specific indicator to each task/scene so that models can react diversely according to the task/scene they are currently handling.
Unfortunately, the differences among ReID scenes in practice are ambiguous, and therefore it is hard to assign the scene indicator in the inference stage.
To address this problem, we propose a two-stage ReID framework that first leverages scene labels to capture abundant knowledge from different scenes and then aggregates them to a final model that does not require scene labels.

\section{A Versatile Person ReID Framework}
\label{s.3}
\begin{figure*}[t]
  \centering
  \includegraphics[width=1\linewidth]{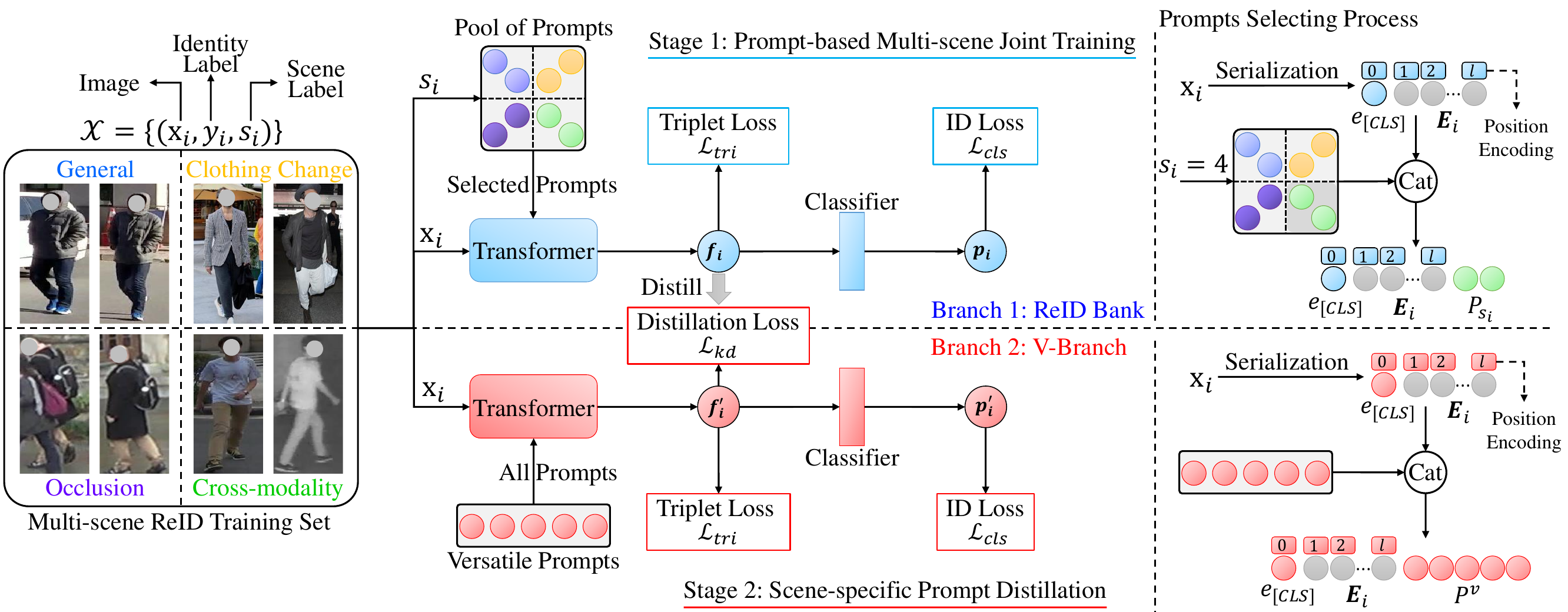}
  \caption{An overview of the VersReID framework. Our VersReID handles the multi-scene ReID by capturing different types of knowledge in a model called ReID Bank (the first stage) under the indication of scene labels and then unifying the knowledge into the V-Branch model (the second stage) to get rid of scene labels. The top side illustrates the first learning stage, namely the prompt-based multi-scene joint training. Based on scene labels, we associate distinctive prompts with images under different scenes, learning to encode scene-specific knowledge in prompts. By integrating the scene-specific prompts with a scene-shared backbone that contains scene-invariant knowledge, we build a multi-scene ReID Bank. 
  Although with abundant knowledge, the usability of ReID Bank is limited since it requires scene labels to select scene-specific prompts.
  We solve this problem in the second stage, namely the scene-specific prompt distillation.
  As illustrated in the bottom part, we distillate the ReID Bank's knowledge to a versatile branch with versatile prompts called V-Branch. Through the distillation, the knowledge from different scenes is unified into the V-Branch model. Hence, the V-Branch model can handle multiple ReID scenes simultaneously without using the scene labels. The ``Cat'' in the figure is the concatenate operation, and $l$ indicates the length of the image tokens.}
  \label{f.2}
\end{figure*}

We aim to construct a versatile ReID framework, \textbf{VersReID}.
The VersReID framework is a prompt-based twin modeling framework that integrates the ``Multi-scene ReID Bank'' (ReID Bank) and the ``Versatile ReID Branch'' (V-Branch), which are learned in two stages as illustrated in Figure~\ref{f.2}.

In the first training stage for learning the ReID Bank, we employ several groups of scene-specific prompts for adaptively modeling different scene-specific knowledge under the indication from scene labels.
Learning scene-specific knowledge that describes distinctive characteristics of different scenes is important in the multi-scene ReID.
For instance, although each person's attire is a crucial discriminative characteristic in the general ReID scene, such a pattern might not be effective in clothing change ReID scene~\cite{prcc,fsam,cg_chen_2022_tmm}.
Besides, a scene-shared backbone is employed for scene-invariant knowledge.
Benefiting the high flexibility of prompt modeling, we integrate the scene-specific prompts with the scene-share backbone to form the ReID Bank.  
Driven by scene labels, the ReID Bank learns abundant knowledge for handling various scenes.

However, the scene labels are not always available in practice. 
Therefore, in the second training stage, namely the scene-specific prompt distillation, 
different knowledge captured by scene-specific prompts in the ReID Bank is adaptively unified to a V-Branch model with versatile prompts for effectively handling various scenes.

\subsection{Prompt-based Multi-scene Joint Training} 
\label{s.3.1}
Suppose that a training set from multiple ReID scenes is available, denoted as $\mathcal{X}=\left\{ (\mathbf{x}_i, y_i, s_i) \right\}$, where $\mathbf{x}_i$ and $y_i$ indicate the $i$-th image and the corresponding identity, respectively. $s_i$ is the scene label denoting which scene $\mathbf{x}_i$ is from, \eg, occlusion, low-resolution, and so forth.

We aim to capture different types of knowledge from different scenes using scene labels and form a ``Multi-scene ReID Bank'' (ReID Bank). 
For this purpose, we introduce a prompt-based multi-scene joint training procedure to learn the ReID Bank, where the scene-specific prompts adaptively encode scene-specific knowledge and a fundamental backbone captures scene-invariant knowledge. 

\vspace{0.6\baselineskip}\noindent\textbf{- Prompt modeling for scene-specific knowledge.} 
We learn a fixed number of learnable prompts under each ReID scene, and the learnable prompts of a scene are supposed to contain corresponding scene-specific knowledge.
Under the multi-scene conditions, a pool of prompts is denoted as $P = \{P_j \mid j=1,2, \cdots, S\}$, where $S$ is the number of total scenes and $P_j$ represents a group of $N$ prompts for the $j$-th scene.
In the top part of Figure~\ref{f.2}, we show an example of $S=4$ and $N=2$.

Based on the pool of prompts, for each training image $\mathbf{x}_i$, we are able to select the scene-specific prompts according to its scene label $s_i$ and feed the image with the selected prompts into the scene-shared vision transformer (ViT)~\cite{vit}. 
For the feature extraction, the input image is first converted into a sequence of tokens following the standard serialization strategy in the ViT. 
The operations include the patch embedding operation, appending the class token $e_{[\mathrm{CLS}]}$, and adding learnable positional encodings.
We denote the sequence of tokens after the above operations as $\{e_{[\mathrm{CLS}]};\boldsymbol{\mathrm{E}_i}\}$, where $\boldsymbol{\mathrm{E}_i}$ is the sequence of the input image. 
We then concatenate the selected prompts to the end of the image sequence. 
Note that the position encodings are not applied to scene-specific prompts because there is no spatial association between these prompts. 
Finally, we obtain the complete input sequence for the transformer blocks, denoted as $\{e_{[\mathrm{CLS}]};\boldsymbol{\mathrm{E}_i};P_{s_i}\}$. 
The above modeling is illustrated in the top-right corner of Figure~\ref{f.2}, where we take $s_i=4$ for example.
After several transformer blocks process the complete input sequence, we separate the class token $e_{[\mathrm{CLS}]}$ from the sequence as the global feature representation of the input image, denoted as $\boldsymbol{f}$, and adopt an identity classifier on it. The predicted probability is denoted as $\boldsymbol{p}$, as shown in the top-middle part of Figure~\ref{f.2}.

For capturing abundant knowledge from data, we employ the triplet loss $\mathcal{L}_{tri}$ with hard-sample mining~\cite{triplet_loss} for the global feature representation $\boldsymbol{f}$ and the classification loss $\mathcal{L}_{cls}$ implemented as the cross-entropy between the one-hot label $y$ and the predicted probability $\boldsymbol{p}$, to train the ReID Bank. 
The entire objective function is formulated as:
\begin{equation}
\label{eq.1}
   \mathcal{L} = \mathcal{L}_{tri} + \mathcal{L}_{cls}.
\end{equation}
By training the ReID Bank via coupling scene-specific prompts with the corresponding scene, the prompts can help the ReID Bank adaptively focus on learning the specific patterns appropriate to the scene they are associated with.

\vspace{0.6\baselineskip}\noindent\textbf{- Discussions.} 
Unlike the hard parameter-sharing mode~\cite{hardparashare} that utilizes scene-specific layers or blocks to extract the specific knowledge among scenes, our prompt-based modeling is more flexible since we do not need to 
 elaborately design which layers should be shared among scenes and which layers should be scene-specific.
We experimentally demonstrate that modeling the scene-specific knowledge by learnable prompts is more elegant and effective than the hard parameter-sharing mode in the multi-scene ReID conditions. 
Details can be found in the Supplementary.

\subsection{Scene-specific Prompt Distillation} 
\label{s.3.2}

\begin{figure*}[t]
  \centering
  \includegraphics[width=1\linewidth]{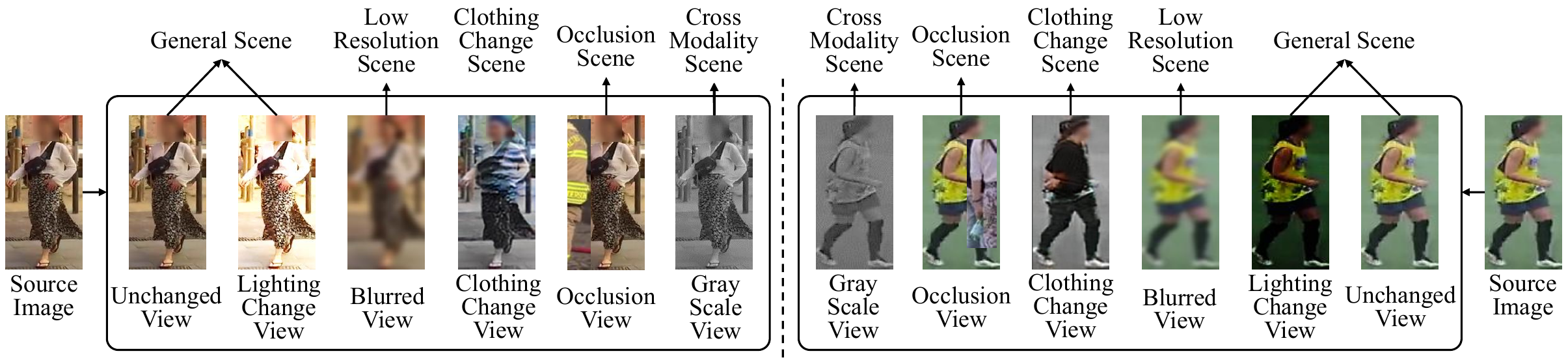}
  \caption{The illustration of the proposed Multi-scene Prioris Data Augmentation (MPDA), which generates multiple augmented views for each source image to simulate the cross-scene variations. As a plug-and-play learning strategy, MPDA is applicable to typical self-supervised contrastive learning methods. The geometric data augmentation (random cropping, resizing, and horizontal flipping) is omitted for simplicity.} 
  \label{f.3}
\end{figure*}

In our formulation, we assume that the scene label of the image is known during inference and use the global feature representation $\boldsymbol{f}$ to retrieve people. 
However, such an assumption of scene label does not always hold in actual scenes, where it is hard to determine which ReID scene the sample should belong to.
Besides, the sample may contain the characteristics of multiple scenes at the same time, \eg, low-resolution and occlusion, in which case using knowledge from merely one scene could be sub-optimal.

To solve the above problems, we explore unifying the knowledge in scene-specific prompts to overcome the multi-scene ReID without scene labels. 
Our idea is to distill the knowledge of the multi-scene ReID Bank into a versatile branch, namely the \textbf{V-Branch}, which is specified by several versatile prompts.
We call this distillation stage as ``scene-specific prompt distillation'' and illustrate it in the bottom part of Figure~\ref{f.2}.
Through the distillation procedure, the scene-shared backbone network with versatile prompts contains sufficient knowledge for resolving ReID across scenes, and the learned V-Branch model does not require the scene label during inference any longer.

\vspace{0.6\baselineskip}\noindent\textbf{- V-Branch with versatile prompts.} 
The V-Branch model has the same backbone architecture as its teacher model ReID Bank without sharing weights. 
The main difference between the V-Branch and the ReID Bank is in the process of selecting prompts.
Specifically, we randomly initialize $M$ versatile prompts $P^v$ to augment the scene-shared backbone of the V-Branch.
Different from the ReID Bank that selects prompts according to the specific scene label of the sample, the V-Branch applies all versatile prompts for arbitrary images among every scene, \ie, the input of transformer blocks in the V-Branch model is denoted as $\{e_{[\mathrm{CLS}]};\boldsymbol{\mathrm{E}_i};P^v\}$, as shown by the bottom-right corner of Figure~\ref{f.2}.

\vspace{0.6\baselineskip}\noindent\textbf{- Distillation procedure.} 
The ReID Bank is frozen during the distillation procedure, and its outputs remain the same as described in the previous subsection.
We denote the global feature representation and the predicted probability of the V-Branch as $\boldsymbol{f'}$ and $\boldsymbol{p'}$, respectively. 
We leverage the global feature representation $\boldsymbol{f}$ output by the ReID Bank that contains scene-invariant and scene-specific information as the supervision to guide the global output $\boldsymbol{f'}$ of the V-Branch. 
The distillation process is carried out via a knowledge distillation~\cite{kd} loss $\mathcal{L}_{kd}$. 
\major{Since ReID is a ranking problem and the relations between pairwise images are the keys, we instantiate $\mathcal{L}_{kd}$ as the relational knowledge distillation loss~\cite{rkd}.
Formally, given two images $\mathbf{x}_i$ and $\mathbf{x}_j$, we utilize the ReID Bank to extract the features denoted as $\boldsymbol{f}_i$ and $\boldsymbol{f}_j$ using the scene-specific prompts according to their scene labels $s_i$ and $s_j$.
Similarly, denote the features from V-Branch as $\boldsymbol{f'}_i$ and $\boldsymbol{f'}_j$.
Notably, as mentioned above, the V-Branch does not require scene labels to extract features.
The relational knowledge distillation $\mathcal{L}_{kd}$ is formulated as:
\begin{equation}
\mathcal{L}_{kd} = \sum_{i,j}  \Vert \boldsymbol{d}'_{i,j} - \boldsymbol{d}_{i,j} \Vert_2^2,
\end{equation}
where $\boldsymbol{d}_{i,j} = \Vert \boldsymbol{f}_i -\boldsymbol{f}_j \Vert_2^2, \boldsymbol{d}'_{i,j} =  \Vert  \boldsymbol{f}'_i -\boldsymbol{f}'_j \Vert_2^2$.
}

Except for the $\mathcal{L}_{kd}$, V-Branch is likewise trained using the previously described triplet loss $\mathcal{L}_{tri}$ and classification loss $\mathcal{L}_{cls}$. 
The objective function for scene-specific prompt distillation is formulated as follows: 
\begin{equation}
\label{eq.2}
   \mathcal{L}_{V-Branch} = \mathcal{L}_{tri} + \mathcal{L}_{cls} + \alpha\mathcal{L}_{kd},
\end{equation}
where $\alpha$ is a non-negative weight to balance the supervision from the identity labels and the ReID Bank. 
By optimizing Equation~\ref{eq.2}, the V-Branch is supposed to unify the scene-invariant and scene-specific knowledge from the ReID Bank to function across multiple scenes with versatile prompts.

\vspace{0.6\baselineskip}\noindent\textbf{- Discussions: Utilization of scene priors.}
\major{\yxa{Introducing the scene priors can help the model handle ReID under various scenes.
In our work, the scene priors actually exist in the ReID Bank.
Based on the scene labels, we utilize a group of scene-specific prompts to model the scene-specific knowledge in ReID Bank.
When learning V-Branch from the ReID Bank, we use scene labels to guide the ReID Bank to process images with different scene-specific prompts.
Hence, the supervision provided by ReID Bank contains scene priors and is perceived by the V-Branch during the distillation from the ReID Bank to the V-Branch, although label information is not required for the V-Branch in inference.}}

\major{We do not make more scene-specific modifications to the V-Branch, because we aim to develop a versatile ReID model adaptive to multiple scenes during inference, where scene labels are always NOT available in inference.
We also investigate learning scene-specific models (fully being aware of scenes) and conduct model ensemble for multi-scene ReID, but it exhibits inferior performance (please refer to Supplementary).
Beyond our method, deliberately exploiting more scene priors may bring further benefits to the V-Branch model, which will be left for future works.}

\section{Self-supervised learning with Multi-scene Prioris}
\label{s.4}
Due to the drastic cross-scene variations, it is challenging to drive a deep model to automatically learn a feature embedding applicable to various scenes, especially considering the data-hungry characteristic of deep models.
Since the labeled data is costly and difficult to scale up, we propose to facilitate learning our VersReID model using tremendous unlabeled data in a self-supervised manner. 
Recently, self-supervised learning (SSL) on large-scale unlabeled person datasets~\cite{lup,transreid_ssl} shows promising results in general ReID, especially contrastive learning~\cite{dino,mocov1,mocov3,byol,simclr} that helps models learn to discriminate different instances by maximizing the similarity between different augmented views of the same source image. 

Benefiting from the advanced contrastive learning methods, we make a further step and introduce multi-scene prioris for the self-supervised multi-scene ReID, which is expected to exploit a large amount of unlabeled data to help the model better adapt to various downstream ReID scenes.
Our core intuition is to simulate the specific ReID scenes by considering their properties and then overcome the variations in a self-supervised learning fashion. 
Specifically, we apply a Multi-scene Prioris Data Augmentation (MPDA) strategy to generate images with various characteristics according to different ReID scenes.
Without losing generality, in this work, we mainly focus on five mainstream ReID scenes commonly occurring in practice, including general, low-resolution, clothing change, occlusion, and cross-modality scenes. 

Note that the above scenes are not absolutely independent of each other, \eg, occlusion and low-resolution may appear in any scenes inadvertently.
However, as other scenes do not customize these properties, \ie, artificially selecting occluded or low-resolution images, these properties are sparse in those scenes.

\subsection{Multi-scene Prioris Data Augmentation} 
\label{s.4.1}
To simulate various ReID scenes, we first associate each ReID scene with an augmented view.
Specifically, as shown in Figure~\ref{f.3}, we design six augmented views for a source image according to the five related ReID scenes, including an unchanged view and a lighting change view for the general scene, a blurred view for the low-resolution scene, an occlusion view for the occlusion scene, a clothing change view for the clothing change scene, and a gray-scale view for the cross-modality scene.

\vspace{0.6\baselineskip}\noindent\textbf{- The unchanged view and lighting change view for the general scene.} 
The unchanged view is the source image itself, which retains most of the semantic information of the identity.
Due to various illumination, lighting change widely exists in the general scene.
The lighting change view is generated by applying the brightness and contrast ratio jitter to the source image for simulating lighting variations.

\vspace{0.6\baselineskip}\noindent\textbf{- The blurred view for the low-resolution scene.}
To simulate the low-resolution scene, we utilize a Gaussian blur operation to augment the source image for generating blurred views containing resolution change. 

\vspace{0.6\baselineskip}\noindent\textbf{- The clothing change view for the clothing change scene.}
To simulate the clothing change person ReID scene, we adopt a pre-trained DGNet~\cite{dgnet} to change the clothes of the person on the source image to another one. 
The clothing change operation changes both the color and texture of the source clothing compared to the color jitter that only changes the color, which greatly increases the data diversity.

\vspace{0.6\baselineskip}\noindent\textbf{- The occlusion view for the occlusion scene.}
We perform a crop-and-paste strategy on the source image to simulate the ReID task under the occlusion scene. 
Specifically, we randomly crop a region from another image and then paste this region to the candidate region on the source image to form an occlusion view. 

\vspace{0.6\baselineskip}\noindent\textbf{- The gray-scale view for the cross-modality scene.}
Cross-modality ReID aims to handle person ReID under invisible light conditions, \ie, night, which consists of infrared (IR) images and RGB images. 
However, it is challenging to collect IR images or synthesize virtual IR images. 
Fortunately, IR images look similar to gray-scale images to a certain extent. 
Thus, we apply a gray-scale data view to simulate the IR image for considering the cross-modality scene. 

\vspace{0.6\baselineskip}\noindent\textbf{- Discussions.}
The proposed MPDA is a plug-and-play augmentation strategy for introducing multi-scene variations in self-supervised learning and thus can benefit from the advanced contrastive learning methods.
Recall that we capture the knowledge to the ReID Bank before we unify the knowledge from different scenes to the V-Branch.
Consequently, if we integrate MPDA with contrastive SSL methods that aim to learn invariant representations against data augmentations, we can enrich the knowledge of ReID Bank by self-supervised learning the backbone of the ReID Bank on unlabeled data, which ultimately benefits the V-Branch through knowledge distillation.

\section{Experiments}
\label{s.5}
We conduct experiments to show the effectiveness of our VersReID. 
Firstly, we introduce the datasets and evaluation metrics in Section~\ref{s.5.1}.
Then, in Section~\ref{s.5.2}, we compare the proposed VersReID with other advanced ReID methods.
In Section~\ref{s.5.3}, we perform extensive ablation studies to evaluate the contribution of each design in our VersReID.
Finally, we visualize and discuss the self-attention maps of our VersReID for better understanding in Section~\ref{s.5.4}.

We focus on five ReID scenes in the experiments, including the general, low-resolution, clothing change, occlusion, and cross-modality scenes, as mentioned in Section~\ref{s.4}.

\subsection{Datasets and Evaluation Metrics}
\label{s.5.1}
\vspace{0.6\baselineskip}\noindent\textbf{- LUPerson.}
LUPerson (short for LUP)~\cite{lup} is a large-scale unlabeled person dataset with $\sim$4.18M images for self-supervised learning in ReID.
Since the LUP was collected from the Internet, to protect personal privacy, we use face detection\footnote{FaceMaskDetection \\ https://github.com/AIZOOTech/FaceMaskDetection} and obfuscation~\cite{face_blur} to blur the person face in the dataset.
The LUP dataset with additional face blur processing is denoted as \textbf{LUP-B}.

\vspace{0.6\baselineskip}\noindent\textbf{- Datasets of general ReID.} 
The downstream General ReID scene contains Market-1501~\cite{market} and MSMT17~\cite{msmt}. 
Market-1501 contains 12,936 training images, 3,368 query images, and 15,913 gallery images from 1,501 identities under 6 cameras.
The MSMT17 dataset is collected from 15 cameras, containing 32,621 training images, 11,659 query images, and 82,161 gallery images from 4,101 identities. 
The above two datasets are general ReID datasets without artificial selection of occlusion, low-resolution, clothing change, \etc.

\vspace{0.6\baselineskip}\noindent\textbf{- Datasets of low-resolution ReID.} 
The low-resolution ReID scene contains the MLR-CUHK03~\cite{dslr_market} dataset, a variant of CUHK03~\cite{cuhk} that aims to simulate a real low-resolution scene. 
Thus, it applies down-sampling to query images for creating low-resolution queries, and its gallery set is the same as the original CUHK03. 
Moreover, half of the training images of MLR-CUHK03 are also the down-sampling version from CUHK03.
In summary, the MLR-CUHK03 dataset contains 965 low-resolution query images, 977 high-resolution gallery images, and 2,6251 training images.

\vspace{0.6\baselineskip}\noindent\textbf{- Datasets of clothing change ReID.} 
To form the clothing change ReID scene, we leverage the Celeb-ReID dataset~\cite{cele} and PRCC dataset~\cite{prcc}. 
Celeb-ReID dataset contains 20,208 training images, 2,972 query images, and 11,006 gallery images from 1,052 identities, and over 70\% of the samples show different clothes. 
PRCC dataset~\cite{prcc} is collected under 3 cameras, \ie, cameras A, B, and C. 
The clothing of the same identity under cameras A and B is the same but different from that of camera C.
The training set of PRCC consists of 17,896 person images of 150 identities.
We follow the cross-clothes evaluation protocol, where 3,384 query images and 3,543 gallery images of 71 identities are used for evaluation.

\vspace{0.6\baselineskip}\noindent\textbf{- Datasets of occlusion ReID.} 
The Occluded-Duke dataset \cite{pgfa} (short for Occ-Duke) is selected to represent the occlusion ReID scene, where the images are manually selected to simulate the occlusion scene.
It contains 2,331 identities over 8 cameras, including 15,618 training images, 2,210 occluded query images, and 17,661 gallery images.

\vspace{0.6\baselineskip}\noindent\textbf{- Datasets of cross-modality ReID.} 
We evaluate the cross-modality ReID task on the SYSU-mm01~\cite{mm01} dataset, which is collected by 6 cameras, including 4 visible (RGB) and
2 infrared~(IR) cameras. 
The training set of SYSU-mm01 consists of 22,258 RGB and 11,909 IR images of 395 identities.
The testing set contains 96 identities, including 3,803 IR query images and 6,775 RGB gallery images. 
We conduct evaluations following the all-search mode, where the gallery set contains all the RGB images captured from all images from the 4 visible cameras. 

\vspace{0.6\baselineskip}\noindent\textbf{- Multi-scene ReID training set.} 
We combine the training sets of all the downstream ReID datasets mentioned before to form a multi-scene ReID training set, including 7 datasets, Market-1501, MSMT17, MLR-CUHK03, Celeb-ReID, PRCC, Occ-Duke, and SYSU-mm01.
The built training set contains 5 ReID scenes including the general, low-resolution, clothing-change, occlusion, and cross-modality scenes, and it includes 5,038 identities and 159,697 training images. 

\vspace{0.6\baselineskip}\noindent\textbf{- Multi-scene ReID joint testing set.} 
Similar to the multi-scene ReID training set, we combine the query and gallery sets of all the downstream ReID scenes to form a multi-scene ReID joint testing set, which is more practical and challenging than a specific ReID testing set as various scenes occur simultaneously. 
The query set of the joint testing set consists of the counterparts of Market-1501, MSMT17, MLR-CUHK03, Celeb-ReID, PRCC, Occ-Duke, and SYSU-mm01, where there are 7 query sets, including 28,361 query images of 5,016 identities.
The above datasets' gallery sets containing 138,036 images of 5,608 identities contribute to the gallery set of the joint testing set. 

\vspace{0.6\baselineskip}\noindent\textbf{- Evaluation metrics.}
Following the previous works, we adopt Cumulative Matching Characteristics of Rank-1 Accuracy (short for R-1) and mean Average Precision (short for mAP) as the evaluation metrics. 
No post-processing operations are used in all the experiments, \eg, re-ranking~\cite{rk}.

\subsection{Implementation Details}
\label{s.5.2.1}
We apply the proposed MPDA strategy to DINO~\cite{dino} and learn the ViT-Base (ViT-B)~\cite{vit} with 50\% training data of the LUP-B dataset for 100 epochs following~\cite{transreid_ssl} in a self-supervised manner.
The method is denoted as DINO + MPDA.
We train the model on 4 GPUs using the AdamW optimizer~\cite{adamw} with a batch size of 512. 
The data augmentation is introduced in Section~\ref{s.4.1}.
\major{In each training iteration, we generate two different views of each source image following previous work in self-supervised learning~\cite{dino, mocov3}.
Note that the MPDA aims to introduce multi-scene variations and is designed for self-supervised contrastive learning on unlabeled person datasets.
Since the multi-scene training set is labeled and each identity has multiple samples in different scenes, we do not use MPDA in the multi-scene training set to generate fake samples to avoid disturbance.
\textbf{All the comparison methods in multi-scene setup are pre-trained on LUP-B dataset using MPDA + DINO.
}}

\vspace{0.6\baselineskip}\noindent\textbf{- Prompt-based multi-scene joint training.} 
In the first stage, we train ReID Bank on the built multi-scene ReID training set mentioned in Section~\ref{s.5.1}, and \textbf{do not} fine-tune the model on any specific dataset.
We adopt a ViT-B model learned from the DINO + MPDA as the scene-shared transformer backbone in ReID Bank.
No camera information or local features are used. 

The number of scene-specific prompts for one scene $N$ is 2, and there are 10 prompts from 5 scenes in the prompt pool.
\major{By default, all prompts are initialized randomly. We also attempt to adopt a pre-trained CLIP~\cite{clip} text encoder to extract the textual embedding for the prompt initialization but do not gain improvements.  Details are in Supplementary.} 
Moreover, the model is trained on 1 GPU for 120 epochs using an SGD with momentum optimizer~\cite{sgd}.
The mini-batch size is set to 128 consisting of 32 persons (4 images per person).
A cosine decaying scheduler is used with a base learning rate of 4e-4. 
We resize the input images to $384 \times 128$ and employ Equation~\ref{eq.1} as the loss function. 

\vspace{0.6\baselineskip}\noindent\textbf{- Scene-specific prompt distillation.}
The V-Branch shares a similar architecture with the ReID Bank except that we replace the scene-specific prompts with $M$ randomly initialized versatile prompts and we set $M=5$ in our experiments. 
We use the scene-shared transformer backbone of the ReID Bank to initialize the transformer backbone of the V-Branch model.
The ReID Bank is frozen during the distillation stage.
Equation~\ref{eq.2} is employed as the loss function with $\alpha=1$. 
Other settings including hyper-parameters are the same as the multi-scene joint training stage.
\major{Please refer to our Supplementary for experiment details.}

\subsection{Evaluations of VersReID}
\label{s.5.2}
We divide the methods into two categories: 1)~single-scene models that are trained on a specific dataset, and 2) multi-scene models that are trained on the multi-scene dataset.

Since our VersReID aims to resolve different scenes simultaneously, the VersReID is trained on the multi-scene dataset and is directly applied to various scenes \textbf{\major{without fine-tuning}} on the specific single-scene dataset. 
Because we are the first to consider addressing the multi-scene ReID problem, except for VersReID, we train six typical ReID models on the multi-scene dataset for comparisons, \major{including PCB~\cite{pcb} and TransReID~\cite{transreid} designed for the general ReID, AIM~\cite{yang2023good} and ReIDCaps~\cite{cele} for the clothing change ReID, and PFD~\cite{pfd} as well as ETNDNET~\cite{etndnet} for the occlusion ReID. 
We denote the above methods as the ``multi-scene methods''.
\major{For a fair comparison, they are pre-trained on the LUP-B dataset using DINO + MPDA.}
For other methods that train models on the specific single-scene dataset, we denote them as ``single-scene methods''.}

\subsubsection{Comparisons on Single Scene}
\label{s.5.2.2}

\renewcommand{\arraystretch}{1.05}
\begin{table}[t]
  \caption{Comparison with other methods on the general scene (Market-1501 and MSMT17). ``*'' means using an overlapping patch embedding layer.}
  \label{t.1}
  \centering
  \resizebox{0.95\linewidth}{!}{
  \begin{tabular}{c | l | c c | c c }
    \toprule
    Training & \makecell[c]{\multirow{2}{*}{Method}} & \multicolumn{2}{c|}{Market-1501} & \multicolumn{2}{c}{MSMT17} \\
    \cline{3-6}
    Set &  & R-1 & mAP & R-1 & mAP \\
    \hline
    & PCB~\cite{pcb} & 93.8 & 81.6 & 68.2 & 40.4 \\
    & ViT-B~\cite{vit} & 94.0 & 87.6 & 82.8 & 63.6 \\
    & IANet~\cite{ianet} & 94.4 & 83.1 & 75.5 & 46.8 \\
    & BOT~\cite{bot} & 94.5 & 85.9 & 74.1 & 50.2 \\
    & DG-Net~\cite{dgnet} & 94.8 & 86.0 & 77.2 & 52.3 \\
    & TransReid~\cite{transreid} & 95.0 & 88.8 & 84.6 & 66.6 \\
    & MGN~\cite{mgn} & 95.1 & 87.5 & 85.1 & 63.7 \\
    Single & TransReid*~\cite{transreid} & 95.2 & 89.5 & 86.2 & 69.4 \\
    Scene & AAformer~\cite{aaformer} & 95.4 & 87.7 & 83.6 & 63.2 \\
    & AGW~\cite{agw} & 95.5 & 89.5 & 81.2 & 59.7 \\
    & FlipReID~\cite{flipreid} & 95.5 & 89.6 & 85.6 & 68.0 \\
    & PFD*~\cite{pfd} & 95.5 & 89.7 & 83.8 & 64.4 \\
    & LDS~\cite{lds} & 95.8 & 90.4 & 86.5 & 67.2 \\
    & DiP*~\cite{dip} & 95.8 & 90.3 & 87.3 & 71.8 \\
    & MPN~\cite{mpn} & 96.4 & 90.1 & 83.5 & 62.7 \\
    \hline \hline
    \specialrule{0em}{0.5pt}{0.5pt}
    & PCB~\cite{pcb} & 93.3 & 83.1 & 69.9 & 48.5 \\
    & \major{AIM}(R50)~\cite{yang2023good} & \major{91.9} & \major{81.6} & \major{72.6} & \major{46.4} \\
    & \major{ETNDNET}(R50)~\cite{etndnet} & \major{93.7} & \major{84.3} & \major{75.7} & \major{50.5} \\
    & \major{AIM}(ViT-B)~\cite{yang2023good} & \major{94.4} & \major{88.5} & \major{83.5} & \major{65.3} \\
    Multiple & \major{ETNDNET}(ViT-B)~\cite{etndnet} & \major{92.7} & \major{81.6} & \major{63.8} & \major{41.5} \\
    Scenes & \major{PFD}~\cite{pfd} & \major{94.3} & \major{86.8} & \major{82.0} & \major{63.3} \\
    & TransReid~\cite{transreid} & 95.0 & 89.7 & 85.8 & 69.4 \\
    & \major{ReIDCaps}~\cite{cele} & \major{95.6} & \major{90.6} & \major{86.6} & \major{69.6} \\
    \specialrule{0em}{0.5pt}{0.5pt}
    \cline{2-6}
    \specialrule{0em}{0.5pt}{0.5pt}
    & \textbf{VersReID (ours)} & 96.9 & 93.0 & 88.5 & 73.6 \\
    & \textbf{VersReID* (ours)} & 96.8 & 93.2 & 88.8 & 74.2 \\
    \bottomrule
  \end{tabular}
  }
\end{table}

\begin{table}[t]
  \caption{Comparison with other methods on the low-resolution ReID scene (MLR-CUHK03). ``*'': using an overlapping patch embedding layer.}
  \label{t.1.5}
  \centering
  \resizebox{0.8\linewidth}{!}{
  \begin{tabular}{c | l | c c }
    \toprule
    Training & \makecell[c]{\multirow{2}{*}{Method}} & \multicolumn{2}{c}{MLR-CUHK03} \\
    \cline{3-4}
    Set &  & R-1 & mAP \\
    \hline
    & SING~\cite{dslr_market} & 67.7 & - \\
    & PCB~\cite{pcb} & 75.3 & - \\
    & CAD-Net~\cite{li_2019} & 82.1 & - \\
    & OSNet~\cite{OSNET} & 83.8 & 85.4 \\
    \multirow{2}{*}{Single} & ABD-Net~\cite{ABDNet} & 84.3 & 85.9 \\
    \multirow{2}{*}{Scene} & PRI~\cite{han2020prediction} & 85.2 & - \\
    & INTACT~\cite{cheng_2020} & 86.4 & - \\
    & AGW~\cite{agw} & 87.3 & 88.4 \\
    & JBIM~\cite{zheng_2022} & 88.7 & 90.3 \\
    & PS-HRNet~\cite{zhang_2021_TIP} & 92.6 & - \\
    \hline \hline
    \specialrule{0em}{0.5pt}{0.5pt}
    & PCB~\cite{pcb} & 87.3 & 90.4 \\
    & TransReid~\cite{transreid} & 92.9 & 93.6 \\
     & \major{AIM}(R50)~\cite{yang2023good} & \major{89.3} & \major{89.9} \\
    & \major{ETNDNET}(R50)~\cite{etndnet} & \major{90.2} & \major{89.8} \\
    Multiple & \major{AIM}(ViT-B)~\cite{yang2023good} & \major{92.5} & \major{91.9}  \\
    Scenes & \major{ETNDNET}(ViT-B)~\cite{etndnet} & \major{90.8} & \major{89.7} \\
    & \major{ReIDCaps}~\cite{cele} & \major{93.3} & \major{94.6} \\
    & \major{PFD}~\cite{pfd} & \major{94.1} & \major{95.5} \\
    \specialrule{0em}{0.5pt}{0.5pt}
    \cline{2-4}
    \specialrule{0em}{0.5pt}{0.5pt}
    & \textbf{VersReID (ours)} & 97.1 & 98.2 \\
    & \textbf{VersReID* (ours)} & 97.5 & 98.4 \\
    \bottomrule
  \end{tabular}
  }
\end{table}

The experimental results on general, low-resolution, clothing change, occlusion, and cross-modality scenes are reported in Table~\ref{t.1}, Table~\ref{t.1.5}, Table~\ref{t.2}, and Table~\ref{t.3} respectively.
Unless otherwise specified, the VersReID in the table denotes the performance of the V-Branch model, which does not require scene labels in inference.

\vspace{0.6\baselineskip}\noindent\textbf{- General ReID scene.}
As shown in Table~\ref{t.1}, our VersReID achieves 96.9\%/93.0\% and 88.5\%/73.6\% in terms of the Rank-1 accuracy/mAP on Market-1501~\cite{market} and MSMT17~\cite{msmt}, respectively.
Compared to the second-best multi-scene model ReIDCaps~\cite{cele}, the VersReID shows non-trivial improvements, which demonstrates the effectiveness of our prompt-based modeling method. 
For example, VersReID outperforms ReIDCaps by +1.3\% and +1.9\% at the Rank-1 accuracy on Market-1501 and MSMT17, respectively. 
Furthermore, the performance of VersReID is comparable to or even better than most of the single-scene methods. 
These results indicate that VersReID can handle general ReID scene effectively.

\vspace{0.6\baselineskip}\noindent\textbf{- Low-resolution ReID scene.}
The results on the low-resolution ReID dataset MLR-CUHK03~\cite{dslr_market} are shown in Table~\ref{t.1.5}. 
From the table, our VersReID obtains 97.1\% Rank-1 accuracy and 98.2\% mAP, which achieves significant improvements compared to other multi-scene methods, gaining +3.0\% at the Rank-1 accuracy against the PFD~\cite{pfd}.
Besides, the best single scene method PS-HRNet~\cite{zhang_2021_TIP} achieves a Rank-1 accuracy of 92.6\%, which is -4.5\% lower than ours.
These results indicate that the VersReID can effectively work on the low-resolution ReID scene.

\vspace{0.6\baselineskip}\noindent\textbf{- Clothing change ReID scene.}
The experimental results of clothing change ReID are shown in Table~\ref{t.2}. 
Note that our VersReID only leverages the global feature representation of images without employing other artificially designed priors such as the face, contour, body parts, and shape information.

Firstly, we compare the multi-scene models. 
According to the results, compared to the ReIDCaps~\cite{cele} designed for the clothing change scene, our VersReID gains +0.9\% and +11.0\% improvements at the Rank-1 accuracy on Celeb-ReID and PRCC~\cite{prcc}, respectively.
Moreover, the proposed VersReID outperforms all the single-scene methods that do not utilize auxiliary information by a large margin. 
For example, VersReID achieves 61.3\% Rank-1 accuracy on the PRCC dataset, which is +10.8\% higher than LightMBN~\cite{lightmbn}.
Regarding the methods using auxiliary information, IRANet\ddag~\cite{iranet}, RCSANet\ddag~\cite{rcsanet}, and CASE-Net\ddag~\cite{casenet} surpass our VersReID by approximately +0.4$\sim$1.9\% mAP on Celeb-ReID. 
Since such methods require auxiliary models or handcrafted designs for extracting auxiliary information, their performance is affected by the auxiliary models and consumes more computation cost.
In comparison, our VersReID does not require any auxiliary information, which could be more robust and practical.

\vspace{0.6\baselineskip}\noindent\textbf{- Occlusion ReID scene.}
Table~\ref{t.3} (left) shows the results on the occlusion ReID dataset Occ-Duke~\cite{pgfa}.
From the table, the proposed VersReID obtains 72.9\% Rank-1 accuracy and 64.9\% mAP, outperforming the single-scene PFD*~\cite{pfd} that uses key point information by +3.4\% Rank-1 accuracy and +3.1\% mAP, respectively.
Although the BPBreID~\cite{bpbreid} achieves +2.2\% higher Rank-1 accuracy than our VersReID, VersReID achieves +2.4\% higher mAP.
Besides, compared to other multi-scene methods, our VersReID also achieves significant improvements.
For example, our VersReID surpasses the ReIDCaps~\cite{cele} by +6.2\% Rank-1 accuracy on Occ-Duke.
These results clearly demonstrate the effectiveness of our VersReID for solving the occlusion in ReID.

\renewcommand{\arraystretch}{1.05}
\begin{table}[t]
  \caption{Comparison with other methods on the clothing change ReID scene (Celeb-ReID and PRCC). ``\ddag'': the methods use an \textbf{additional model} to extract auxiliary information like face, body parts, or human shape. \\ ``*'': using an overlapping patch embedding layer. \vspace{-6pt}}
  \label{t.2}
  \centering
  \resizebox{0.95\linewidth}{!}{
  \begin{tabular}{c | l | c c | c c }
    \toprule
    Training & \makecell[c]{\multirow{2}{*}{Method}} & \multicolumn{2}{c|}{Celeb-ReID} & \multicolumn{2}{c}{PRCC} \\
    \cline{3-6}
    Set &  & R-1 & mAP & R-1 & mAP \\
    \hline
    & FSAM\ddag~\cite{fsam} & - & - & 54.5 & - \\
    & PCB~\cite{pcb} & 37.1 & 8.2 & 41.8 & 38.7 \\
    & HACNN~\cite{hacnn} & 47.6 & 9.5 & 21.8 & - \\
    & MGN~\cite{mgn} & 49.0 & 10.8 & 33.8 & 35.9 \\
    & ReIDCaps~\cite{cele} & 51.2 & 9.8 & - & - \\
    & AFD-Net~\cite{afdnet} & 52.1 & 10.6 & 42.8 & - \\
    \multirow{2}{*}{Single} & RCSANet~\cite{rcsanet} & 55.6 & 11.9 & 31.6 & 31.5 \\
    \multirow{2}{*}{Scene} & LightMBN~\cite{lightmbn} & 57.3 & 14.3 & 50.5 & 51.2 \\
    & TransReid~\cite{transreid} & 58.9 & 14.6 & 45.0 & 57.9 \\
    & ViT-B~\cite{vit} & 59.7 & 13.4 & 44.3 & 57.5 \\
    & ReIDCaps~\cite{cele} & 63.0 & 15.8 & 38.0 & - \\
    & IRANet\ddag~\cite{iranet} & 64.1 & 19.0 & 54.9 & 53.0 \\
    & RCSANet\ddag~\cite{rcsanet} & 65.3 & 17.5 & 50.2 & 48.6 \\
    & CASE-Net\ddag~\cite{casenet} & 66.4 & 18.2 & 39.5 & - \\
    \hline \hline
    \specialrule{0em}{0.5pt}{0.5pt}
    & PCB~\cite{pcb} & 10.1 & 1.7 & 44.8 & 57.5 \\
        & \major{AIM}(R50)~\cite{yang2023good} & \major{54.5} & \major{12.4} & \major{55.4} & \major{66.0} \\
    & \major{ETNDNET}(R50)~\cite{etndnet} & \major{55.8} & \major{12.7} & \major{45.5} & \major{57.2} \\
    & \major{AIM}(ViT-B)~\cite{yang2023good} & \major{51.3} & \major{10.5} & \major{48.2} & \major{60.7} \\
    Multiple & \major{ETNDNET}(ViT-B)~\cite{etndnet} & \major{49.8} & \major{8.5} & \major{47.6} & \major{59.2} \\
    
    Scenes & \major{PFD}~\cite{pfd} & \major{54.0} & \major{11.0} & \major{55.5} & \major{65.8} \\
    & TransReid~\cite{transreid} & 57.4 & 14.8 & 48.5 & 62.2 \\
    & \major{ReIDCaps}~\cite{cele} & \major{59.9} & \major{17.7} & \major{50.3} & \major{64.4} \\
    \specialrule{0em}{0.5pt}{0.5pt}
    \cline{2-6}
    \specialrule{0em}{0.5pt}{0.5pt}
    & \textbf{VersReID (ours)} & 60.8 & 17.1 & 61.3 & 72.6 \\
    & \textbf{VersReID* (ours)} & 61.7 & 18.7 & 60.7 & 71.4 \\
    \bottomrule
  \end{tabular}
  }
\end{table}

\renewcommand{\arraystretch}{1.1}
\begin{table}[t]
  \caption{Comparison with other methods on the occlusion ReID scene (Occ-Duke) and the cross-modality ReID scene (SYSU-mm01). \\ ``*'' means using an overlapping patch embedding layer. \vspace{-6pt}}
  \label{t.3}
  \centering
  \resizebox{\linewidth}{!}{
  \begin{tabular}{c | l | c c || l | c c}
    \toprule
    Training & \makecell[c]{\multirow{2}{*}{Method}} & \multicolumn{2}{c||}{Occ-Duke} & \makecell[c]{\multirow{2}{*}{Method}} & \multicolumn{2}{c}{SYSU-mm01}\\
    \cline{3-4}\cline{6-7}
    Set &  & R-1 & mAP &   & R-1 & mAP\\
    \hline
    & FD-GAN~\cite{fdgan} & 40.8 & - & JSIA~\cite{jsia} & 38.1 & 36.9 \\
    & PCB~\cite{pcb} & 42.6 & 33.7 & HPILN~\cite{hpiln} & 41.4 & 43.0 \\
    & Ad-Occ~\cite{ad-occ} & 44.5 & 32.2 & AliGAN~\cite{aligan} & 42.4 & 70.7 \\
    & PVPM~\cite{pvpm} & 47.0 & 37.7 & CMSP~\cite{mm01} & 43.6 & 45.0 \\
    & PGFA~\cite{pgfa} & 51.4 & 37.3 & TransReid~\cite{transreid} & 45.2 & 46.8 \\
    & HOReID~\cite{horeid} & 55.1 & 43.8 & Attri~\cite{attri} & 47.1 & 47.1 \\
    & CACE-Net~\cite{cace} & 58.8 & 50.8 & AGW~\cite{agw} & 47.5 & 47.7 \\
    & MoS~\cite{mos} & 61.0 & 49.2 & DFE~\cite{dfe} & 48.7 & 48.6 \\
    \multirow{2}{*}{Single} & OAMN~\cite{occ_chen_2021_iccv} & 62.6 & 46.1 & XIV~\cite{li2020infrared} & 49.9 & 50.7 \\
    \multirow{2}{*}{Scene} & ISP~\cite{isp} & 62.8 & 52.3 & ViT-B~\cite{vit} & 50.1 & 50.9 \\
    & ViT-B~\cite{vit} & 63.0 & 55.2 & HAT~\cite{hat} & 55.3 & 53.9 \\
    & RFCNet~\cite{occ_recover_3} & 63.9 & 54.5 & HC~\cite{hc} & 57.0 & 55.0 \\
    & TransReid~\cite{transreid} & 64.2 & 55.7 & VT-ReID~\cite{liu2020parameter} & 61.7 & 57.5 \\
    & PAT~\cite{pat} & 64.5 & 53.6 & CM-NAS~\cite{fu2021cm} & 62.0 & 60.0 \\
    & TransReid*~\cite{transreid} & 66.4 & 59.2 & MCLNet~\cite{hao2021cross} & 65.4 & 62.0 \\
    & FED~\cite{occ_wang_2022} & 68.1 & 56.4 & SMCL~\cite{wei2021syncretic} & 67.4 & 61.8 \\
    & PFD*~\cite{pfd} & 69.5 & 61.8 & MPANet~\cite{wu2021discover} & 70.6 & 68.2 \\
    & BPBreID~\cite{bpbreid} & 75.1 & 62.5 & CMT~\cite{jiang2022cross} & 71.9 & 68.6 \\
    \hline \hline
    \specialrule{0em}{0.5pt}{0.5pt}
    & PCB~\cite{pcb} & 46.7 & 39.0 & PCB~\cite{pcb} & 52.7 & 48.3 \\
        & \major{AIM}(R50)~\cite{yang2023good} & \major{52.3} & \major{42.9} & \major{AIM}(R50)~\cite{yang2023good} & \major{58.2} & \major{56.4} \\
    & \major{ETNDNET}(R50)~\cite{etndnet} & \major{58.2} & \major{48.7} & \major{ETNDNET}(R50)~\cite{etndnet}  & \major{55.1} & \major{50.0} \\
    & \major{AIM}(ViT-B)~\cite{yang2023good} & \major{62.9} & \major{54.2} & \major{AIM}(ViT-B)~\cite{yang2023good} & \major{60.6} & \major{60.2} \\
    Multiple & \major{ETNDNET}(ViT-B)~\cite{etndnet} & \major{44.8} & \major{38.4} & \major{ETNDNET}(ViT-B)~\cite{etndnet} & \major{51.4} & \major{43.4} \\
    
    Scenes & \major{PFD~\cite{pfd}} & \major{59.5} & \major{52.0} & \major{PFD~\cite{pfd}} & \major{62.7} & \major{62.0} \\     
    & TransReid~\cite{transreid} & 64.2 & 56.6 & TransReid~\cite{transreid} & 58.8 & 59.5 \\
    & \major{ReIDCaps~\cite{cele}} & \major{66.7} & \major{58.8} & \major{ReIDCaps~\cite{cele}} & \major{63.2} & \major{59.9} \\    
    \specialrule{0em}{0.5pt}{0.5pt}
    \cline{2-7}
    \specialrule{0em}{0.5pt}{0.5pt}
    & \textbf{VersReID (ours)} & 72.9 & 64.9 & \textbf{VersReID (ours)} & 67.7 & 65.7 \\
    & \textbf{VersReID* (ours)} & 75.2 & 66.1 & \textbf{VersReID* (ours)} & 69.3 & 66.9 \\
    \bottomrule
  \end{tabular}
  }
\end{table}

\vspace{0.6\baselineskip}\noindent\textbf{- Cross-modality ReID scene.}
The right part of Table~\ref{t.3} shows the performance on the cross-modality ReID dataset SYSU-mm01~\cite{mm01}.
Similar to the comparison on the other scenes, our VersReID outperforms other multi-scene methods significantly. 
Specifically, VersReID achieves 67.7\% Rank-1 accuracy and 65.7\% mAP, which outperform the PFD~\cite{pfd} by +5.0\% and +3.7\%, respectively.
VersReID shows a decent performance compared to most of the single-scene methods designed for the cross-modality ReID problem. 
For example, VersReID outperforms SMCL~\cite{wei2021syncretic} by +3.9\% mAP.
Although the performance of VersReID is slightly behind the best cross-modality method CMT~\cite{jiang2022cross} by -2.9\% mAP, our VersReID can simultaneously handle other scenes. 
In contrast, the other single-scene methods are tailored for the cross-modality ReID problem and hardly be applied to other scenes.
Overall, the experimental results reflect that our method can handle the cross-modality ReID scene well.

\vspace{0.6\baselineskip}\noindent\textbf{- Discussions.}
The above experimental results have shown that the proposed VersReID consistently achieves impressive performance on various ReID scenes \textbf{\major{without fine-tuning}} on any specific datasets.
It is notable that when we use an overlapping patch embedding layer to extract fine-grained feature representations, the performance will be further improved.
For example, as shown in Table~\ref{t.2} and Table~\ref{t.3}, VersReID* outperforms VersReID by +1.6\%, +1.2\%, and +1.2\% in terms of the mAP on Celeb-ReID~\cite{cele}, Occ-Duke~\cite{pgfa}, and SYSU-mm01~\cite{mm01}, respectively.

In summary, we demonstrate the possibility of using a versatile model to solve the ReID task in various scenes, which is of great significance for ReID in practice because the conventional assumptions of only testing under a specific ReID scene are hardly held in reality.

\renewcommand{\arraystretch}{1.2}
\begin{table*}[t]

  \caption{\major{Comparison to single-dataset setting. When there is only one dataset, our VersReID degenerates to the single-dataset ViT-B (MPDA) model, which is a ViT-Base model with our MPDA pre-training.
  The single-dataset ViT-B is trained and evaluated on each dataset separately while our VersReID is trained on the multi-scene training set and evaluated on each dataset without fine-tuning. 
  In other words, \textbf{the prerequisite of evaluating single-dataset ViT-B is knowing scene labels in inference}, which is not adaptive for multi-scene ReID.
  Our VersReID can be further enhanced by overlapping patch embedding layer, denoted as VersReID*.
  We also report the average performance among seven datasets for comparison.\vspace{-8pt}}}
  \label{rt.8}
  \centering
  \resizebox{\linewidth}{!}{
  
  \begin{tabular}{c  c | c c | c c | c c | c c | c c | c c | c c | c c  }
    \toprule
    \makecell[c]{\multirow{3}{*}{Model}} & \textbf{Requiring} & \multicolumn{4}{c|}{General} & \multicolumn{2}{c|}{Low-resolution} & \multicolumn{4}{c|}{Clothing Change} & \multicolumn{2}{c|}{Occlusion} & \multicolumn{2}{c}{Cross-modality} & \multicolumn{2}{|c}{\multirow{2}*{Average}} \\
    \cline{3-16}
    & \textbf{Scene} & \multicolumn{2}{c|}{Market-1501} & \multicolumn{2}{c|}{MSMT17} & \multicolumn{2}{c|}{MLR-CUHK03} & \multicolumn{2}{c|}{Celeb-ReID} & \multicolumn{2}{c|}{PRCC} & \multicolumn{2}{c|}{Occ-Duke} & \multicolumn{2}{c}{SYSU-mm01} & \multicolumn{2}{|c}{} \\
    \cline{3-18}
    & \textbf{Label} & R-1 & mAP & R-1 & mAP & R-1 & mAP & R-1 & mAP & R-1 & mAP & R-1 & mAP & R-1 & mAP & R-1 & mAP \\
    \hline
    Single-dataset & \multirow{2}{*}{\large{\ding{51}}} & \multirow{2}{*}{96.1} & \multirow{2}{*}{91.9} & \multirow{2}{*}{88.6} & \multirow{2}{*}{74.1} & \multirow{2}{*}{95.8} & \multirow{2}{*}{94.9} & \multirow{2}{*}{61.6} & \multirow{2}{*}{17.2} & \multirow{2}{*}{54.4} & \multirow{2}{*}{66.6} & \multirow{2}{*}{72.7} & \multirow{2}{*}{64.0} & \multirow{2}{*}{64.6} & \multirow{2}{*}{63.5}  & \multirow{2}{*}{76.3} & \multirow{2}{*}{67.5} \\
    ViT-B (MPDA) \\
    \hline
    VersReID & \large{\ding{55}} & \textbf{96.9} & 93.0 & 88.5 & 73.6 & {97.1} & {98.2} & {60.8} & {17.1} & \textbf{61.3} & \textbf{72.6} & {72.9} & {64.9} & {67.7} & {65.7}  & 77.9 & 69.3 \\ 
    VersReID* & \large{\ding{55}} & 96.8 & \textbf{93.2} & \textbf{88.8} & \textbf{74.2} & \textbf{97.5} & \textbf{98.4} & \textbf{61.7} & \textbf{18.7} & {60.7} & {71.4} & \textbf{75.2} & \textbf{66.1} & \textbf{69.3} & \textbf{66.9}  & \textbf{78.6}  & \textbf{69.8} \\    
    \bottomrule
  \end{tabular}
  }
  \vspace{-8pt}
\end{table*}

\renewcommand{\arraystretch}{1.2}
\begin{table}[t]
  \caption{\major{
  Comparisons with other methods on the computational costs and the performance. All methods share the same training recipe and do not use scene labels for evaluation. \S~denotes the parameters of an extra pose estimator used in PFD~\cite{pfd}.
  Our VersReID achieves the best performance and competitive computational costs in methods with the ViT-Base backbone.
  Our method can be enhanced by overlapping patch embedding layer, denoted as VersReID*.  } }
  \label{t.4.A}
  \centering
  \resizebox{1.0\linewidth}{!}{
  \begin{tabular}{l | c c | c c c}
    \toprule
    \makecell[c]{\multirow{2}{*}{Method}} & \multicolumn{2}{c|}{Joint Testing Set} & \multicolumn{3}{c}{Computational Costs}\\
    \cline{2-6}
    & R-1 & mAP & Params (M) & GMAC & TP (imgs/s) \\
    \hline 
    \multicolumn{6}{c}{ResNet50 backbone} \\
    PCB~\cite{pcb} & 48.5 & 36.8 & 24.03 & 6.12 & 813.92 \\
    AIM(R50)~\cite{yang2023good} & 65.2 & 46.4 & 23.52 & 6.12 & 1045.33 \\
    ETNDNET(R50)~\cite{etndnet} & 66.4 & 47.7 & 23.51 & 6.12 & 1075.08 \\
    \cdashline{1-6}
    \multicolumn{6}{c}{ViT-Base backbone} \\ 
    ETNDNET(ViT-B)~\cite{etndnet} & 58.8 & 41.7 & 85.80 & 16.52 & 276.47 \\
    AIM(ViT-B)~\cite{yang2023good} & 68.7 & 54.9 & 85.80 & 16.52 & \textbf{277.38} \\
    PFD~\cite{pfd} & 69.8 & 55.9 & 142.52 + 63.59$^\S$ & 26.17 & 90.11 \\
    TransReid~\cite{transreid} & 72.4 & 59.9 & 108.37 & 17.91 & 257.80 \\
    ReIDCaps~\cite{cele} & 74.0 & 60.6 & 96.10 & 16.55 & 182.52 \\
    \textbf{VersReID (ours)} & 76.3 & 63.8 &  \textbf{85.80} &  \textbf{16.52} &  274.56 \\
    \textbf{VersReID* (ours)} & \textbf{77.2} & \textbf{64.7} & 85.90 & 27.04 & 158.72 \\
    \bottomrule

  \end{tabular}
  }
\end{table}

\subsubsection{\major{Comparisons on the Joint Testing Set}}
\label{s.5.2.3}
\major{We compare the performance of different multi-scene methods on the joint testing set, which is more challenging than single-scene testing. We also report the computational costs, including the number of parameters (M), computational amount (GMAC), and throughput (abbreviated as TP, images per second under 256 batch size on an RTX-8000 GPU).}

\major{
The results are shown in Table~\ref{t.4.A}.
According to the results, the methods designed for the specific scene cannot achieve satisfactory performance on multi-scene ReID.
For example, AIM(R50) only achieves 65.2\% Rank-1 accuracy on the joint testing set.
With the transformer backbone, AIM(ViT-B) achieves 68.7\% Rank-1 accuracy, while our VersReID achieves 76.3\% Rank-1 accuracy largely outperforming AIM(ViT-B).
Besides, PFD designed for the occlusion scene cannot be consistently generalized to other scenes and only achieves 69.8\% Rank-1 accuracy.
It is mainly because the knowledge for the specific scene is not equal to the knowledge for multi-scene ReID. For example, while the visible clothes could be an important cue for occluded ReID, the clothes are not helpful for clothing change ReID.}

\major{Moreover, benefiting from the representation ability of the capsule network, ReIDCaps achieves 74.0\% Rank-1 accuracy on the joint testing set, which is 2.3\% lower than our VersReID.
\yxa{However, the competitive performance is because the joint testing set is dominated by the images from the general scene and ReIDCaps can effectively handle ReID under the general scene.
Only performing well on the joint testing set does not mean ReIDCaps is effective in multi-scene ReID.}
As shown in Table~\ref{t.3}, ReIDCaps achieves 66.7\% and 63.2\% Rank-1 accuracy in the occlusion scene and cross-modality scene, which is 6.2\% and 4.5\% lower than ours, respectively.
Besides, as shown in Table~\ref{t.4.A}, ReIDCaps is computationally costly and our method significantly outperforms ReIDCaps in terms of throughput.
Overall, our VersReID has significant advantages in both performance and computational overhead.
}

\vspace{0.6\baselineskip}\noindent\textbf{- Indispensability of a versatile ReID model.}
\major{To show the advantages of learning a versatile ReID model on the multi-scene dataset, we report the result of training our model in a single dataset.
When restricting training our VersReID in one single dataset, the scene-specific prompts and the knowledge distillation for unifying knowledge from different scenes are not needed. Consequently, our VersReID degenerates to a simple ViT-B (MPDA) model, which is a ViT-B model with our MPDA pre-training. 
We report the comparison results with the single-dataset ViT-B (MPDA) models that are separately trained in individual datasets in Table~\ref{rt.8}.
From the results, we can observe that in most of the time our VersReID achieves superior performance to the single-dataset ViT-B (MPDA) models, especially under clothing change, occlusion, and cross-modality scenes.
Our VersReID achieves 77.9\% Rank-1 accuracy averaged from seven datasets, which is 1.6\% higher than that of the single-dataset ViT-B (MPDA).
Especially, single-dataset ViT-B (MPDA) achieves 54.4\% Rank-1 accuracy on PRCC and 64.6\% Rank-1 accuracy on SYSU-mm01, which is 6.9\% and 3.1\% lower than VersReID. 
It is possibly because of the overfitting of the model, as the PRCC and SYSU-mm01 datasets are small in data scales. These results demonstrate that unifying knowledge from different scenes can promote the forming of a powerful versatile model.}

\major{Apart from the performance gap, another advantage is that our method is adaptive for multi-scene ReID without requiring scene labels for inference. In contrast, knowing scene labels is an essential prerequisite for applying multiple single-dataset models since we have to manually choose the correct model for the input images from different scenes.
}

\major{We also discuss the ensemble of multiple single-scene models for multi-scene ReID. Experimental results show that our method outperforms various ensemble methods by a clear margin. Please refer to Supplementary for details. }

\renewcommand{\arraystretch}{1.3}
\begin{table*}[t]
  \caption{Comparisons between the two branches of VersReID: the ReID Bank and the V-Branch. ``Require Scene Lable'' denotes whether scene labels are required for the evaluation. ``Joint Testing Set'' is the build multi-scene ReID joint testing set mentioned in Section~\ref{s.5.1}. \vspace{-8pt}}
  \label{t.4}
  \centering
  \resizebox{\linewidth}{!}{
  \begin{tabular}{c | c | c c | c c | c c | c c | c c | c c | c c | c c}
    \toprule
    & \textbf{Requiring} & \multicolumn{4}{c|}{General} & \multicolumn{2}{c|}{Low-resolution} & \multicolumn{4}{c|}{Clothing Change} & \multicolumn{2}{c|}{Occlusion} & \multicolumn{2}{c|}{Cross-modality} & \multicolumn{2}{c}{Joint}\\
    \cline{3-16}
    \makecell[c]{Model} & \textbf{Scene} & \multicolumn{2}{c|}{Market-1501} & \multicolumn{2}{c|}{MSMT17} & \multicolumn{2}{c|}{MLR-CUHK03} & \multicolumn{2}{c|}{Celeb-ReID} & \multicolumn{2}{c|}{PRCC} & \multicolumn{2}{c|}{Occ-Duke} & \multicolumn{2}{c|}{SYSU-mm01} & \multicolumn{2}{c}{Testing Set} \\
    \cline{3-18}
    & \textbf{Label} & R-1 & mAP & R-1 & mAP & R-1 & mAP & R-1 & mAP & R-1 & mAP & R-1 & mAP & R-1 & mAP & R-1 & mAP \\
    \hline 
    ReID Bank  & \large{\ding{51}} & 96.4 & 92.5 & 87.9 & 72.5 & 96.3 & 97.6 & \textbf{61.9} & \textbf{17.3} & {60.0} & 71.2 & 71.6 & 63.9 & \textbf{68.6} & \textbf{68.5} & \textbf{78.7} & \textbf{66.2} \\
    \hline
    V-Branch & \large{\ding{55}} & \textbf{96.9} & \textbf{93.0} & \textbf{88.5} & \textbf{73.6} & \textbf{97.1} & \textbf{98.2} & 60.8 & 17.1 & \textbf{61.3} & \textbf{72.6} & \textbf{72.9} & \textbf{64.9} & 67.7 & 65.7 & 76.3 & 63.8 \\
    \bottomrule
  \end{tabular}
  }
  \vspace{-4pt}
\end{table*}

\subsection{Ablation Studies}
\label{s.5.3}
To show the effectiveness of each key component contributing to the versatility of our model, we conduct evaluations on the joint testing set mentioned in Section~\ref{s.5.1}, which forms a more complicated and challenging ReID task.
We report the performance on each dataset in our Supplementary.
Unless otherwise stated, in ablation studies, the models are pre-trained in the LUP-B dataset with DINO + MPDA.

\subsubsection{Comparisons between the Two Model Branches}
\label{s.5.3.1}
We compare the two model branches of the VersReID including the ReID Bank obtained at the first stage and the V-Branch obtained at the second stage, to study how the distillation procedure transfers the knowledge from the ReID Bank. 
Note that evaluating the V-Branch does not require scene labels, while the ReID Bank needs scene labels for selecting scene-specific prompts.

According to the experimental results in Table~\ref{t.4}, we can observe that for most of the scenes, the V-Branch achieves comparable performance as the ReID Bank, and even surpasses the ReID Bank on the MSMT17 and Occ-Duke datasets. 
These results indicate the different types of scene-specific knowledge are effectively unified into the V-Branch so that the V-Branch can work well in each specific scene without requiring scene labels. 

\subsubsection{Ablation Studies on the ReID Bank in VersReID}
\label{s.5.3.2}
\vspace{0.6\baselineskip}\noindent\textbf{- The effectiveness of scene-specific prompts.}
\major{\yxa{Our scene-specific prompts have the following advantages. (i) Our scene-specific prompts learn the scene-specific knowledge from different scenes, which is crucial for multi-scene ReID because not all knowledge can be shared among different scenes. 
To further show the importance of scene-specific modeling, we organized the ``Joint Testing Set V2'', which is comprised of the data from clothing change, occlusion, and cross-modality scenes. The images in these scenes exhibit severe shape and texture variations.
As shown in Table~\ref{rt.5}, learning scene-specific prompts brings clear improvement. 
Specifically, without the prompts, we can only achieve 55.7\% and 42.5\% mAP in the joint testing set V2, which is 4.1\% and 3.6\% lower than our full model.
(ii) Our scene-specific prompts make our model more explainable. (iii) In addition to the performance gains, our scene-specific prompts are easy to implement without complicated design or training procedures. We also show that the scene-specific prompts harmonize the ReID learning under different scenes and stabilize the training process. Please refer to Supplementary for details of (ii) and (iii).}}

\major{We also investigate the impact of the number of scene-specific prompts per scene in the ReID Bank (the parameter $N$ in Section~\ref{s.3.1}).
Our model is non-sensitive to $N$ when at least one scene-specific prompt exists.
Please refer to the Supplementary for experiment details.
Since using more prompts increases computational costs, we set $N=2$ to balance the model complexity and the performance.
}

\vspace{0.6\baselineskip}\noindent\textbf{- The robustness of our scene label.}
\major{\yxa{In our work, we define the scene label based on the research subject corresponding to the dataset.
For example, the PRCC~\cite{prcc} and Cele-ReID~\cite{cele} are proposed for the clothing change ReID. 
Thus, we assign the ``clothing change ReID scene'' to the two datasets.
Indeed, we agree that the property of a certain scene can inevitably appear in other scenes, \eg, occlusion can also appear in the clothing change scene.}
Regarding the robustness of the scene label,  we conduct a trial, where we randomly assign scene labels to datasets in training while still using default scene labels for testing\footnote{The evaluation of ReID Bank requires scene labels while V-Branch does not.}.
In this case, the scene labels are considerably noisy. Fortunately, we find that a noisy scene label could reduce the performance but not very serious. Please refer to our Supplementary for details.}

\subsubsection{Ablation Studies on the V-Branch in VersReID}
\label{s.5.3.3}

\begin{figure}[t]
  \centering
  \includegraphics[width=0.85\linewidth]{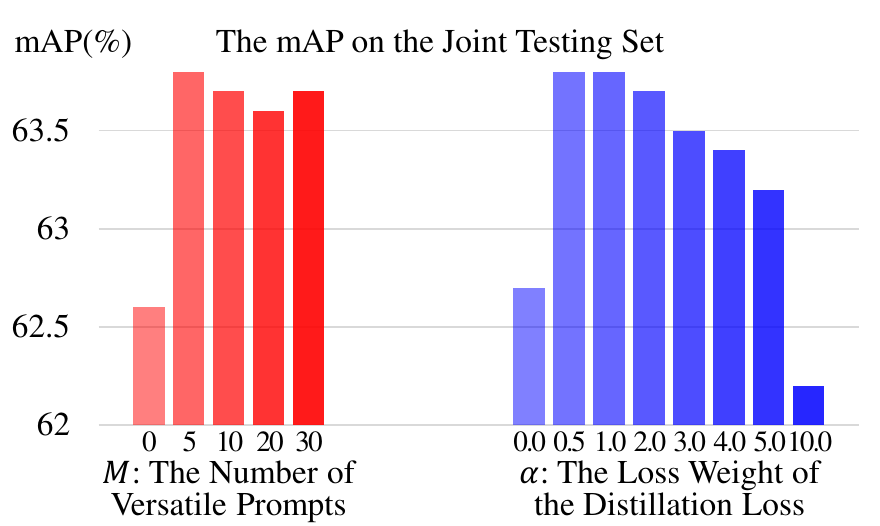}
  \vspace{-8pt}
  \caption{Ablation studies on the number of versatile prompts (left) and the loss weight $\alpha$ (right) in the scene-specific prompts distillation stage for training the \textbf{V-Branch}. We show the mAP on the joint testing set. By default, we utilize five versatile prompts and set the $\alpha$ as 1.0.} 
  \vspace{-8pt}
  \label{f.5}
  
\end{figure}

\begin{table*}[t]
  \caption{\major{The indispensability of scene-specific prompts for modeling scene-specific knowledge. Removing the scene-specific prompts leads to a significant performance drop. All the methods in this table do not use scene labels for evaluation.
  \colorbox{gray}{``Joint Testing Set V2''} is comprised of the testing sets from clothing change, cross-modality, and occlusion ReID datasets, which focuses on severe shape and texture variations while the original joint testing set contains additional general ReID datasets and low-resolution ReID dataset.}}
  \vspace{-7pt}
  \label{rt.5}
  \centering
  \resizebox{\linewidth}{!}{
  \begin{tabular}{c | c c | c c | c c | c c | c c | c c | c c | c c | c c}
    \toprule
    \makecell[c]{\multirow{3}{*}{Model}} & \multicolumn{4}{c|}{General} & \multicolumn{2}{c|}{Low-resolution} & \multicolumn{4}{c|}{Clothing Change} & \multicolumn{2}{c|}{Occlusion} & \multicolumn{2}{c|}{Cross-modality} & \multicolumn{2}{c|}{Joint} & \multicolumn{2}{c}{ \cellcolor{gray}Joint} \\
    \cline{2-19}
    & \multicolumn{2}{c|}{Market-1501} & \multicolumn{2}{c|}{MSMT17} & \multicolumn{2}{c|}{MLR-CUHK03} & \multicolumn{2}{c|}{Celeb-ReID} & \multicolumn{2}{c|}{PRCC} & \multicolumn{2}{c|}{Occ-Duke} & \multicolumn{2}{c|}{SYSU-mm01} & \multicolumn{2}{c|}{Testing Set} &   \multicolumn{2}{c}{ \cellcolor{gray}Testing Set V2} \\
    \cline{2-19}
    & R-1 & mAP & R-1 & mAP & R-1 & mAP & R-1 & mAP & R-1 & mAP & R-1 & mAP & R-1 & mAP & R-1 & mAP & \cellcolor{gray} R-1 & \cellcolor{gray} mAP\\
    \hline
    
    V-Branch (ours) & \textbf{96.9} & \textbf{93.0} & \textbf{88.5} & \textbf{73.6} & \textbf{97.1} & \textbf{98.2} & \textbf{60.8} & \textbf{17.1} & \textbf{61.3} & \textbf{72.6} & \textbf{72.9} & \textbf{64.9} & \textbf{67.7} & \textbf{65.7} & \textbf{76.3} & \textbf{63.8} &  \cellcolor{gray} \textbf{59.8} &  \cellcolor{gray} \textbf{46.1}  \\
    
    W/o prompts & 96.4 & 92.8 & 87.9 & 72.0 & 96.0 & 97.4 & 59.8 & 16.8 & 58.8 & 70.0 & 71.2 & 62.9 & 66.3 & 63.9 & 75.3 & 61.9 &  \cellcolor{gray} 55.7 &  \cellcolor{gray} 42.5 \\
    \bottomrule
  \end{tabular}
  }
  \vspace{-7pt}
\end{table*}

\vspace{0.6\baselineskip}\noindent\textbf{- The number of versatile prompts ($M$).}
According to the left side of Figure~\ref{f.5}, we observe that employing any number of versatile prompts results in promising improvements on most scenes as compared to the method without versatile prompts, \ie, $M=0$.
For example, leveraging five versatile prompts brings +1.2\% improvements on the mAP metrics compared to the performance without versatile prompts (63.8\% \vs 62.6\%).
Regarding Rank-1 accuracy, setting $M=5$ also outperforms setting $M=0$ by +0.8\% performance gains (see details in the Supplementary). 
Note that due to the large number of identities in the joint testing set, improving +0.8\% Rank-1 accuracy means accurately retrieving about 230 query images from approximately 138,000 gallery images, which is a clear performance improvement.

Moreover, using more than five versatile prompts brings limited improvement. 
Considering the computational costs, we set $M=5$, which achieves 76.3\% and 63.8\% in terms of the Rank-1 accuracy and the mAP on the joint testing set.

\vspace{0.6\baselineskip}\noindent\textbf{- The loss weight ($\alpha$).}
$\alpha$ is the trade-off parameter in the loss function (Equation~\ref{eq.2}).
We evaluate our method with different $\alpha$ and show the results on the right side of Figure~\ref{f.5}.
When $\alpha=0.0$, the learning of V-Branch can not utilize the knowledge from the ReID Bank, and the mAP on the joint testing set degrades -1.1\% compared to the performance of $\alpha=1.0$.
Moreover, the performance starts to degrade when $\alpha > 1.0$, indicating that excessive supervision from the ReID Bank will hurt the V-Branch. 
By default, we set $\alpha=1.0$ to balance the supervision of labels and the ReID Bank.

\major{We also investigate the impact of different distillation functions. Please refer to Supplementary for details.}

\renewcommand{\arraystretch}{1.05}
\begin{table}[t] 
  \caption{Ablation studies on \textbf{MPDA}. We study how augmented views affect performance. The used model is V-Branch with the ViT-S backbone. The ``All'' in the table represents all augmented views used, which is the default setting of our MPDA. The ``W/o'' is an abbreviation of without. \vspace{-4pt}}
  \label{t.8}
  \centering
  \resizebox{0.85\linewidth}{!}{
  \begin{tabular}{c | c | c | c}
     \toprule
     \multirow{2}{*}{Index} & \multirow{2}{*}{Augmented Views} & \multicolumn{2}{c}{The Corresponding Scene} \\
     \cline{3-4}
     & & ~~~~~~R-1~~~~~~ & mAP \\
     \hline
     & & \multicolumn{2}{c}{MSMT17} \\
     \cline{3-4}
     1 & All & 63.2 & 41.2 \\
     & W/o Lighting Change & 62.1 & 39.6 \\
     \hline
     & & \multicolumn{2}{c}{MLR-CUHK03} \\
     \cline{3-4}
     2 & All & 89.4 & 90.5 \\
     & W/o Blurred & 86.1 & 88.8 \\
     \hline
     & & \multicolumn{2}{c}{PRCC} \\
     \cline{3-4}
     3 & All & 35.1 & 49.5 \\
     & W/o Clothing Change & 32.9 & 48.7 \\
     \hline
     & & \multicolumn{2}{c}{Occ-Duke} \\
     \cline{3-4}
     4 & All & 46.5 & 37.7 \\
     & W/o Occlusion & 42.4 & 33.7 \\
     \hline
     & & \multicolumn{2}{c}{SYSU-mm01} \\
     \cline{3-4}
     5 & All & 35.8 & 38.6 \\
     & W/o Gray-scale & 35.0 & 37.8 \\    
     \hline
     & & \multicolumn{2}{c}{Joint Testing Set} \\
     \cline{3-4}
     6 & All & 56.2 & 40.8 \\
     & W/o MPDA & 53.7 & 36.9 \\    
    \bottomrule
\end{tabular}
  }
  
\end{table}

\subsubsection{Ablation Studies on MPDA}
\label{s.5.3.5}
In this section, we ablate the main designs in MPDA. 

\vspace{0.6\baselineskip}\noindent\textbf{- Settings.}
Since self-supervised learning is costly, we train a ViT-Small (ViT-S)~\cite{vit} model instead of the ViT-B on the LUP-B dataset to conduct the ablation studies of MPDA. 
For a quick check, we randomly select 500k images from the LUP-B dataset to train the model for 100 epochs in the ablation studies. 
All the experiments share the same training images.
In MPDA, we randomly select two augmented views from the generated candidate views for each sample in each training iteration and adopt the objection functions in DINO~\cite{dino}.
We use the obtained model to initialize the ViT-S backbone of the VersReID framework and then train the VersReID on the multi-scene training set.

\vspace{0.6\baselineskip}\noindent\textbf{- Augmented views.}
Based on the full MPDA with self-supervised learning method DINO\cite{dino}, we remove each specific augmented view from the candidate views to study the impact of each augmented view.
Note that the unchanged view is always preserved.
The ablation results are shown in Table~\ref{t.8}, where we mainly show the performance on the specific scene corresponding to the augmented view. 

Based on the results, it can be observed that the lack of certain augmented views harms the performance of the corresponding scene. 
For instance, when the lighting change view is removed, the mAP on the general scene dataset MSMT17 degrades -1.6\% as shown by Table~\ref{t.8}~(1).
Similarly, when other views such as blurred, clothing change, occlusion, and gray-scale views are removed, the model's performances on the low-resolution dataset MLR-CUHK03~\cite{dslr_market}, clothing change dataset PRCC~\cite{prcc}, occlusion dataset Occ-Duke~\cite{pgfa}, cross-modality dataset SYSU-mm01~\cite{mm01} also decreased accordingly. 
Specifically, as shown by Table~\ref{t.8} (2), (3), (4), and (5), they underperform the full MPDA by -3.3\%, -2.2\%, -4.1\%, and -0.8\% at the Rank-1 accuracy on the corresponding scene, respectively.
The results show the effect of each augmented view, where the performance in related scenes can be enhanced by its corresponding image views. 
Moreover, when the MPDA is removed and only uses standard DINO for self-supervised learning, the mAP on the joint testing test decreases from 40.8\% to 36.9\%.
The considerable performance drops show the necessity of the MPDA for self-supervised multi-scene ReID learning.

\major{We further analyze the impact of each augmented view in different datasets. Details are in Supplementary. }

\begin{figure*}[t]
  \centering
  \includegraphics[width=1\linewidth]{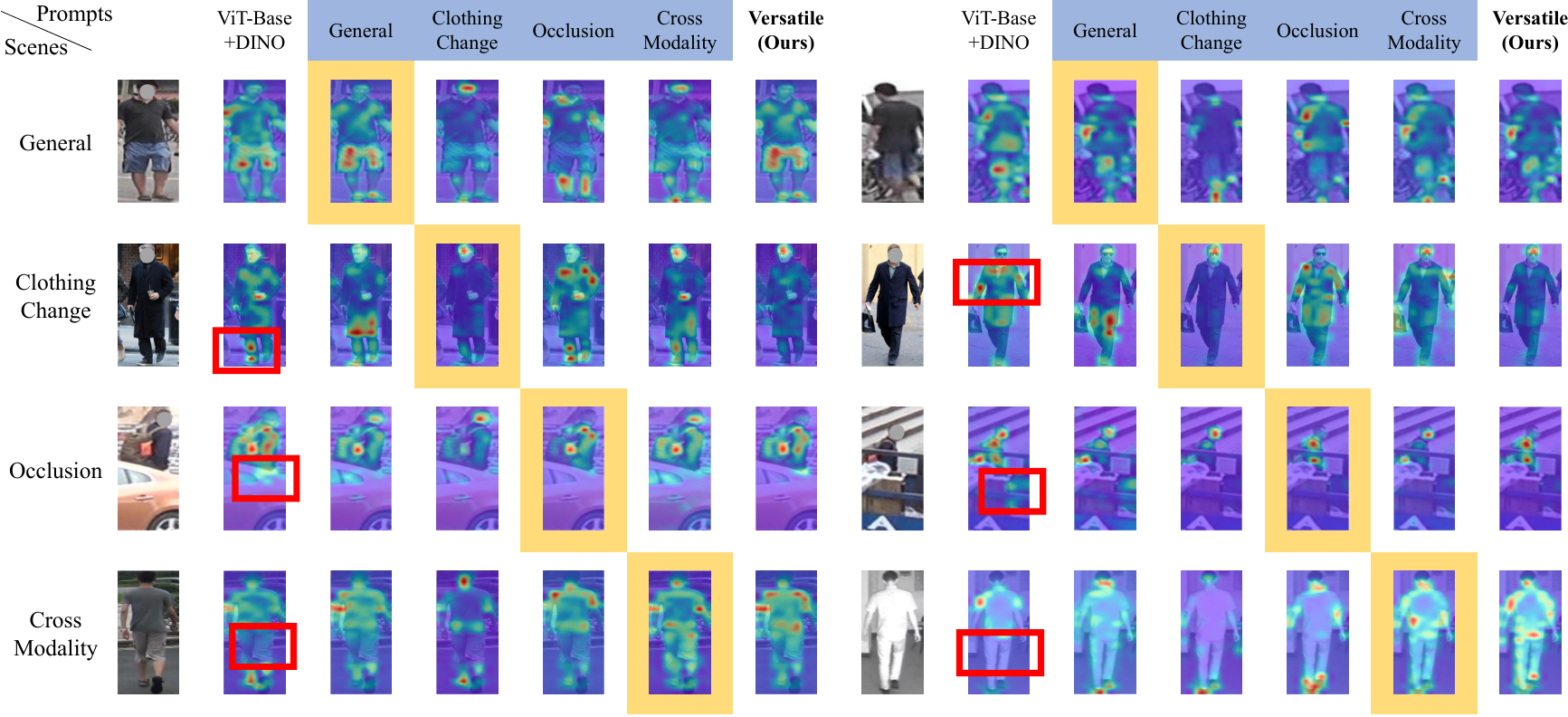}
  \caption{Visualizations of the self-attention map of the class token $e_{[\mathrm{CLS}]}$ in the last transformer block. Attention maps are the average value among all heads in the self-attention module. Each row represents two images of the same person in a specific scene and their attention maps. Each column represents applying the same scene-specific prompts to different input images. Prompts with \colorbox{blue}{blue} background: attention maps in this column are from the ReID Bank. Attention maps with \colorbox{lightyellow}{yellow} background: the scene and prompts correspond with each other.
  \major{The red boxes highlight that the vanilla ViT-B model pays the wrong attention to image parts. Such as focusing on clothes in the clothing-change scene, focusing on the occlusion part in the occlusion scene, and failing to capture shape information in the cross-modality scene.}}
  \label{f.6}
\end{figure*}

\vspace{0.6\baselineskip}\noindent\textbf{- Applying MPDA to other SSL methods.}
In our previous experiments, we applied MPDA to the DINO for self-supervised learning. 
To demonstrate the ``plug-and-play'' nature of MPDA, we apply MPDA to the Mocov3~\cite{mocov3} model. 
We find that augmenting Mocov3 with MPDA outperforms the Mocov3 baseline by +4.7\% and +5.2\% at the Rank-1 accuracy and mAP (details are in Supplementary), respectively. 
The results demonstrate that MPDA can be integrated with other self-supervised learning methods to enhance the ReID in multi-scene conditions.

\subsection{Visualizations}
\label{s.5.4}
To better understand how the proposed VersReID works, we visualize the self-attention maps of the ReID Bank and V-Branch models of VersReID.
\md{For each scene, we take two images from the testing set as the input and leverage different prompts to guide the model.}
We use the ReID Bank for the scene-specific prompts and utilize the V-Branch model for the versatile prompts.
The visualizations are illustrated in Figure~\ref{f.6}, and we analyze the results from the perspective of scenes and prompts, respectively. 

\vspace{0.6\baselineskip}\noindent\textbf{- The perspective of scenes.}
\md{When we view Figure~\ref{f.6} in rows, we can observe that the attention maps are different for the same image with different prompts.
For example, the attention maps of general prompts focus more on the human body, while those of the clothing change prompts suppress the activation of the main body (clothes) and show high responses to the human face and feet (shoes). 
The results suggest that face and foot information may be more effective than attire for ReID under the clothing change since a person may wear various clothes in different cameras.}

\md{Moreover, the occlusion prompts guide the ReID Bank to capture the information about human key points such as shoulders, elbows, and feet. 
Differently, the attention maps guided by other prompts do not pay special attention to human key points. 
Besides, in the fourth row, the ReID Bank overlooks the car and obstruction that cover the person under the guidance of occlusion prompts, while using other prompts will more or less focus on the car or obstruction.}

\md{Furthermore, the attention maps of the ReID Bank with cross-modality prompts seem like a combination of the attention maps of general and clothing change prompts. 
In other words, both the human body and face are highly attended to. 
This phenomenon is probably because infrared images are different from RGB images that contain abundant color attributes. 
Thus the model needs to focus further on facial features to distinguish identities.}

\md{Last but not least, the attention maps obtained by applying versatile prompts to the V-Branch are similar to those of the ReID Bank using scene-specific prompts corresponding to the specific scene, indicating that the distillation successfully transfers the knowledge from ReID Bank to V-Branch.}

\section{Conclusions}%
\label{s.6}
In this work, we have successfully shown the possibility of learning a versatile model for handling multiple ReID scenes, namely different ReID challenges. We accomplish this task by proposing a prompt-based twin modeling framework called \textbf{VersReID}. In the proposed VersReID, a versatile model branch named V-Branch with embedded learnable versatile prompts is distilled across a number of scenes from a multi-scene ReID Bank model with various learnable scene-specific prompts. We also facilitate the effectiveness of learning multi-scene ReID models for distillation by proposing a self-supervised data augmentation strategy with multi-scene prioris named MPDA. With these efforts, when no scene label is available in the inference stage, we have demonstrated that an effective versatile ReID model is constructed to perform re-identification under the general ReID scene as well as under the low-resolution ReID, clothing change ReID, occlusion ReID, and cross-modality ReID scenes. The performance of VersReID is either comparable or even better than any ReID model that is fine-tuned on a specific ReID scene. Our work has overcome the limitation of existing ReID modeling, always only working for a specific scene but degrading largely for others.

\section*{Acknowledgment}
This work was supported partially by the NSFC (U21A20471,U1911401), Guangdong NSF Project (No. 2023B1515040025, 2020B1515120085).
The corresponding author and principal investigator for this work is Yi-Xing Peng.
Wei-Shi Zheng and Junkai Yan contribute equally to this work.

\appendices

\ifCLASSOPTIONcaptionsoff
  \newpage
\fi

\bibliographystyle{IEEEtran}
\bibliography{VersReID}

\begin{IEEEbiography}[{\includegraphics[width=1in,height=1in,clip,keepaspectratio]{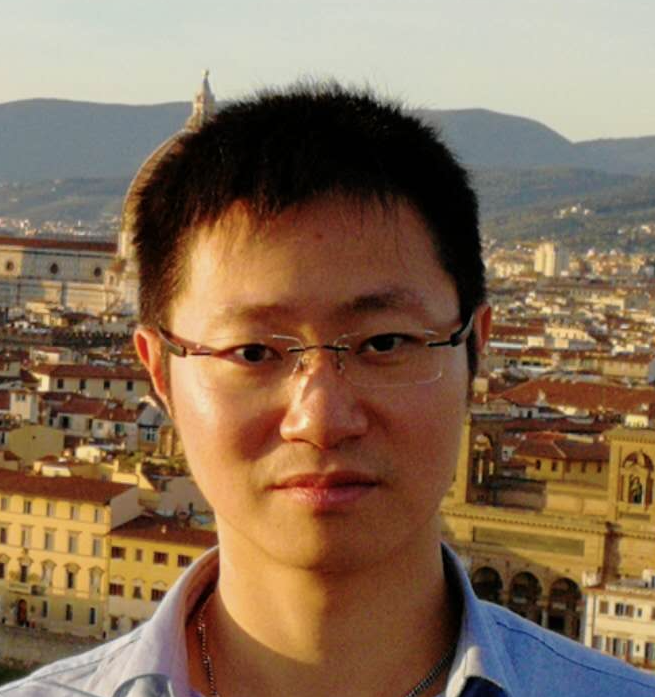}}]{Dr. Wei-Shi Zheng} 
is now a full Professor with Sun Yat-sen University. His research interests include person/object association and activity understanding, and the related weakly supervised/unsupervised and continuous learning machine learning algorithms. He has ever served as area chairs of ICCV, CVPR, ECCV, BMVC, etc. He is associate editors of IEEE-TPAMI, Pattern Recognition. He has ever joined Microsoft Research Asia Young Faculty Visiting Programme. He is a Cheung Kong Scholar Distinguished Professor, a recipient of the Excellent Young Scientists Fund of the National Natural Science Foundation of China, and a recipient of the Royal Society-Newton Advanced Fellowship of the United Kingdom.
\end{IEEEbiography}

\begin{IEEEbiography}[{\includegraphics[width=1in,height=1in,clip,keepaspectratio]{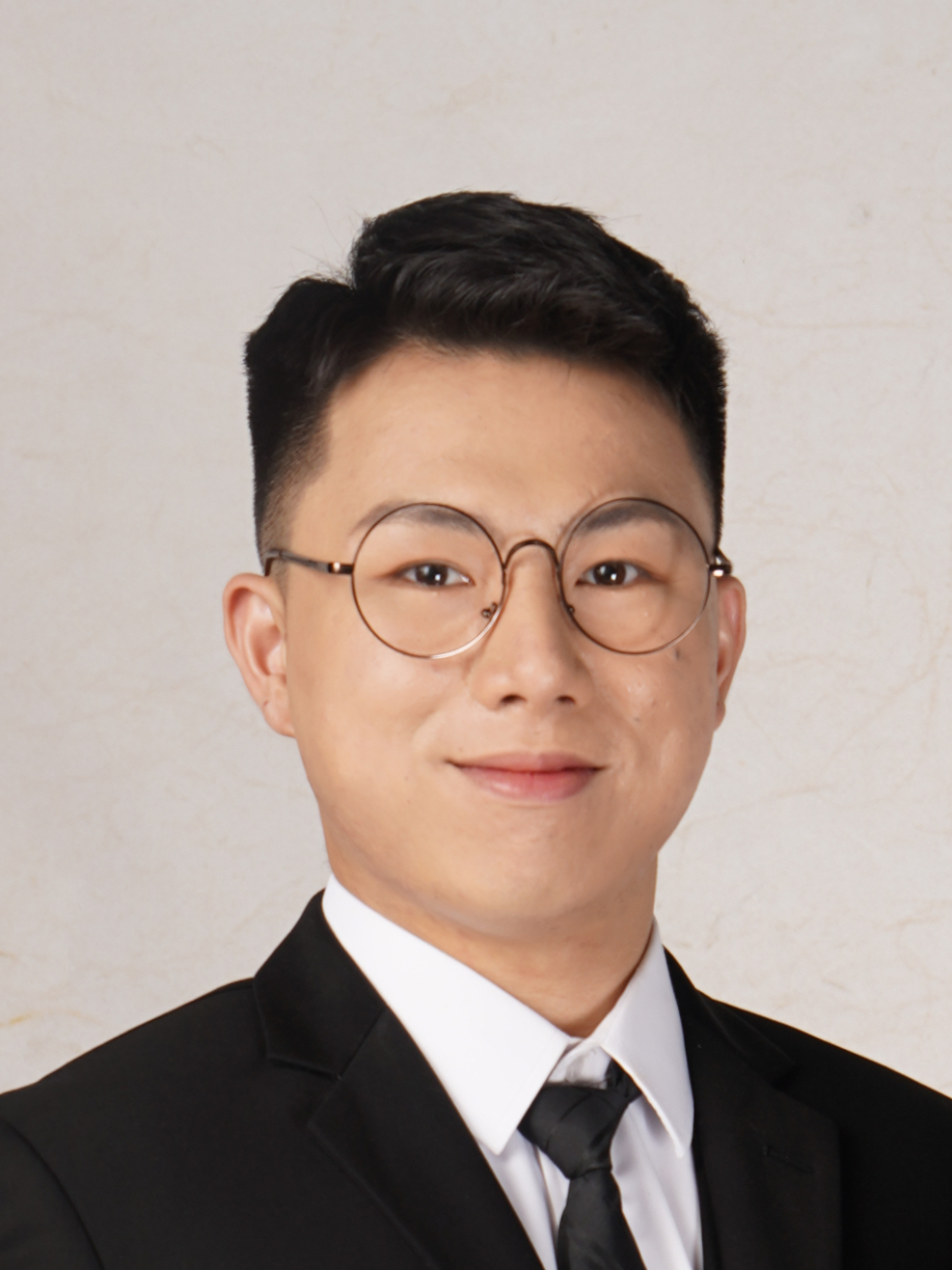}}]{Junkai Yan}
received the bachelor’s degree in computer science and technology from Sun Yat-sen University in 2021. He is working toward the Master degree at the School of Computer Science and Engineering at Sun Yat-sen University. His research interests include self-supervised learning and its application.
\end{IEEEbiography}

\begin{IEEEbiography}[{\includegraphics[width=1in,height=1in,clip,keepaspectratio]{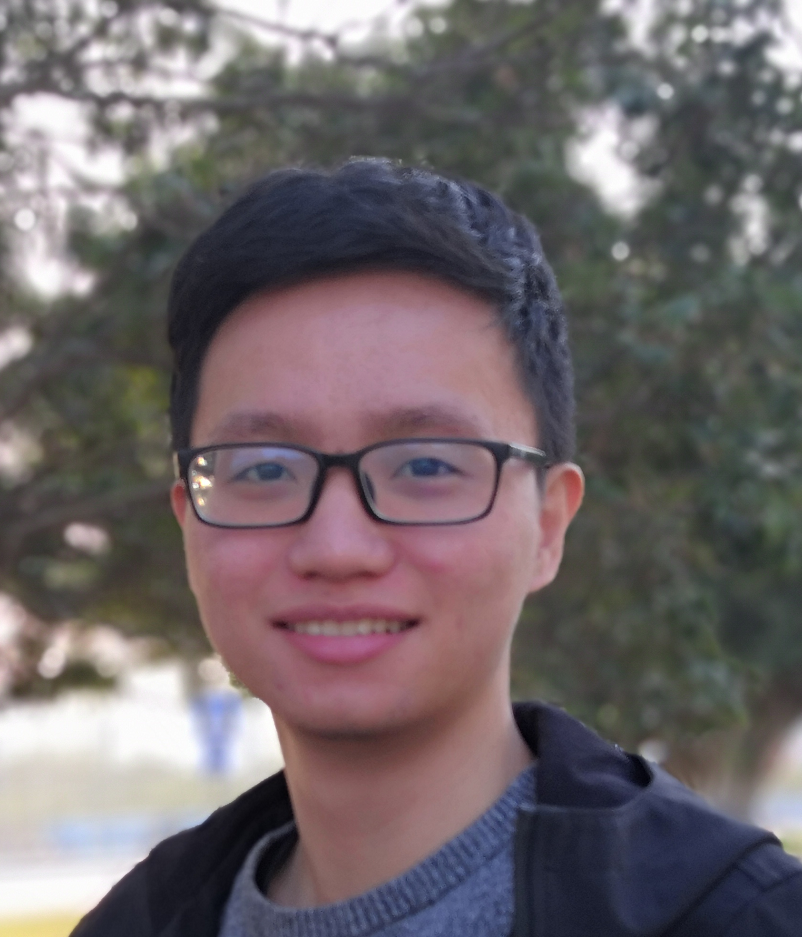}}] {Yi-Xing Peng} received the bachelor’s degree in computer science and technology from Sun Yat-sen University in 2020. He is working toward the Ph.D. degree at the School of Computer Science and Engineering at Sun Yat-sen University. His research interests include computer vision and machine learning.
  \end{IEEEbiography}

}
\end{document}


\title{Supplementary for ``A Versatile Framework for\\Multi-scene Person Re-identification''}
\author{Wei-Shi Zheng, Junkai Yan, Yi-Xing Peng}

\maketitle
\IEEEdisplaynontitleabstractindextext
\IEEEpeerreviewmaketitle

In this supplemental material, we present detailed experimental results of our proposed versatile ReID framework, namely VersReID, and the multi-scene prioris data augmentation strategy (MPDA). 

\section{Detailed Experimental Results on the ReID Bank}
In this section, we provide the comprehensive ablation results of the number of scene-specific prompts $N$ in the ReID Bank. 
Moreover, we also compare the modeling of scene-specific knowledge via scene-specific prompts or scene-specific layers~\cite{hardparashare}.
The results are shown in Table~\ref{t.a1}.
{We also discuss the robustness of the scene labels, using CLIP for prompt initialization, and present more analysis on the scene-specific prompts.}

\vspace{0.5\baselineskip}\noindent\textbf{- Comparisons between scene-specific prompts and scene-specific layers.}
Firstly, we compare the effectiveness of scene-specific knowledge modeling methods, namely the scene-specific prompts, and layers. 
In our experiments, the scene-specific layers are implemented by configuring the last batch normalization (BN) layer and the last $L$ transformer blocks as scene-specific layers, denoted as ``BN + $L$''. 
For example, ``BN + 0'' means only the last BN layer is considered scene-specific.
The results in Table~\ref{t.a1} show that modeling scene-specific knowledge using specific prompts is superior to specific layers. 

\begin{figure*}[ht]
  \centering
  \includegraphics[width=0.55\linewidth]{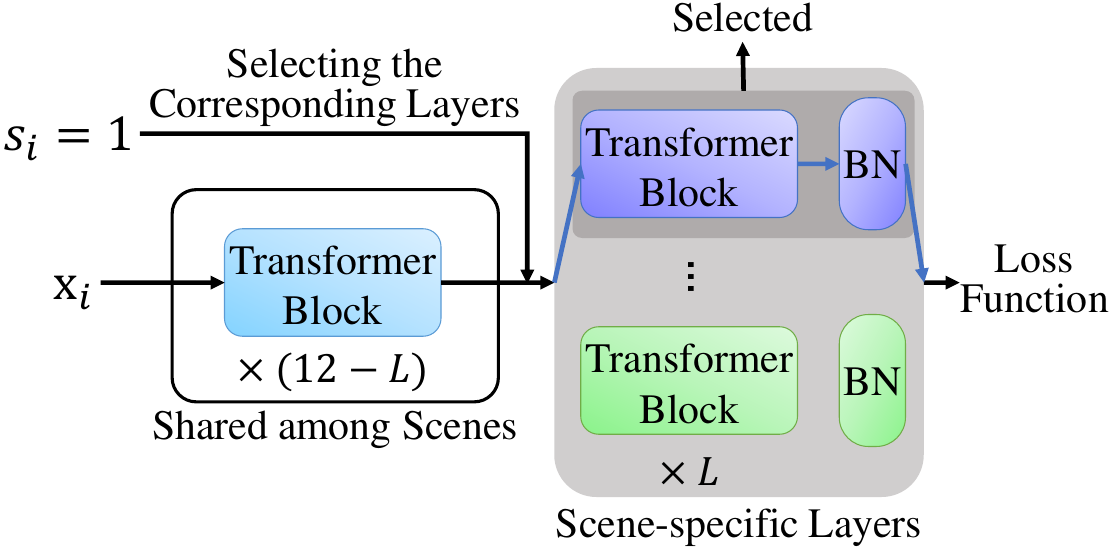}
  \caption{An illustration of modeling scene-specific knowledge via scene-specific layers~\cite{hardparashare}. The ``12'' in the figure indicates the model contains 12 transformer blocks in total. We show an example of $s_i=1$, where the first group of scene-specific layers is selected.}
  \label{f.a1}
\end{figure*}

\renewcommand{\arraystretch}{1.2}
\begin{table*}[ht]
  \caption{Comparisons between modeling scene-specific knowledge via scene-specific layers~\cite{hardparashare} and scene-specific prompts (ours) in the \textbf{ReID Bank}. The $L$ or $N$ represents the number of scene-specific layers or the number of scene-specific prompts for each scene. The ``BN'' represents the batch normalization layer before the classification head, details can be found in~\cite{bot}. The ``BN~+~$L$'' means the last BN layer and the last $L$ transformer blocks are regarded as scene-specific layers. The cell with \colorbox{lightgray}{gray} background in the table is the default setting of the proposed method. }
  \label{t.a1}
  \centering
  \resizebox{\linewidth}{!}{
  \begin{tabular}{c | c | c c | c c | c c | c c | c c | c c | c c | c c}
    \toprule
    & $L$ & \multicolumn{4}{c|}{General} & \multicolumn{2}{c|}{Low-resolution} & \multicolumn{4}{c|}{Clothing Change} & \multicolumn{2}{c|}{Occlusion} & \multicolumn{2}{c|}{Cross-modality} & \multicolumn{2}{c}{Joint}\\
    \cline{3-16}
    Method & or & \multicolumn{2}{c|}{Market-1501} & \multicolumn{2}{c|}{MSMT17} & \multicolumn{2}{c|}{MLR-CUHK03} & \multicolumn{2}{c|}{Celeb-ReID} & \multicolumn{2}{c|}{PRCC} & \multicolumn{2}{c|}{Occ-Duke} & \multicolumn{2}{c|}{SYSU-mm01} & \multicolumn{2}{c}{testing set} \\
    \cline{3-18}
    & $N$ & R-1 & mAP & R-1 & mAP & R-1 & mAP & R-1 & mAP & R-1 & mAP & R-1 & mAP & R-1 & mAP & R-1 & mAP \\
    \hline
    Scene & BN + 0 & 96.0 & 91.0 & 88.1 & 70.9 & 95.2 & 94.7 & 58.8 & 15.9 & 50.7 & 63.0 & 71.8 & 62.9 & 64.3 & 61.1 & 76.2 & 62.9 \\
    Specific & BN + 1 & 95.8 & 91.1 & 87.8 & 71.3 & 95.4 & 95.1 & 60.0 & 16.6 & 51.1 & 63.8 & 71.6 & 62.9 & 61.0 & 60.9 & 76.8 & 63.4 \\
    Layers & BN + 2 & 96.0 & 91.2 & 87.8 & 71.5 & 96.1 & 96.8 & 60.2 & 16.7 & 50.2 & 63.0 & 71.1 & 62.0 & 61.6 & 61.7 & 77.1 & 63.3 \\
    \hline \hline
    & 0 & 96.2 & 92.2 & 87.6 & 71.7 & 95.9 & 95.1 & 60.8 & 16.9 & 57.2 & 68.9 & 70.7 & 62.7 & 67.2 & 64.7 & 74.8 & 61.9 \\
    & 1 & 96.0 & 92.2 & 88.0 & 72.6 & 96.2 & 96.9 & 61.7 & 17.5 & 60.3 & 70.5 & 71.4 & 63.3 & 68.2 & 66.5 & 78.5 & 66.1 \\
    Scene & \cellcolor{lightgray}2 & 96.4 & 92.5 & 87.9 & 72.5 & 96.3 & 97.6 & {61.9} & {17.3} & {60.0} & 71.2 & 71.6 & 63.9 & {68.6} & {68.5} & {78.7} & {66.2} \\
    Specific & 3 & 96.3 & 92.2 & 88.1 & 72.6 & 96.4 & 97.4 & 61.4 & 17.5 & 58.7 & 69.6 & 71.6 & 64.0 & 67.9 & 67.4 & 78.6 & 66.1 \\
    Prompts & 4 & 96.4 & 92.3 & 88.1 & 72.8 & 96.6 & 97.5 & 61.6 & 17.5 & 59.9 & 71.2 & 71.9 & 63.9 & 68.1 & 67.8 & 78.8 & 66.4 \\
    & 5 & 96.3 & 92.2 & 87.8 & 72.5 & 96.6 & 97.7 & 61.4 & 17.3 & 60.1 & 71.0 & 72.0 & 63.8 & 68.8 & 69.0 & 78.8 & 66.3 \\
    & 10 & 96.3 & 92.3 & 88.0 & 72.8 & 96.3 & 97.5 & 61.6 & 17.4 & 60.6 & 71.4 & 72.2 & 64.2 & 68.4 & 68.3 & 78.7 & 66.4 \\
    \bottomrule
  \end{tabular}
  }
\end{table*}

\vspace{0.5\baselineskip}\noindent\textbf{- Impact of the number of scene-specific prompts.}
We investigate the impact of the number of scene-specific prompts per scene (denoted as $N$) in the ReID Bank.
Note that when $N=0$, there is no scene-specific prompt, and the method can represent the intuitive solution, which lacks the modeling of scene-specific knowledge.
Notably, for models with $N>0$, the scene label is required for selecting scene-specific prompts.

\begin{figure}[t]
  \centering
  \includegraphics[width=0.5\linewidth]{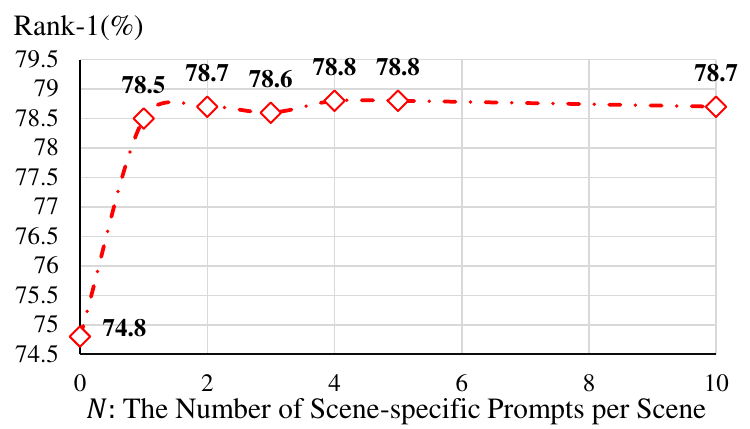}
  \caption{Effectiveness and robustness of the scene-specific prompts in the \textbf{ReID Bank}. We evaluate our model with different $N$ and show the Rank-1 accuracy on the joint testing set. By default, we set $N=2$.}
  \label{f.4}
\end{figure}

According to Figure~\ref{f.4}, our model is non-sensitive to the hyper-parameter $N$ when at least one prompt exists. 
Specifically, regardless of the number of prompts, employing scene-specific prompts to model scene-specific knowledge produces promising improvements compared to the performance when no scene-specific prompt is enabled ($N=0$).
For example, compared to disabling the scene-specific prompts on the joint testing set, our ReID Bank achieves +3.7\% $\sim$ +4.0\% performance gains at the Rank-1 accuracy.
As using more prompts means more computational costs, we set $N=2$ in our main experiments to balance the model complexity and the performance.

\subsection{{Robustness of the scene-specific labels}}
{In our work, the definition of ``scene'' is based on the research subject corresponding to the dataset.
For example, the PRCC~\cite{prcc} and Cele-ReID~\cite{cele} datasets are proposed for studying the clothing change ReID. 
Thus, we assign the ``clothing change ReID scene'' to both of the two datasets.%
Indeed, under such a definition, we agree that the property of a certain scene can inevitably appear in other scenes, \eg, occlusion can also appear in the clothing change scene.
However, the occlusion in the clothing change datasets is much less frequent than that in the occlusion datasets. 
Therefore, it is reasonable to assign scene labels based on the main properties of the dataset.}

{Regarding the robustness of the scene label, we find that a noisy scene label could reduce the performance but is not very serious. We conduct a trial, where we randomly assign scene labels to datasets in training while still using default scene labels for testing\footnote{\yx{The evaluation of ReID Bank requires scene labels while V-Branch does not.}}.
In this case, the scene labels are considerably noisy according to the definition of scene.}

{The results under this setting are shown in the Table~\ref{rt.2}.
We can observe that using random scene labels leads to inferior performance.
The Rank-1 accuracy and mAP drop by 3.0\% and 3.2\% in the joint testing set, respectively.
The inferior performance indicates that the ReID Bank fails to capture the scene-specific knowledge and thus the V-Branch cannot effectively unify knowledge from different scenes.
Nevertheless, despite the clear performance drop, using the random labels does not make our model collapse, which demonstrates our robustness to the scene labels.}

{We emphasize that the experimental results show the robustness of our scene-specific modeling and do not mean the scene-specific modeling is not important.
The scene-specific modeling is crucial for the multi-scene ReID. 
As shown in Table~\ref{rt.5}, removing the scene-specific prompts leads to clear performance drops. For example, the Rank-1 accuracy drops by 2.5\% at PRCC dataset and the mAP drops by 1.9\% at the joint testing set. Considering there are about 138,000 gallery images in the joint testing set, 1.9\% drops at mAP means the gap between the probes and the ground-truth matches is considerably enlarged. 
\textbf{Especially}, we also construct the joint testing set V2 by combining the testing data from clothing change, occlusion, and cross-modality ReID datasets, which contain severe shape and texture variations. The experimental results in Table~\ref{rt.5} show that the scene-specific prompts are crucial to adaptively overcome the shape and texture variations because the Rank-1 accuracy (mAP) in the joint testing set V2 drops by 4.1\% (3.6\%) without scene-specific prompts.}

{Note that the robustness we achieve above is probably attributed to the learning of the scene-shared knowledge along with our scene-specific modeling.
Specifically, the scene labels only affect the learning of scene-specific knowledge as the images are associated with different scene-specific prompts according to the scene labels.
However, since all images are processed by a scene-shared backbone, the scene labels do not hamper the learning of the scene-shared knowledge.
Hence, our method can at least learn the scene-shared knowledge for multi-scene ReID even when the scene labels are extremely noisy.}

{In general, the scene labels characterize distinct challenges in ReID. The scene-specific modeling is beneficial for adaptively handling images from various scenes and stabilizes the learning process on different datasets with distinct challenges.
By exploring scene labels for learning scene-specific prompts, our model can adaptively focus on different parts of the image regardless of which scene the image is from. Our visualization results (Figure 6 in the main manuscript) qualitatively show the benefits of our scene labels.}

\renewcommand{\arraystretch}{1.3}
\begin{table*}[t]
  \caption{
  {Validation of the robustness of the scene label. By default, we use five groups of prompts corresponding to five different scenes. To show the robustness of the scene modeling, we further evaluate our model using random scene labels for datasets in training and use the correct scene labels (\eg, general, occlusion, etc.) for evaluation, denoted as \colorbox{yellow}{ReID Bank (random)}.
  Experimental results show the effectiveness and robustness of our scene modeling, and randomly assigning scene labels in training significantly hinders the multi-scene ReID.} }
  \label{rt.2}
  \centering
  \resizebox{\linewidth}{!}{
  \begin{tabular}{c | c | c c | c c | c c | c c | c c | c c | c c | c c}
    \toprule
    \makecell[c]{\multirow{3}{*}{Method}} & \multirowcell{3}{Scene \\ labels}  & \multicolumn{4}{c|}{General} & \multicolumn{2}{c|}{Low-resolution} & \multicolumn{4}{c|}{Clothing Change} & \multicolumn{2}{c|}{Occlusion} & \multicolumn{2}{c|}{Cross-modality} & \multicolumn{2}{c}{Joint}\\
    \cline{3-16}
    & & \multicolumn{2}{c|}{Market-1501} & \multicolumn{2}{c|}{MSMT17} & \multicolumn{2}{c|}{MLR-CUHK03} & \multicolumn{2}{c|}{Celeb-ReID} & \multicolumn{2}{c|}{PRCC} & \multicolumn{2}{c|}{Occ-Duke} & \multicolumn{2}{c|}{SYSU-mm01} & \multicolumn{2}{c}{Testing Set} \\
    \cline{3-18}
    & & R-1 & mAP & R-1 & mAP & R-1 & mAP & R-1 & mAP & R-1 & mAP & R-1 & mAP & R-1 & mAP & R-1 & mAP \\
    \hline

\cellcolor{yellow}ReID Bank (random) & \cellcolor{yellow}\Checkmark & \cellcolor{yellow}96.4 & \cellcolor{yellow}92.5 & \cellcolor{yellow}87.7 & \cellcolor{yellow}72.1 & \cellcolor{yellow}96.1 & \cellcolor{yellow}96.3 & \cellcolor{yellow}61.2 & \cellcolor{yellow}17.2 &\cellcolor{yellow}59.2 & \cellcolor{yellow}71.1 & \cellcolor{yellow}70.8 & \cellcolor{yellow}62.9 & \cellcolor{yellow}66.7 & \cellcolor{yellow}65.9 & \cellcolor{yellow}75.7 & \cellcolor{yellow}63.0 \\ 

    \cellcolor{yellow}V-Branch (random) & \cellcolor{yellow}\XSolid & \cellcolor{yellow}96.7 & \cellcolor{yellow}92.9 & \cellcolor{yellow}88.0 & \cellcolor{yellow}73.0 & \cellcolor{yellow}96.5
    & \cellcolor{yellow}96.1 & \cellcolor{yellow}60.4 & \cellcolor{yellow}16.9 & \cellcolor{yellow}59.5 & \cellcolor{yellow}71.1 & \cellcolor{yellow}72.1 
    & \cellcolor{yellow}64.0 & \cellcolor{yellow}65.3 & \cellcolor{yellow}64.9 & \cellcolor{yellow}75.4 & \cellcolor{yellow}62.9 \\ \hline

    \textbf{ReID Bank (not random)} & \Checkmark & 96.4 & 92.5 & 87.9 & 72.5 & 96.3 & 97.6 & \textbf{61.9} & \textbf{17.3} & 60.0 & 71.2 & 71.6 & 63.9 & \textbf{68.6} & \textbf{68.5} & \textbf{78.7} & \textbf{66.2} \\
    \textbf{V-Branch(not random)} & \XSolid & \textbf{96.9} & \textbf{93.0} & \textbf{88.5} & \textbf{73.6} & \textbf{97.1} & \textbf{98.2} & 60.8 & 17.1 & \textbf{61.3} & \textbf{72.6} & \textbf{72.9} & \textbf{64.9} & 67.7 & 65.7 & 76.3 & 63.8 \\
    
    \bottomrule
  \end{tabular}
  }
\end{table*}

\subsection{{Using CLIP to extract textual prompts as scene-specific prompts}}
{In our main manuscript, we randomly initialize the scene-specific prompts and optimize the prompts through training.
Since visual-language models like CLIP~\cite{clip} have shown impressive performance in downstream tasks, we investigate adopting a pre-trained CLIP-Large text encoder\footnote{We use the pre-trained model in https://huggingface.co/openai/clip-vit-large-patch14.} to extract the textual embedding for the prompt initialization, and then fix the textual prompts all the time or optimize the textual prompts during the training.}

{Specifically, we use a template of ``A person photo from the [X] re-identification scene'' and replace the ``[X]'' with ``general'', ``low-resolution'', ``clothing change'', ``occlusion'', and ``cross-modality''.
We input the texts into the CLIP text encoder, obtain the textual embeddings, and then use the embeddings to initialize the five groups of scene-specific prompts.
Then, we can conduct the VersReID training in two ways: freezing the textual prompts or optimizing them with the backbone.}

{The experimental results in Table~\ref{rt.4} show that in most scenes, the impact of different initialization on performance is not significant.
But on the clothing change scene, using CLIP textual embedding for prompt initialization performs relatively worse compared to other scenes.
For example, when using CLIP embedding for initialization and optimizing the prompts during training, ReID Bank only achieves 58.6\% Rank-1 accuracy on the PRCC, which is 1.4\% lower than that of using random initialization. 
This may be because the scene-specific prompts are used to guide the ReID Bank to handle images from different scenes.
However, since CLIP is trained to match a text and a single image, the text that describes the changes \textbf{between} images like ``clothing changing'' is out of the training scope of CLIP. Thus, CLIP cannot effectively provide guidance to the ReID model based on the text ``A person photo from the clothing change re-identification scene'' (which implies the clothing between images is changing), which limits its performance for prompt initialization. Hence, randomly initializing groups of learnable prompts can achieve promising results after optimization, and we do not adopt a text encoder like CLIP for extracting scene-specific prompts.}

{We also investigate using CLIP for zero-shot cross-scene ReID. Specifically, we employ the textual prompts from CLIP mentioned above as the scene-specific prompts. In our work, there are five ReID scenes. We regard the cross-modality scene as the testing scene and use the other four scenes for training.
We train our model in four scenes with the fixed textual prompts (``A person photo from the [X] re-identification scene'', where ``[X]'' $\in$ [``general'', ``low-resolution'', ``clothing change'', ``occlusion'']).
Then, we evaluate our model in the cross-modality scene with different prompts (four prompts in training scenes and one extra prompt in cross-modality scene). Unfortunately, compared with the prompts from the training scenes, using the cross-modality prompts damages the model in the cross-modality scene, which indicates that simply using textual prompts for zero-shot cross-scene ReID is unsuitable. We leave this for further future investigation. }

\renewcommand{\arraystretch}{1.3}
\begin{table*}[ht]
  \caption{{Comparisons between different prompt initialization strategies. The CLIP (F) and CLIP (T) denote the textual prompts output by the CLIP text encoder are frozen (F) and trainable~(T), respectively. Random prompt initialization is the default setting in the main manuscript, which performs slightly better than the CLIP initialization.}}
  \label{rt.4}
  \centering
  \resizebox{\linewidth}{!}{
  \begin{tabular}{c | c | c | c c | c c | c c | c c | c c | c c | c c | c c}
    \toprule
   &  & \multirow{2}{*}{Prompt} & \multicolumn{4}{c|}{General} & \multicolumn{2}{c|}{Low-resolution} & \multicolumn{4}{c|}{Clothing Change} & \multicolumn{2}{c|}{Occlusion} & \multicolumn{2}{c|}{Cross-modality} & \multicolumn{2}{c}{Joint}\\
    \cline{4-17}
    \makecell[c]{Model} & \makecell[c]{Scene \\ labels} & \multirow{2}{*}{Init.} & \multicolumn{2}{c|}{Market-1501} & \multicolumn{2}{c|}{MSMT17} & \multicolumn{2}{c|}{MLR-CUHK03} & \multicolumn{2}{c|}{Celeb-ReID} & \multicolumn{2}{c|}{PRCC} & \multicolumn{2}{c|}{Occ-Duke} & \multicolumn{2}{c|}{SYSU-mm01} & \multicolumn{2}{c}{Testing Set} \\
    \cline{4-19}
    & & & R-1 & mAP & R-1 & mAP & R-1 & mAP & R-1 & mAP & R-1 & mAP & R-1 & mAP & R-1 & mAP & R-1 & mAP \\
    \hline
     & &  Random & 96.4 & 92.5 & 87.9 & \textbf{72.5} & \textbf{96.3} & \textbf{97.6} & \textbf{61.9} & \textbf{17.3} & \textbf{60.0} & \textbf{71.2} & \textbf{71.6} & \textbf{63.9} & \textbf{68.6} & \textbf{68.5} & \textbf{78.7} & \textbf{66.2} \\
    ReID Bank & \Checkmark & CLIP (F) & \textbf{96.6} & \textbf{92.6} & \textbf{88.0} & \textbf{72.5} & 95.8 & 97.4 & 60.4 & 17.0 & 58.5 & 66.7 & 70.8 & 63.2 & 67.6 & 66.9 & 78.0 & 65.5 \\
    & & CLIP (T) & 96.5 & \textbf{92.6} & \textbf{88.0} & \textbf{72.5} & 96.0 & 97.4 & 60.1 & 17.0 & 58.6 & 66.7 & 71.3 & 63.2 & 67.8 & 67.0 & 78.1 & 65.5 \\
     \hline
     & & Random & \textbf{96.9} & \textbf{93.0} & 88.5 & \textbf{73.6} & \textbf{97.1} & \textbf{98.2} & \textbf{60.8} & \textbf{17.1} & \textbf{61.3} & \textbf{72.6} & \textbf{72.9} & \textbf{64.9} & \textbf{67.7} & 65.7 & \textbf{76.3} & \textbf{63.8} \\
    V-Branch & \XSolid & CLIP (F) & \textbf{96.9} & \textbf{93.0} & \textbf{88.6} & \textbf{73.6} & 96.9 & 97.9 & 59.9 & 17.0 & 59.3 & 67.3 & 71.9 & 64.4 & 67.0 & \textbf{65.8} & 75.5 & 63.2 \\
     & & CLIP (T) & \textbf{96.9} & 92.9 & \textbf{88.6} & \textbf{73.6} & 97.0 & 97.9 & 59.7 & 16.9 & 59.5 & 67.5 & 72.6 & 64.7 & 67.3 & 65.5 & 75.7 & 63.3 \\
    \bottomrule
  \end{tabular}
  }
\end{table*}

\renewcommand{\arraystretch}{1.3}
\begin{table*}[t]
  \caption{{The indispensability of scene-specific prompts for modeling scene-specific knowledge. Removing the scene-specific prompts leads to a significant performance drop. All the methods in this table do not use scene labels for evaluation.
  \colorbox{gray}{``Joint Testing Set V2''} is comprised of the testing sets from clothing change, cross-modality, and occlusion ReID datasets, which focuses on severe shape and texture variations while the original joint testing set contains additional general ReID datasets and low-resolution ReID dataset.}}
  \label{rt.5}
  \centering
  \resizebox{\linewidth}{!}{
  \begin{tabular}{c | c c | c c | c c | c c | c c | c c | c c | c c | c c}
    \toprule
    \makecell[c]{\multirow{3}{*}{Model}} & \multicolumn{4}{c|}{General} & \multicolumn{2}{c|}{Low-resolution} & \multicolumn{4}{c|}{Clothing Change} & \multicolumn{2}{c|}{Occlusion} & \multicolumn{2}{c|}{Cross-modality} & \multicolumn{2}{c|}{Joint} & \multicolumn{2}{c}{ \cellcolor{gray}Joint} \\
    \cline{2-19}
    & \multicolumn{2}{c|}{Market-1501} & \multicolumn{2}{c|}{MSMT17} & \multicolumn{2}{c|}{MLR-CUHK03} & \multicolumn{2}{c|}{Celeb-ReID} & \multicolumn{2}{c|}{PRCC} & \multicolumn{2}{c|}{Occ-Duke} & \multicolumn{2}{c|}{SYSU-mm01} & \multicolumn{2}{c|}{Testing Set} &   \multicolumn{2}{c}{ \cellcolor{gray}Testing Set V2} \\
    \cline{2-19}
    & R-1 & mAP & R-1 & mAP & R-1 & mAP & R-1 & mAP & R-1 & mAP & R-1 & mAP & R-1 & mAP & R-1 & mAP & \cellcolor{gray} R-1 & \cellcolor{gray} mAP\\
    \hline
    
    V-Branch (ours) & \textbf{96.9} & \textbf{93.0} & \textbf{88.5} & \textbf{73.6} & \textbf{97.1} & \textbf{98.2} & \textbf{60.8} & \textbf{17.1} & \textbf{61.3} & \textbf{72.6} & \textbf{72.9} & \textbf{64.9} & \textbf{67.7} & \textbf{65.7} & \textbf{76.3} & \textbf{63.8} &  \cellcolor{gray} \textbf{59.8} &  \cellcolor{gray} \textbf{46.1}  \\
    
    W/o prompts & 96.4 & 92.8 & 87.9 & 72.0 & 96.0 & 97.4 & 59.8 & 16.8 & 58.8 & 70.0 & 71.2 & 62.9 & 66.3 & 63.9 & 75.3 & 61.9 &  \cellcolor{gray} 55.7 &  \cellcolor{gray} 42.5 \\
    
    \bottomrule
  \end{tabular}
  }
\end{table*}

\subsection{{Advantages of scene-specific prompts}}
{We clarify that our scene-specific prompts have the following advantages. (i) Our scene-specific prompts learn the scene-specific knowledge from different scenes, which is crucial for multi-scene ReID because not all knowledge can be shared among different scenes. (ii) Our scene-specific prompts make our model more explainable. (iii) In addition to the performance gains, our scene-specific prompts are easy to implement without complicated design or training procedures. We also show that the scene-specific prompts harmonize the ReID learning under different scenes and stabilize the training process.
We detail the advantages as follows.}

\vspace{0.5\baselineskip}
{\noindent \textbf{- Regarding (i) learning scene-specific knowledge}, the ReID model should have the flexibility to adapt the changes of identity cues in the multi-scenes ReID.
For example, the clothing of identity is a useful cue in the general scene while the clothes tend to confuse the ReID model in the clothing change scene.}

\begin{figure*}[t]
  \centering
  \includegraphics[width=1\linewidth]{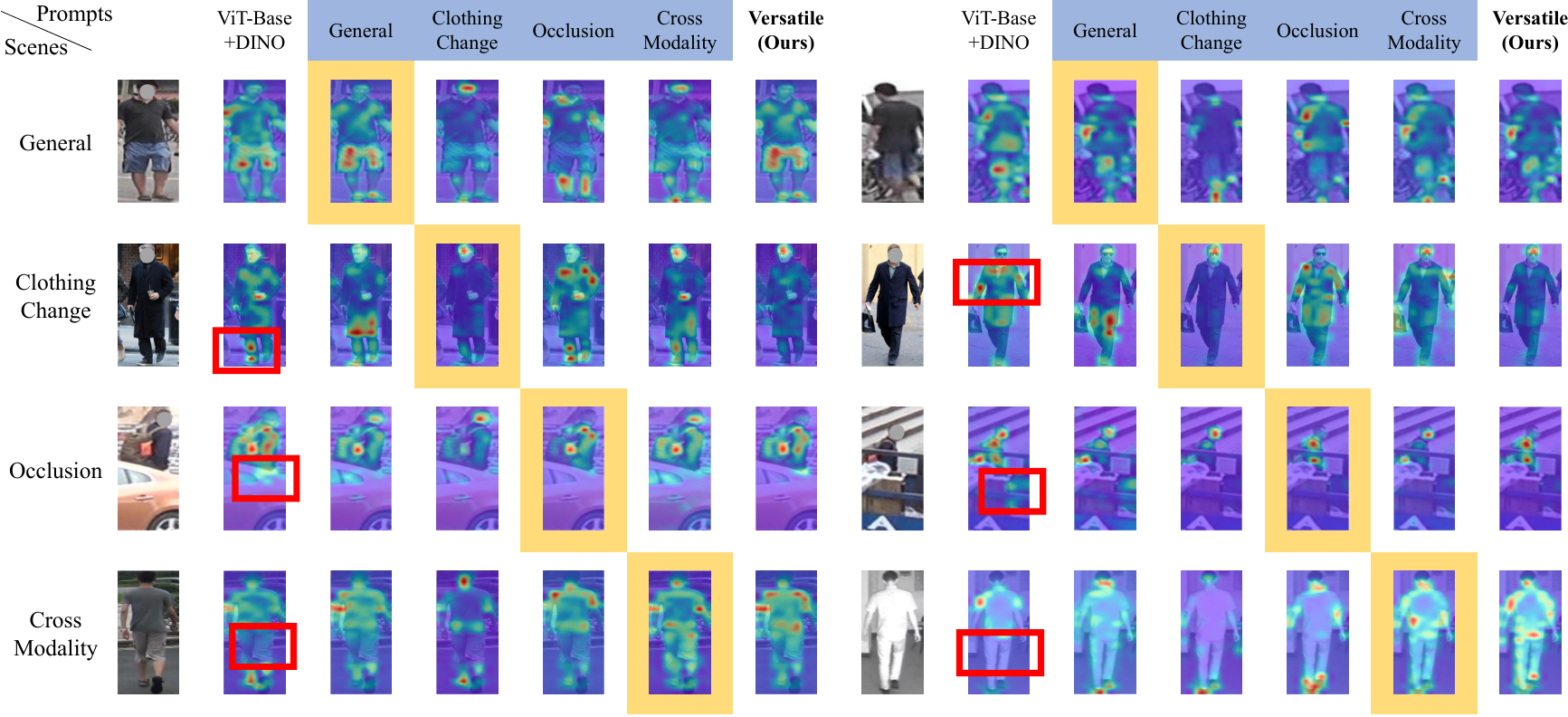}
  \caption{Visualizations of the self-attention map of the class token $e_{[\mathrm{CLS}]}$ in the last transformer block. Attention maps are the average value among all heads in the self-attention module. Each row represents two images of the same person in a specific scene and their attention maps. Each column represents applying the same scene-specific prompts to different input images. Prompts with \colorbox{blue}{blue} background: attention maps in this column are from the ReID Bank. Attention maps with \colorbox{lightyellow}{yellow} background: the scene and prompts correspond with each other.
  {We can observe that our V-Branch can adaptively focus on different image parts while the vanilla ViT-Base cannot as highlighted in red boxes.}}
  \label{f.6}
\end{figure*}

{Importantly, in our modeling, each group of scene-specific prompts in the ReID Bank is coupled with a kind of scene label so that each group can focus on learning the corresponding scene-specific knowledge, which can be observed from the attention maps (Figure 6 in the main manuscript and Figure~\ref{f.6} in response).
Differently, \textbf{only} learning a shared backbone on all images can merely model the scene-invariant knowledge and ignore the scene-specific knowledge, which leads to inferior performance.}

{We have conducted experiments to show the effectiveness of the scene-specific prompts by removing the specific prompts. 
The experimental results are shown in Table~\ref{rt.5}.
We can observe that, except for the general scene, removing the scene-specific prompts leads to a clear performance drop.
For example, the Rank-1 accuracy in MLR-CUHK03, PRCC, Occ-Duke, and SYSU-mm01 drops by 1.1\%, 2.5\%, 1.7\% and 1.4\%, respectively.
The performance drops on different scenes are because the model cannot effectively capture the scene-specific knowledge without the scene-specific prompts.
Especially, the Rank-1 accuracy and mAP in the joint testing set increased by 1.0\% and 1.9\% in the presence of scene-specific prompts, respectively.
Due to the large number of identities in the joint testing set,  +1.0\% Rank-1 accuracy means accurately retrieving about 288 query images from approximately 138,000 gallery images from various scenes (occlusion, low-resolution, and so forth) and +1.9\% mAP means a number of ground-truth matches are ranked properly, which is a clear performance improvement demonstrating the effectiveness of the scene-specific prompts. }

{To further demonstrate the importance of scene-specific modeling, we organized the ``Joint Testing Set V2'', which is comprised of the data from clothing change, occlusion, and cross-modality scenes. The images in these scenes exhibit severe shape and texture variations.
As shown in Table~\ref{rt.5}, we observe that learning scene-specific prompts brings clear improvement. 
Specifically, without scene-specific prompts, we can only achieve 55.7\% and 42.5\% mAP in the joint testing set V2, which is 4.1\% and 3.6\% lower than our full model.}

\begin{figure*}[t]
  \centering
  \includegraphics[width=1\linewidth]{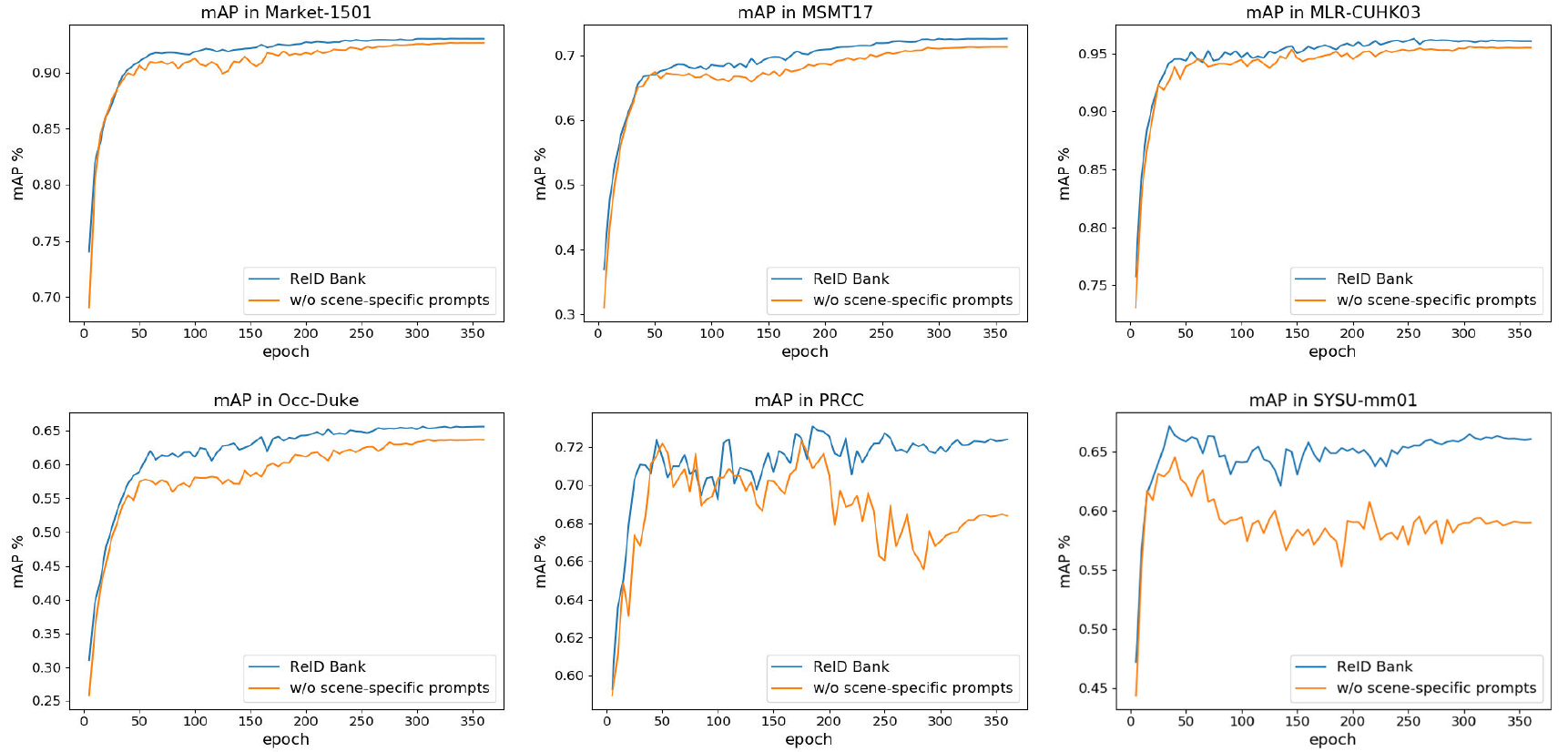}
  \caption{{Performance curves in six datasets. We can observe that the scene-specific prompts help the model to overcome the shape and texture variations in ReID, which commonly exist in the clothing-change ReID, occlusion ReID, and cross-modality ReID. }}
  \label{f.7}
\end{figure*}

\vspace{0.5\baselineskip}
{\noindent \textbf{- Regarding (ii) the explanability}, by visualizing the feature maps from the ReID Bank with different scene-specific prompts, we can understand the scene-specific knowledge modeled by our method. Thus, our model is more explainable.
We provide visualization results to qualitatively verify the effectiveness of our scene-specific prompts in our manuscript.
We show the visualization results in Figure~\ref{f.6}.
From the figure, we can show that using different scene-specific prompts in ReID Bank leads to different attention maps on images.
Specifically, when using cloth-changing prompts, our ReID Bank focuses on the face as well as shape, and when using occlusion prompts, ReID Bank concentrates on visible parts. Especially, by comparing the attention map of the V-Branch and the corresponding attention map from ReID Back with correct scene-specific prompts, we observe that our V-Branch can adaptively extract ReID features from different scenes, which qualitatively verifies the effectiveness of our method in unifying knowledge from different scenes. 
In contrast, the vanilla ViT-Base model with DINO pre-training struggles to handle images from various scenes.
As highlighted in the red box, the vanilla ViT-Base model fails to capture shape information in cross-modality scene and is disturbed by the background or occlusion part. Besides, the vanilla model pays considerable attention to the clothes when handling clothing change ReID.
Overall, we can conclude that our scene-specific prompts can effectively learn scene-specific knowledge \yx{with explainability} and are crucial for multi-scene ReID.}

\vspace{0.5\baselineskip}
{\noindent \textbf{- Regarding (iii) the optimization}, our scene-specific prompts work as additional informative tokens in the transformer backbone, which do not bring complicated training procedures and are easy to optimize. 
We show the performance curves in Figure~\ref{f.7}. 
From the figure, we can observe that the scene-specific prompts help the model handle ReID under different scenes and do not slow down the convergence speed of the model. Besides, we find that learning scene-specific prompts helps to stabilize the training process.
As shown in the figure, in the absence of scene-specific prompts, the mAP in the PRCC dataset increases first but turns to decrease after a period of optimization process. 
It is mainly because of the difficulty of harmonizing a scene-shared backbone for ReID under different scenes and the potential overfitting.
Differently, applying scene-specific prompts can alleviate this problem and the performance becomes more stable during the optimization process.}

\section{Detailed Experimental Results on the V-Branch}
In this section, we provide the comprehensive results of the ablation studies on the V-Branch including the number of versatile prompts ($M$), the distillation loss function $\mathcal{L}_{KD}$ and its weight $\alpha$. The experimental results are shown in Table~\ref{t.a2}.

\vspace{0.5\baselineskip}\noindent\textbf{- Distillation loss.}
We study the impact of different distillation loss functions in the second part of Table~\ref{t.a2}.
We investigate four typical and effective distillation methods as follows. 
The vanilla distillation loss $\mathcal{L}_{KL}$~\cite{kd} guides the student model to produce similar logits as the teacher model by optimizing the KL divergence between the logits from the two models.
The relational knowledge distillation loss $\mathcal{L}_{RKD}$~\cite{rkd} enforces the relations between samples are consistent in the feature spaces of both the teacher and student model. 
Another two losses $\mathcal{L}_{1}, \mathcal{L}_2$ constrain the student model to extract similar features with the teacher model by minimizing the $\ell_1$ and $\ell_2$ distance between features from the two models, respectively.

{Formally, assuming there are two images $\mathbf{x}_i$ and $\mathbf{x}_j$, the ReID Bank extract features $\boldsymbol{f}_i$ and $\boldsymbol{f}_j$ for them, respectively, and denote the features from the V-Branch as $\boldsymbol{f}'_i$ and $\boldsymbol{f}'_j$.
Denote the identity prediction probabilities of image $\mathbf{x}_i$ from ReID Bank and V-Branch as $\boldsymbol{p}_i$ and $\boldsymbol{p}'_i$.
Let $KLDiv(.)$ represent the KL divergence. We have:}
\begin{equation}
\mathcal{L}_1 = \sum_i  \Vert  \boldsymbol{f}'_i -\boldsymbol{f}_i \Vert_1,
\end{equation}
\begin{equation}
\mathcal{L}_2 = \sum_i  \Vert  \boldsymbol{f}'_i -\boldsymbol{f}_i \Vert_2,
\end{equation}
\begin{equation}
\mathcal{L}_{KL} = \sum_i  KLDiv(\boldsymbol{p}_i, \boldsymbol{p}'_i),
\end{equation}
\begin{equation}
\mathcal{L}_{RKD} = \sum_{i,j}  \Vert \boldsymbol{d}'_{i,j} - \boldsymbol{d}_{i,j} \Vert_2^2, \textbf{where } \boldsymbol{d}_{i,j} = \Vert \boldsymbol{f}_i -\boldsymbol{f}_j \Vert_2^2, \boldsymbol{d}'_{i,j} =  \Vert  \boldsymbol{f}'_i -\boldsymbol{f}'_j \Vert_2^2.
\end{equation}

The results in Table~\ref{t.a2} show that different distillation loss functions have little difference in the performance in each scene. 
Among the compared loss functions, the relational knowledge distillation loss $\mathcal{L}_{RKD}$ performs best on the joint testing set because the knowledge in the form of relations between samples is distilled, and such knowledge is important for Re-ID since Re-ID is a ranking task.
Thus, the $\mathcal{L}_{RKD}$ is chosen to be the distillation objective function $\mathcal{L}_{kd}$ in the main experiments.

\begin{table*}[t]
  \caption{Ablation studies on the number of versatile prompts $M$, the distillation loss, and the loss weight $\alpha$ in the scene-specific prompts distillation stage for training the \textbf{V-Branch}.  The ``Num. Versatile Prompt'' represents the number of versatile prompts in the V-Branch. The ``Dis. Loss'' denotes the distillation objective functions. The cell with \colorbox{lightgray}{gray} background in the table is the default setting of the proposed method. By default, we instantiate our distillation function $\mathcal{L}_{kd}$ as $\mathcal{L}_{RKD}$ and use loss weight $\alpha=1$.}
  \label{t.a2}
  \centering
  \resizebox{\linewidth}{!}{
  \begin{tabular}{c | c | c | c c | c c | c c | c c | c c | c c | c c | c c }
    \toprule
    Num. & \multirow{2}{*}{Dis.} & Loss & \multicolumn{4}{c|}{General} & \multicolumn{2}{c|}{Low-resolution} & \multicolumn{4}{c|}{Clothing Change} & \multicolumn{2}{c|}{Occlusion} & \multicolumn{2}{c|}{Cross-modality} & \multicolumn{2}{c}{Joint} \\
    \cline{4-17}
    Versatile & \multirow{2}{*}{Loss} & Weight & \multicolumn{2}{c|}{Market-1501} & \multicolumn{2}{c|}{MSMT17} & \multicolumn{2}{c|}{MLR-CUHK03} & \multicolumn{2}{c|}{Celeb-ReID} & \multicolumn{2}{c|}{PRCC} & \multicolumn{2}{c|}{Occ-Duke} & \multicolumn{2}{c|}{SYSU-mm01} & \multicolumn{2}{c}{Testing Set} \\
    \cline{4-19}
    Prompt &  & $\alpha$ & R-1 & mAP & R-1 & mAP & R-1 & mAP & R-1 & mAP & R-1 & mAP & R-1 & mAP & R-1 & mAP & R-1 & mAP \\
    \hline
    0 & \multirow{5}{*}{$\mathcal{L}_{RKD}$} & \multirow{5}{*}{1.0} & 96.5 & 92.5 & 88.1 & 72.5 & 96.5 & 96.9 & 60.6 & 17.0 & 60.0 & 70.5 & 72.4 & 63.7 & 67.0 & 64.9 & 75.5 & 62.6 \\
    \cellcolor{lightgray}5 &  &  & 96.9 & 93.0 & 88.5 & 73.6 & 97.1 & 98.2 & 60.8 & 17.1 & 61.3 & 72.6 & 72.9 & 64.9 & 67.7 & 65.7 & 76.3 & 63.8 \\   
    10 &  &  & 96.4 & 92.9 & 88.6 & 73.8 & 97.2 & 98.2 & 61.3 & 17.1 & 60.8 & 71.5 & 72.8 & 64.5 & 67.9 & 65.8 & 76.3 & 63.7 \\
    20 &  &  & 96.7 & 92.9 & 88.6 & 73.8 & 96.9 & 98.0 & 60.9 & 17.2 & 61.4 & 72.0 & 72.7 & 64.6 & 67.9 & 65.5 & 76.2 & 63.6 \\
    30 &  &  & 96.8 & 93.0 & 88.5 & 73.7 & 97.1 & 98.2 & 60.6 & 17.2 & 60.9 & 71.4 & 72.9 & 64.7 & 67.6 & 65.4 & 76.3 & 63.7 \\
    \hline \hline
    \multirow{4}{*}{5} & \cellcolor{lightgray}$\mathcal{L}_{RKD}$ & \multirow{4}{*}{1.0} & 96.9 & 93.0 & 88.5 & 73.6 & 97.1 & 98.2 & 60.8 & 17.1 & 61.3 & 72.6 & 72.9 & 64.9 & 67.7 & 65.7 & 76.3 & 63.8 \\    
    & $\mathcal{L}_1$ &  & 96.5 & 92.8 & 88.6 & 73.8 & 96.9 & 98.1 & 61.1 & 17.4 & 60.9 & 71.4 & 72.5 & 64.6 & 67.4 & 66.6 & 76.1 & 63.4 \\
    & $\mathcal{L}_2$ &  & 96.5 & 92.9 & 88.5 & 73.7 & 97.3 & 98.2 & 60.8 & 17.3 & 61.0 & 71.9 & 72.5 & 65.2 & 67.6 & 66.0 & 76.1 & 63.5 \\
    & $\mathcal{L}_{KL}$ &  & 96.7 & 93.0 & 88.8 & 73.9 & 96.8 & 97.9 & 60.5 & 17.3 & 60.4 & 71.3 & 73.3 & 64.5 & 67.1 & 65.7 & 76.0 & 63.5 \\    
    \hline \hline
    \multirow{8}{*}{5} & \multirow{8}{*}{$\mathcal{L}_{RKD}$} & 0.0 & 96.6 & 92.6 & 88.1 & 72.6 & 96.7 & 97.8 & 59.3 & 16.4 & 60.2 & 71.0 & 72.3 & 63.8 & 66.7 & 63.1 & 75.4 & 62.7 \\
    &  & 0.5 & 97.0 & 93.1 & 88.6 & 73.5 & 97.0 & 98.0 & 60.7 & 17.1 & 62.2 & 73.2 & 72.7 & 64.9 & 67.5 & 65.6 & 76.3 & 63.8 \\
    &  & \cellcolor{lightgray}1.0 & 96.9 & 93.0 & 88.5 & 73.6 & 97.1 & 98.2 & 60.8 & 17.1 & 61.3 & 72.6 & 72.9 & 64.9 & 67.7 & 65.7 & 76.3 & 63.8 \\
    &  & 2.0 & 97.0 & 93.0 & 88.5 & 73.5 & 97.0 & 98.0 & 60.7 & 17.1 & 61.1 & 72.3 & 72.9 & 64.8 & 67.8 & 64.3 & 76.3 & 63.7 \\
    &  & 3.0 & 97.0 & 93.0 & 88.5 & 73.5 & 97.1 & 98.2 & 61.1 & 17.1 & 60.7 & 71.7 & 73.0 & 64.9 & 67.8 & 63.8 & 76.3 & 63.5 \\
    &  & 4.0 & 96.8 & 93.0 & 88.6 & 73.6 & 97.6 & 98.5 & 60.8 & 17.2 & 59.6 & 70.6 & 72.9 & 64.8 & 67.6 & 62.1 & 76.2 & 63.4 \\
    &  & 5.0 & 96.7 & 93.0 & 88.5 & 73.8 & 96.4 & 97.7 & 60.8 & 17.2 & 58.8 & 69.4 & 72.4 & 64.5 & 66.4 & 62.6 & 75.8 & 63.2 \\
    &  & 10.0 & 96.6 & 92.9 & 88.7 & 73.7 & 96.5 & 97.7 & 60.5 & 17.3 & 54.0 & 66.3 & 72.3 & 64.7 & 66.0 & 61.9 & 75.2 & 62.2 \\
    \bottomrule
  \end{tabular}
  }
\end{table*}

\section{Detailed Experimental Results on the MPDA}
In this section, we show the comprehensive results of the ablation studies on the MPDA, which are shown in Table~\ref{t.a3}.
To simplify the notation, we denote the clothing change, gray-scale, occlusion, blurred, and lighting change views as CC, GS, OC, BL, and LC, respectively.

\begin{table*}[ht] 
  \caption{Ablation studies on \textbf{MPDA}. We study how the augmented views affect the transfer performance and if MPDA can generalize to another SSL method, Mocov3~\cite{mocov3}. The used model is V-Branch with the ViT-S backbone. \textbf{LC: lighting change view, BL: blurred view, CC: clothing change view, OC: occlusion view, GS: gray-scale view. Row (7) is the Mocov3 baseline which applies two randomly augmented views. Row (8) presents the results of Mocov3 + MPDA.} The cell with \colorbox{lightgray}{gray} background in the table is the default setting of the proposed method. }
  \label{t.a3}
  \centering
  \resizebox{\linewidth}{!}{
  \begin{tabular}{c | c | c c c c c | c c | c c | c c | c c | c c | c c | c c | c c }
     \toprule
    \multirow{3}{*}{Index} & Self-supervised  & \multicolumn{5}{c|}{Augmented views} & \multicolumn{4}{c|}{General} & \multicolumn{2}{c|}{Low-resolution} & \multicolumn{4}{c|}{Clothing Change} & \multicolumn{2}{c|}{Occlusion} & \multicolumn{2}{c|}{Cross-modality} & \multicolumn{2}{c}{Joint} \\
     \cline{3-21}
     & Learning & \multicolumn{1}{c|}{L} & \multicolumn{1}{c|}{B} & \multicolumn{1}{c|}{C} & \multicolumn{1}{c|}{O} & \multicolumn{1}{c|}{G} & \multicolumn{2}{c|}{Market-1501} & \multicolumn{2}{c|}{MSMT17} & \multicolumn{2}{c|}{MLR-CUHK03} & \multicolumn{2}{c|}{Celeb-ReID} & \multicolumn{2}{c|}{PRCC} & \multicolumn{2}{c|}{Occ-Duke} & \multicolumn{2}{c|}{SYSU-mm01} & \multicolumn{2}{c}{testing set} \\
    \cline{8-23}
     & Method & \multicolumn{1}{c|}{C} & \multicolumn{1}{c|}{L} & \multicolumn{1}{c|}{C} & \multicolumn{1}{c|}{C} & \multicolumn{1}{c|}{S} & R-1 & mAP & R-1 & mAP & R-1 & mAP & R-1 & mAP & R-1 & mAP & R-1 & mAP & R-1 & mAP & R-1 & mAP \\
    \hline
    \cellcolor{lightgray}1 &  & \ding{51} & \ding{51} & \ding{51} & \ding{51} & \ding{51} & 90.5 & 79.5 & 63.2 & 41.2 & 89.4 & 90.5 & 45.8 & 7.5 & 35.1 & 49.5 & 46.5 & 37.7 & 35.8 & 38.6 & 56.2 & 40.8 \\
    2 &  &  & \ding{51} & \ding{51} & \ding{51} & \ding{51} & 90.4 & 78.4 & 62.1 & 39.6 & 88.6 & 89.9 & 45.5 & 7.3 & 34.7 & 48.9 & 45.0 & 35.8 & 31.7 & 34.4 & 55.5 & 39.4 \\
    3 & DINO & \ding{51} &  & \ding{51} & \ding{51} & \ding{51} & 89.9 & 78.2 & 61.0 & 38.4 & 86.1 & 88.8 & 43.9 & 7.2 & 30.7 & 46.0 & 44.4 & 35.3 & 31.4 & 33.8 & 54.1 & 38.3 \\
    4 & \cite{dino} & \ding{51} & \ding{51} &  & \ding{51} & \ding{51} & 90.8 & 79.7 & 63.6 & 41.5 & 89.2 & 90.6 & 45.6 & 7.3 & 32.9 & 48.7 & 45.7 & 37.2 & 34.6 & 37.3 & 55.9 & 40.7 \\
    5 &  & \ding{51} & \ding{51} & \ding{51} &  & \ding{51} & 90.3 & 78.2 & 58.6 & 37.0 & 88.6 & 89.6 & 43.7 & 6.8 & 33.3 & 47.8 & 42.4 & 33.7 & 34.5 & 36.8 & 54.0 & 37.4 \\
    6 &  & \ding{51} & \ding{51} & \ding{51} & \ding{51} &  & 90.3 & 79.4 & 63.3 & 41.3 & 89.5 & 90.7 & 45.9 & 7.5 & 34.8 & 49.1 & 46.4 & 37.5 & 35.0 & 37.8 & 56.0 & 40.6 \\
    \hline \hline
    7 & Mocov3 &  &  &  &  &  & 90.4 & 77.0 & 62.6 & 39.0 & 90.7 & 91.5 & 48.2 & 8.1 & 43.8 & 57.8 & 43.0 & 33.9 & 37.6 & 39.5 & 57.2 & 40.4 \\
    8 & \cite{mocov3} & \ding{51} & \ding{51} & \ding{51} & \ding{51} & \ding{51} & 92.3 & 81.8 & 70.6 & 47.4 & 92.2 & 93.1 & 50.1 & 9.1 & 42.2 & 57.1 & 50.8 & 41.7 & 45.4 & 47.5 & 61.9 & 45.6 \\
    \bottomrule
\end{tabular}
  }
\end{table*}

\vspace{0.5\baselineskip}\noindent\textbf{- Effectiveness of each augmented view.}
In general, based on the results, it can be observed that the lack of certain augmented views has a negative impact on the performance of the corresponding scene. 
For instance, when the blurred view is removed, the Rank-1 accuracy on the low-resolution dataset MLR-CUHK03~\cite{dslr_market} degrades -3.3\%, as shown by the comparison of Row (1) and (3) in Table~\ref{t.a3}. 
Similarly, when other views such as clothing change, occlusion, and gray-scale views are removed, the model's performances on the clothing change dataset PRCC~\cite{prcc}, occlusion dataset Occ-Duke~\cite{pgfa}, cross-modality dataset SYSU-mm01~\cite{mm01} also decreased accordingly. 
The above results clearly show the impact of each augmented view, where the performance in related scenes will be affected by its corresponding image views, even if these views are used for self-supervised learning.
These phenomena also suggest the necessity of introducing downstream-related properties to self-supervised learning, which is helpful for the transfer ability of models.

{We further analyze the impact of the augmented views in each dataset.
Regarding the general scene, the results show that, apart from the lighting change view, the other views including the blurred view, occlusion view, also have a significant impact on MSMT17~\cite{msmt}, which may suggest that low-resolution and occlusion also exist in the MSMT17 dataset. 
This is reasonable because MSMT is a general ReID dataset at a large scale, containing abundant variations.}

{Regarding the low-resolution scene, the employment of blur augmentation significantly improves the performance. In the MLR-CUHK03 datasets, removing blur augmentation leads to a 3.3\% performance drop in Rank-1 accuracy while the other data augmentation affects the performance less. This is mainly because the blur augmentation imitates the low-resolution scene.}

{Regarding the clothing-change scene, the blur augmentation, clothing-change augmentation, and occlusion augmentation are helpful and will cause 4.4\%, 2.2\%, and 1.8\% Rank-1 accuracy drops in PRCC dataset if removed. 
While the clothing-change augmentation directly imitates the clothing-change scene, the blur augmentation and occlusion augmentation destroy the information in clothes and force the model to extract identity cues from other ways instead of clothes.  Besides, it also indicates that the PRCC and Celeb-ReID datasets contain low-resolution and occlusion challenges.
Moreover, although clothing-change augmentation works effectively in PRCC dataset, it affects the performance in Celeb-ReID less. It may be because the clothing-change augmentation is conducted upon the generative adversarial model which can be improved in terms of stability.}

{Regarding the occlusion scene, the occlusion augmentation affects the performance most as removing the occlusion views of the images leads to a 4.1\% performance drop in Rank-1 accuracy in Occ-Duke dataset. The blur augmentation also helps the model in the occlusion scene. This may be because the blurred images force the model to extract fine-grained feature representation, which is conducive to ReID in the occlusion scene.}

{Regarding the cross-modality scene, the lighting change augmentation and the blur augmentation contribute remarkably and will cause 4.1\% and 4.4\% drops at Rank-1 accuracy in SYSU-mm01 dataset if being removed.
This is mainly because the lighting change augmentation reveals the dramatic modality gap between RGB and infrared images to some extent. Besides, on the one hand, the blur augmentation forces the model to extract fine-grained feature representation, which is conducive to ReID in the cross-modality scene. On the other hand, it may also indicate that the SYSU-mm01 dataset contains a low-resolution challenge, and hence the blur augmentation affects the performance remarkably. }

\vspace{0.5\baselineskip}\noindent\textbf{- Applying MPDA to Mocov3.}
We report more results of applying our MPDA to the Mocov3 in Table~\ref{t.a3} (7) and (8), respectively. 
Our results show that Mocov3 + MPDA outperforms the Mocov3 baseline with random data augmentations by +4.7\% Rank-1 accuracy on the joint testing set.
Moreover, MPDA improves the Rank-1 accuracy in specific scenes.
For example, the Rank-1 accuracy in occlusion and cross-modality scenes are improved to 50.8\% and 45.4\%, respectively, which are both +7.8\% higher than the baseline. 
These results demonstrate the robustness of MPDA, which can be used with other contrastive learning methods and enhance the ReID models in multi-scene conditions.

\section{{Comparisons with different ensemble methods}} 
{To solve multi-scene ReID, a native solution is to train multiple single-scene models and then conduct the ensemble of models. We conduct ensemble experiments to show the superiority of our method to the ensemble method. }

\vspace{0.5\baselineskip}
\noindent {\textbf{- Experiments setup.} 
We first train two different scene classifiers on the multi-scene training set, including a ResNet-18~\cite{resnet} and a ViT-Tiny~\cite{vit}.
The ResNet-18 and ViT-Tiny achieve 93.6\% and 96.3\% scene classification accuracy on the joint testing set, respectively.
For fair comparisons, since we use ViT-Base as the backbone of our VersReID and do not use any auxiliary information, we train several ViT-Base models on different scenes separately as the scene-specific models. 
The training settings are the same as training the VersReID, and these scene-specific models are also pre-trained on LUP-B dataset using our proposed MPDA.
We denote these models as scene-specific ViT-Base (MPDA).
We try our best to adjust the hyperparameters to improve the performance of scene-specific ViT-Base models in specific scenes.
For testing on multi-scene ReID, we use hard and soft ways to ensemble the specific models. 
Given an image, the hard ensemble uses the scene prediction results to determine which specific model is applied.
The soft ensemble weights and sums the features of different specific models according to the scene prediction probability.}

\vspace{0.5\baselineskip}
\noindent {\textbf{- Experiments on Ensemble of scene-specific models.}
As shown in Table~\ref{rt.1.A} and Table~\ref{rt.1.B}, the performance of ensemble multiple scene-specific models is inferior to our V-Branch no matter whether using ResNet-18 or ViT-Tiny as the scene classifier. 
Specifically, when using ViT-Tiny as the scene classifier, the best performance of the ensemble model on the joint testing set is 72.6\% Rank-1 accuracy (hard ensemble using ViT-Tiny scene classifier, in Table~\ref{rt.1.B}), which is 3.7\% lower than our V-Branch. 
Besides, we find that the ensemble performance is sensitive to the scene-classifier. When using ResNet-18 as the scene classifier rather than ViT-Tiny, the scene classification accuracy is slightly degraded from 96.3\% to 93.6\%. However, the ensemble model suffers from severe performance degradation and only achieves 39.7\% Rank-1 accuracy (hard ensemble using ResNet-18 scene classifier, in Table~\ref{rt.1.A}), which is 32.9\% lower than that of using ViT-Tiny as scene classifier.}

{Moreover, assuming a scene classifier with 100\% classification accuracy is available (equivalent to knowing scene labels during testing),  we can accurately employ the corresponding scene-specific models to process images from different scenes.
With the scene label, the ensemble method is expected to perform well (the row with\colorbox{myblue}{gray}background in each table).
However, our V-Branch still outperforms the ensemble model in most of the time without using the scene labels.
For example, our V-Branch achieves 67.7\% Rank-1 accuracy on SYSU-mm01 dataset without using scene labels, which is 3.1\% higher than the scene-specific ViT-Base (MPDA) model that uses scene labels.
The inferior performance of scene-specific models is mainly because the scene-shared knowledge from different scenes is ignored.
In contrast, our V-Branch effectively unifies knowledge (both scene-shared and scene-specific) from different scenes.
As a result, our method is adaptive to multi-scene ReID and achieves superior performance without requiring scene labels in inference.}

\renewcommand{\arraystretch}{1.3}
\begin{table*}[tb]
  \caption{
  {The ensemble results of scene-specific ViT-Base (MPDA) models using \textbf{ResNet-18} (93.6\% scene classification accuracy) as the scene classifier.
  Scene-specific ViT-Base (MPDA) models are trained on different scenes separately.
\textbf{\colorbox{myblue}{Ensemble with scene labels} means the scene labels are used in inference to apply the corresponding ViT-Base model to handle images from different scenes and ignore the scene classifier.}
  Otherwise, the single-scene models are ensemble in a hard or soft way according to the scene classifier. 
  Given an image, the scene classifier predicts the probability of the scenes.
 \textbf{The hard ensemble strategy selects the feature of the specific model from the scene with maximum probability, and the soft ensemble strategy weights and sums the features from all scene-specific models according to the scene classification probability.}
  Notably, our V-Branch is evaluated without requiring scene labels and achieves superior performance, demonstrating the effectiveness of unifying knowledge from different scenes.
  Our V-Branch can be further enhanced by overlapping patch embedding layer, denoted as V-Branch*.
  }}
  \label{rt.1.A}
  \centering
  \resizebox{\linewidth}{!}{
  \begin{tabular}{l | c | c c | c c | c c | c c | c c | c c | c c | c c}
    \toprule
    \makecell[c]{\multirow{3}{*}{Methods}} & \multirowcell{3}{Scene \\ labels} & \multicolumn{4}{c|}{General} & \multicolumn{2}{c|}{Low-resolution} & \multicolumn{4}{c|}{Clothing Change} & \multicolumn{2}{c|}{Occlusion} & \multicolumn{2}{c|}{Cross-modality} & \multicolumn{2}{c}{Joint}\\
    \cline{3-16}
    & & \multicolumn{2}{c|}{Market-1501} & \multicolumn{2}{c|}{MSMT17} & \multicolumn{2}{c|}{MLR-CUHK03} & \multicolumn{2}{c|}{Celeb-ReID} & \multicolumn{2}{c|}{PRCC} & \multicolumn{2}{c|}{Occ-Duke} & \multicolumn{2}{c|}{SYSU-mm01} & \multicolumn{2}{c}{Testing Set} \\
    \cline{3-18}
    & & R-1 & mAP & R-1 & mAP & R-1 & mAP & R-1 & mAP & R-1 & mAP & R-1 & mAP & R-1 & mAP & R-1 & mAP \\
    \hline
     \multicolumn{16}{c}{Different ensembles of \textbf{scene-specific ViT-Base (MPDA) model} using ResNet-18 as the scene classifier. } \\
     
    \textbf{Hard} Ensemble & \XSolid & 73.7 & 42.6 & 43.9 & 14.5 & 36.4 & 28.1 & 41.0 & 4.9 & 35.0 & 44.1 & 30.1 & 15.7 & 43.9 & 27.0 & 39.7 & 16.7 \\
    
    \textbf{Soft} Ensemble & \XSolid & 68.6 & 32.5 & 39.3 & 9.9 & 33.9 & 23.4 & 35.3 & 3.7 & 28.7 & 38.8 & 27.3 & 11.0 & 40.2 & 30.4 & 36.2 & 11.8 \\
    
    \cellcolor{myblue} Ensemble with Scene Labels & \cellcolor{myblue}\Checkmark  & \cellcolor{myblue}96.1 & \cellcolor{myblue}91.9 & \cellcolor{myblue}88.6 & \cellcolor{myblue}74.1 & \cellcolor{myblue}95.8 & \cellcolor{myblue}94.9 & \cellcolor{myblue}61.6 & \cellcolor{myblue}17.2 & \cellcolor{myblue}54.4 & \cellcolor{myblue}66.6 & \cellcolor{myblue}72.7 & \cellcolor{myblue}64.0 & \cellcolor{myblue}64.6 & \cellcolor{myblue}63.5 & \cellcolor{myblue}77.2 & \cellcolor{myblue}64.6 \\
    \hline \bottomrule
    \textbf{V-Branch (ours)} & \XSolid & 96.9 & 93.0 & 88.5 & 73.6 & 97.1 & 98.2 & 60.8 & 17.1 & \textbf{61.3} & \textbf{72.6} & 72.9 & 64.9 & 67.7 & 65.7 & 76.3 & 63.8 \\
    \textbf{V-Branch* (ours)} & \XSolid & \textbf{96.8} & \textbf{93.2} & \textbf{88.8} & \textbf{74.2} & \textbf{97.5} & \textbf{98.4} & \textbf{61.7} & \textbf{18.7} & 60.7 & 71.4 & \textbf{75.2} & \textbf{66.1} & \textbf{69.3} & \textbf{66.9} & \textbf{77.2} & \textbf{64.7} \\
    \bottomrule
  \end{tabular}
  }
\end{table*}

\renewcommand{\arraystretch}{1.3}
\begin{table*}[ht]
  \caption{{The ensemble results of scene-specific ViT-Base (MPDA) models using \textbf{ViT-Tiny} (96.3\% scene classification accuracy) as the scene classifier. 
  Scene-specific ViT-Base (MPDA) models are trained on different scenes separately.
\textbf{\colorbox{myblue}{Ensemble with scene labels} means the scene labels are used in inference to apply the corresponding ViT-Base model to handle images from different scenes and ignore the scene classifier.}
  Otherwise, the single-scene models are ensemble in a hard or soft way according to the scene classifier. 
    Given an image, the scene classifier predicts the probability of the scenes.
  \textbf{The hard ensemble strategy selects the feature of the specific model from the scene with maximum probability, and the soft ensemble strategy weights and sums the features from all scene-specific models according to the scene classification probability.}
  Notably, our V-Branch is evaluated without requiring scene labels and achieves superior performance, demonstrating the effectiveness of unifying knowledge from different scenes.
  Our V-Branch can be further enhanced by overlapping patch embedding layer, denoted as V-Branch*.}}
  \label{rt.1.B}
  \centering
  \resizebox{\linewidth}{!}{
  \begin{tabular}{l | c | c c | c c | c c | c c | c c | c c | c c | c c}
    \toprule
    \makecell[c]{\multirow{3}{*}{Methods}} & \multirowcell{3}{Scene \\ labels} & \multicolumn{4}{c|}{General} & \multicolumn{2}{c|}{Low-resolution} & \multicolumn{4}{c|}{Clothing Change} & \multicolumn{2}{c|}{Occlusion} & \multicolumn{2}{c|}{Cross-modality} & \multicolumn{2}{c}{Joint}\\
    \cline{3-16}
    &  & \multicolumn{2}{c|}{Market-1501} & \multicolumn{2}{c|}{MSMT17} & \multicolumn{2}{c|}{MLR-CUHK03} & \multicolumn{2}{c|}{Celeb-ReID} & \multicolumn{2}{c|}{PRCC} & \multicolumn{2}{c|}{Occ-Duke} & \multicolumn{2}{c|}{SYSU-mm01} & \multicolumn{2}{c}{Testing Set} \\
    \cline{3-18}
    &  & R-1 & mAP & R-1 & mAP & R-1 & mAP & R-1 & mAP & R-1 & mAP & R-1 & mAP & R-1 & mAP & R-1 & mAP \\
    \hline

     \multicolumn{16}{c}{Different ensembles of \textbf{scene-specific ViT-Base (MPDA) model} using ViT-Tiny as the scene classifier. } \\
     
    \textbf{Hard} Ensemble & \XSolid & 92.4 & 82.5 & 87.9 & 72.7 & 93.0 & 92.9 & 61.5 & 17.1 & 41.4 & 51.3 & 72.5 & 63.8 & 63.3 & 51.7 & 72.6 & 57.8 \\

 \textbf{Soft} Ensemble & \XSolid & 91.8 & 81.7 & 87.9 & 72.5 & 93.3 & 93.3 & 61.4 & 17.1 & 35.1 & 44.1 & 72.5 & 63.7 & 63.3 & 49.9 & 72.2 & 56.8 \\
 
    \cellcolor{myblue} Ensemble with Scene Labels & \cellcolor{myblue}\Checkmark & \cellcolor{myblue}96.1 & \cellcolor{myblue}91.9 & \cellcolor{myblue}88.6 & \cellcolor{myblue}74.1 & \cellcolor{myblue}95.8 & \cellcolor{myblue}94.9 & \cellcolor{myblue}61.6 & \cellcolor{myblue}17.2 & \cellcolor{myblue}54.4 & \cellcolor{myblue}66.6 & \cellcolor{myblue}72.7 & \cellcolor{myblue}64.0 & \cellcolor{myblue}64.6 & \cellcolor{myblue}63.5 & \cellcolor{myblue}77.2 & \cellcolor{myblue}64.6 \\
    \hline \bottomrule
    \textbf{V-Branch (ours)} & \XSolid & 96.9 & 93.0 & 88.5 & 73.6 & 97.1 & 98.2 & 60.8 & 17.1 & \textbf{61.3} & \textbf{72.6} & 72.9 & 64.9 & 67.7 & 65.7 & 76.3 & 63.8 \\
    \textbf{V-Branch* (ours)} & \XSolid & \textbf{96.8} & \textbf{93.2} & \textbf{88.8} & \textbf{74.2} & \textbf{97.5} & \textbf{98.4} & \textbf{61.7} & \textbf{18.7} &60.7 & 71.4 & \textbf{75.2} & \textbf{66.1} & \textbf{69.3} & \textbf{66.9} & \textbf{77.2} & \textbf{64.7} \\
    \bottomrule
  \end{tabular}
  }
\end{table*}

{Apart from the inferior performance, training scene-specific models separately also introduces more computation costs (total of 5 $\times$ 85.5M parameters for 5 scenes).
In contrast, our V-Branch only contains 85.8 M parameters.}

\vspace{0.5\baselineskip}
\noindent{ \textbf{- Experiments on Ensemble of ReID Bank.}
While training models separately will cause a domain gap between models, we consider another ensemble method that selects the specific prompts in the ReID Bank in both hard and soft ways. 
Since our ReID Bank models the scene-specific knowledge in prompts and adopts a shared backbone to model the scene-invariant knowledge, it is expected that our ReID Bank can learn the knowledge from different scenes more effectively than training multiple models under different scenes separately.
The results are shown in Table~\ref{rt.1.C}.
On average, ensembling scene-specific prompts in the ReID Bank results in better performance than the ensemble of scene-specific ViT-B (MPDA) models, as shown by Table~\ref{rt.1.B} and Table~\ref{rt.1.C}.
This is mainly because the scene-specific ViT-B (MPDA) models are separately trained and ignore exploiting the scene-shared knowledge while our ReID Bank captures the scene-shared knowledge by a scene-share backbone. Hence, the ensemble of our ReID Bank achieves superior performance to the ensemble of scene-specific ViT-B (MPDA) models.
However, adopting the hard ensemble strategy or soft ensemble strategy with ReID Bank only achieves inferior performance to our VersReID. It is mainly because our VersReID can unify the knowledge from different scenes more effectively than the ensemble strategies.}

{We also investigate a concatenation ensemble. Specifically, we adopt each group of prompts to extract a feature and then concatenate these features as the final feature representation for the input sample, which is used for inference.
The concatenation ensemble is denoted as ReID Bank$^\oplus$, and the results are shown in Table~\ref{rt.1.C}.
We can observe that the performance of the ReID Bank$^\oplus$ is much worse than the V-Branch. 
For example, ReID Bank$^\oplus$ only achieves 64.3\% Rank-1 accuracy on the joint testing set, which is 12.0\% lower than the V-Branch.
The inferior performance of ReID Bank$^\oplus$ is because only one group of prompts is correctly used in ReID Bank$^\oplus$, and the features obtained via other groups of prompts are noisy.
Specifically, since the ReID Bank learns to handle images under different scenes under the guidance of the corresponding prompts, using different prompts may mislead the ReID Bank.
For example, given an image from the clothing-changing scene, ReID Bank will concentrate on clothes-invariant information with the corresponding scene-specific prompts learned in the clothing-change scene.  However, if the ReID Bank uses specific prompts from the occlusion scene to extract features for images from the clothing change scene, it may try to extract information from visible parts of the image and pay attention to clothes, which leads to problematic feature representations. Hence, using all prompts will lead to ineffective integration of knowledge from different scenes.
Differently, our V-Branch learns to unify knowledge from different scenes for multi-scene ReID.
Besides, using all prompts to extract features will make the computation cost in inference linearly grow with the number of scenes, while the inference time of our V-Branch does not depend on the number of scenes. Our VersReID is superior to the concatenation ensemble.}

\renewcommand{\arraystretch}{1.3}
\begin{table*}[tb]
 \caption{{Experimental results of ensemble our ReID Bank.
  \textbf{\colorbox{myblue}{Ensemble with scene labels} means the scene labels are used in inference to apply the ReID Bank with the correct prompts for multi-scene ReID.}
  Otherwise, we ensemble the ReID Bank using different scene-specific prompts conditioned on the \textbf{ViT-Tiny} scene classifier (96.3\% scene classification accuracy).
  Given an image, the scene classifier predicts the probability of the scenes.
  \textbf{The hard ensemble strategy uses the scene-specific prompt from the scene with maximum probability to extract features, and the soft ensemble strategy weights and sums the features obtained from all scene-specific prompts according to the scene classification probability.}
  The results show that our V-Branch significantly outperforms all the ensemble methods, demonstrating the effectiveness of unifying knowledge from different scenes.
  Notably, our V-Branch is evaluated without requiring scene labels and can be further enhanced by overlapping patch embedding layer, denoted as V-Branch*.}}
  \label{rt.1.C}
  \centering
  \resizebox{\linewidth}{!}{
  \begin{tabular}{l | c | c c | c c | c c | c c | c c | c c | c c | c c}
    \toprule
    \makecell[c]{\multirow{3}{*}{\makecell{Methods}}} & \multirowcell{3}{Scene \\ labels} & \multicolumn{4}{c|}{General} & \multicolumn{2}{c|}{Low-resolution} & \multicolumn{4}{c|}{Clothing Change} & \multicolumn{2}{c|}{Occlusion} & \multicolumn{2}{c|}{Cross-modality} & \multicolumn{2}{c}{Joint}\\
    \cline{3-16}
    & & \multicolumn{2}{c|}{Market-1501} & \multicolumn{2}{c|}{MSMT17} & \multicolumn{2}{c|}{MLR-CUHK03} & \multicolumn{2}{c|}{Celeb-ReID} & \multicolumn{2}{c|}{PRCC} & \multicolumn{2}{c|}{Occ-Duke} & \multicolumn{2}{c|}{SYSU-mm01} & \multicolumn{2}{c}{Testing Set} \\
    \cline{3-18}
    & & R-1 & mAP & R-1 & mAP & R-1 & mAP & R-1 & mAP & R-1 & mAP & R-1 & mAP & R-1 & mAP & R-1 & mAP \\
    \hline

    \multicolumn{16}{c}{Different ensembles of \textbf{ReID Bank} using ViT-Tiny as the scene classifier. } \\
        
     \textbf{Hard} Ensemble & \XSolid & 92.9 & 85.0 & 87.2 & 71.1 & 95.5 & 95.8 & 61.0 & 17.3 & 47.5 & 57.1 & 71.4 & 63.3 & 65.4 & 57.3 & 73.3 & 58.7 \\ 
     \textbf{Soft} Ensemble & \XSolid & 94.3 & 88.2 & 87.2 & 71.5 & 96.0 & 96.9 & 61.0 & 17.3 & 53.8 & 63.9 & 71.6 & 63.4 & 66.7 & 58.9 & 73.8 & 59.8 \\ 
    \cellcolor{myblue} Ensemble with Scene labels & \cellcolor{myblue}\Checkmark & \cellcolor{myblue}96.4 & \cellcolor{myblue}92.5 & \cellcolor{myblue}87.9 & \cellcolor{myblue}72.5 & \cellcolor{myblue}96.3 & \cellcolor{myblue}97.6 & \cellcolor{myblue}61.9 & \cellcolor{myblue}17.3 & \cellcolor{myblue}60.0 & \cellcolor{myblue}71.2 & \cellcolor{myblue}71.6 & \cellcolor{myblue}63.9 & \cellcolor{myblue}68.6 & \cellcolor{myblue}68.5 & \cellcolor{myblue}78.7 & \cellcolor{myblue}66.2 \\ 
    \hline
    ReID Bank$^\oplus$ &  \XSolid &  95.8 & 91.3 & 82.6 & 61.2 & 94.3 & 93.4 & 55.7 & 10.4 & 44.9 & 57.5 & 63.4 & 55.2 & 53.4 & 54.6 & 64.3 & 49.8 \\ \hline \bottomrule
    \textbf{V-Branch (ours)} & \XSolid & 96.9 & 93.0 & 88.5 & 73.6 & 97.1 & 98.2 & 60.8 & 17.1 & \textbf{61.3} & \textbf{72.6} & 72.9 & 64.9 & 67.7 & 65.7 & 76.3 & 63.8 \\
    \textbf{V-Branch* (ours)} & \XSolid & \textbf{96.8} & \textbf{93.2} & \textbf{88.8} & \textbf{74.2} & \textbf{97.5} & \textbf{98.4} & \textbf{61.7} & \textbf{18.7} & 60.7 & 71.4 & \textbf{75.2} & \textbf{66.1} & \textbf{69.3} & \textbf{66.9} & \textbf{77.2} & \textbf{64.7} \\
    \bottomrule
  \end{tabular}
  }
\end{table*}

\vspace{0.5\baselineskip}
\noindent{ \textbf{- Potential overfitting.}
We argue that training the model on one scene could cause overfitting. Specifically, when using scene-labels, the ensemble of scene-specific ViT-B (MPDA) models is evaluated by using the corresponding scene-specific model to extract features for images from different scenes (the row with\colorbox{myblue}{gray}background in each Table~\ref{rt.1.B}).
In other words, with the scene labels, the performance of the ensemble model on each scene is evaluated by the scene-specific ViT-B (MPDA) trained on that scene.
In this case, we find that our VersReID outperforms the scene-specific ViT-B (MPDA) on most datasets.
For example, we can observe that the scene-specific ViT-B (MPDA) underperforms our VersReID on the PRCC~\cite{prcc} and SYSU-mm01~\cite{mm01} datasets, where scene-specific ViT-B (MPDA) achieves 54.4\% Rank-1 accuracy on PRCC and 64.6\% Rank-1 accuracy on SYSU-mm01, which is 6.9\% and 3.1\% lower than VersReID.
Note that our VersReID achieves superior performance without using scene labels in inference time.
The inferior performance of the scene-specific model is due to the overfitting of the model, as the PRCC and SYSU-mm01 datasets are relatively small in data scales while our method unifies knowledge from different scenes.
These results also demonstrate that training the model with enough data can promote the forming of a powerful versatile model.}

\vspace{0.5\baselineskip}
{\textbf{In summary}, even with the scene labels, the ensemble of scene-specific models is inferior to our V-Branch in terms of performance and number of parameters.
As for the ensemble of ReID Bank, it can hardly outperform the V-Branch on the joint testing set without the scene labels. 
This is because our ReID Bank captures the scene-specific knowledge in scene-specific prompts, and thus ReID Bank relies on scene labels to properly select feasible prompts.
Through knowledge distillation from ReID Bank to V-Branch, our V-Branch can further unify the knowledge from different scenes with versatile prompts, thus handling multi-scene ReID more effectively and efficiently without requiring scene labels.}

\section{More Analysis on Visualization Results.}
\vspace{0.5\baselineskip}{\noindent{\textbf{- The perspective of prompts.}
When we view Figure~\ref{f.6} in columns, we find that for the same type of prompts, the properties reflected by attention maps are independent of which scene the image is from.
Specifically, regardless of which scene the person image belongs to, the attention maps guided by the same scene-specific prompts roughly focus on similar positions. 
For example, the clothing change prompts guiding attention maps always show high responses to the face and feet, and the attention maps under the guidance of occlusion prompts pay more attention to key points. 
This phenomenon demonstrates that the scene-specific prompts in the ReID Bank contain scene-specific knowledge. 
Even if the backbone network is shared among scenes, different prompts can still help the ReID Bank capture specific features required by every scene, resulting in impressive ReID performances in various scenes.}}

\section{More Implementation Details.}

{We apply the proposed MPDA strategy to DINO~\cite{dino} and learn the ViT-Base (ViT-B)~\cite{vit} with 50\% training data of the LUP-B dataset for 100 epochs following~\cite{transreid_ssl} in a self-supervised manner.
The method is denoted as DINO + MPDA.
We train the model on 4 GPUs using the AdamW optimizer~\cite{adamw} with a batch size of 512. 
During the first 10 epochs, the learning rate linearly increases to 1e-3.  After the warm-up, the learning rate is decayed according to a cosine schedule~\cite{sgdr}. The weight decay also follows a cosine scheduler from 0.04 to 0.4. 
The data augmentation is introduced in Section 4.
In each training iteration, we generate two different views of each source image following previous work in self-supervised learning~\cite{dino, mocov3}.
Note that the MPDA aims to introduce multi-scene variations and is designed for self-supervised contrastive learning on unlabeled person datasets.
Since the multi-scene training set is labeled and each identity has multiple samples in different scenes, we do not use MPDA in the multi-scene training set to generate fake samples to avoid disturbance.
\textbf{All the comparison methods in multi-scene setup use MPDA + LUP-B pretrianing.}}

\vspace{0.5\baselineskip}\noindent{\textbf{- Prompt-based multi-scene joint training.} 
In the first stage, we train ReID Bank on the built multi-scene ReID training set, and \textbf{do not} fine-tune the model on any specific dataset.
We adopt a ViT-B model learned from the DINO + MPDA as the scene-shared transformer backbone in ReID Bank.
No camera information or local features are used. }

{The number of scene-specific prompts for one scene $N$ is 2, and there are 10 prompts from 5 scenes in the prompt pool.
By default, all prompts are initialized randomly. We also attempt to adopt a pre-trained CLIP~\cite{clip} text encoder to extract the textual embedding for the prompt initialization but do not gain improvements. 
Moreover, the model is trained on 1 GPU for 120 epochs using an SGD with momentum optimizer~\cite{sgd}.
The momentum and weight decay of the optimizer are set to 0.9 and 1e-4, respectively.
The mini-batch size is set to 128 consisting of 32 persons (4 images per person).
A cosine decaying scheduler is used with a base learning rate of 4e-4, and the learning rate linearly increases from 0 to 4e-4 during the first 20 epochs.}

\ifCLASSOPTIONcaptionsoff
\newpage
\fi

\bibliographystyle{IEEEtran}
\bibliography{VersReID}